\documentclass{article}
\usepackage[utf8]{inputenc}
\usepackage[T1]{fontenc}     
\usepackage{ragged2e}        
\usepackage[protrusion=true,expansion=false]{microtype}  
\usepackage[hang,flushmargin]{footmisc} 
\usepackage{graphicx}
\usepackage{multirow}
\usepackage{caption}
\usepackage{float}
\usepackage{booktabs}
\usepackage{multicol}
\usepackage{tablefootnote}
\usepackage{siunitx}
\usepackage{tabularx}        
\usepackage[table,xcdraw]{xcolor}
\usepackage{amsmath,amssymb,amsfonts}
\usepackage{mathtools} 
\usepackage{amsthm}
\usepackage{mathrsfs}
\usepackage{algorithm}
\usepackage{algpseudocode}  
\usepackage{algorithmicx}

\usepackage{textcomp}
\usepackage{pifont}
\usepackage{setspace,lineno}
\usepackage{listings}
\usepackage{soul}            
\sethlcolor{yellow}          
\usepackage[title]{appendix}
\usepackage{cite}
\usepackage{manyfoot}
\usepackage{xcolor}
\definecolor{customgreen}{HTML}{00B050}
\definecolor{captionblue}{HTML}{0070C0}  
\newcommand{\cmark}{\textcolor{customgreen}{\ding{51}}} 
\newcommand{\xmark}{\textcolor{red}{\ding{55}}}
\newcommand{\x}[1]{\textcolor{black}{#1}}
\DeclareCaptionLabelFormat{boldblue}{\textbf{\textcolor{captionblue}{#1 #2}}}
\usepackage{titlesec}
\titleformat{\section}
  {\normalfont\Large\bfseries\color{captionblue}}{\thesection}{1em}{}
\titleformat{\subsection}
  {\normalfont\large\bfseries\color{captionblue}}{\thesubsection}{1em}{}
\titleformat{\subsubsection}
  {\normalfont\normalsize\bfseries\color{captionblue}}{\thesubsubsection}{1em}{}
\usepackage{geometry}
\usepackage{amsmath} 
\allowdisplaybreaks[4]
\usepackage[protrusion=true,expansion=false]{microtype}
\usepackage{hyperref}
\hypersetup{
    colorlinks=true,
    linkcolor=blue,
    citecolor=blue,
    urlcolor=blue
}

\usepackage{fvextra}
\usepackage{xcolor}

\DefineVerbatimEnvironment{PromptBlock}{Verbatim}{
  breaklines=true,
  breakanywhere=true,
  fontsize=\scriptsize,
  frame=single,
  framesep=2mm,
  bgcolor=gray!5,
  baselinestretch=0.95
}
\title{Generative Artificial Intelligence in Bioinformatics: A Systematic Review of Models, Applications, and Methodological Advances}
\author{
Wasimul Karim\textsuperscript{1,2,a},
Riasad Alvi\textsuperscript{1,2,a}, 
Sayeem Been Zaman\textsuperscript{1,2,a},\\
Arefin Ittesafun Abian\textsuperscript{1,2,b}, 
Mohaimenul Azam Khan Raiaan\textsuperscript{3,b,*},
Saddam Mukta\textsuperscript{4}, \\
Md Rafi Ur Rashid\textsuperscript{5}, 
Md Rafiqul Islam\textsuperscript{6}, 
Yakub Sebastian\textsuperscript{6},
Sami Azam\textsuperscript{6,b,*}\\
\small
\textsuperscript{1}Applied Artificial Intelligence and INtelligent Systems (AAIINS) Laboratory, Dhaka 1217, Bangladesh\\
\small
\textsuperscript{2}Department of Computer Science and Engineering, United International University, Dhaka 1212, Bangladesh \\
\small
\textsuperscript{3}Department of Data Science and Artificial Intelligence, Monash University, Clayton, VIC, 3153, Australia \\
\small
\textsuperscript{4}Department of Software Engineering, Lappeenranta-Lahti University of Technology, Lappeenranta, 53850, Finland \\
\small
\textsuperscript{5}Department of Computer Science and Engineering, Pennsylvania State University, State College, PA 16801, USA \\
\small
\textsuperscript{6}Faculty of Science and Technology, Charles Darwin University, Casuarina, NT 0909, Australia\\
\small
\textsuperscript{a}Equal Contributions\\
\small
\textsuperscript{b}Equal Supervision\\
\small
\textsuperscript{*}Correspondence: 
\href{mailto:mohaimenul.raiaan@monash.edu}{mohaimenul.raiaan@monash.edu},
\href{mailto:sami.azam@cdu.edu.au}{sami.azam@cdu.edu.au}
}
\date{} 
\begin{document}
\justifying
\twocolumn[
\maketitle
\begin{abstract} 
\noindent Generative artificial intelligence (GenAI) has become a transformative approach in bioinformatics that often enables advancements in genomics, proteomics, transcriptomics, structural biology, and drug discovery. To systematically identify and evaluate these growing developments, this review proposes six research questions (RQs), according to the preferred reporting items for systematic reviews and meta-analysis methods. The objective is to evaluate \x{influential} GenAI strategies \x{with respect to} methodological advancement, predictive performance, and specialization, and to identify promising approaches for advanced modeling, data-intensive discovery, and integrative biological analysis. RQ1 highlights diverse applications across multiple bioinformatics subfields (sequence analysis, molecular design, and integrative data modeling), which demonstrate superior performance over traditional methods through pattern recognition and output generation. RQ2 reveals that adapted specialized model architectures outperformed general-purpose models, an advantage attributed to targeted pretraining and context-aware strategies. RQ3 identifies significant benefits in the bioinformatics domains, focusing on molecular analysis and data integration, that improves accuracy and reduces errors in complex analysis. RQ4 indicates improvements in structural modeling, functional prediction, and synthetic data generation, validated by established benchmarks. RQ5 suggests the main constraints, such as lack of scalability and biases in data that impact generalizability, and proposes future directions focused on robust evaluation and biologically grounded modeling. RQ6 examines that molecular datasets (such as UniProtKB and ProteinNet12), cellular datasets (such as CELLxGENE and GTEx), and textual resources (such as PubMedQA and OMIM) broadly support the training and generalization of GenAI models. This review highlights the potential of GenAI to progress computational biology by addressing recent advances in bioinformatics.
\end{abstract}

\vspace{0.5cm}
\vspace{1em}
]


\section{Introduction}
\x{Generative artificial intelligence (GenAI) refers to a group of artificial intelligence models capable of creating new content, such as text, images, videos, reports and problem-solving solutions, by learning complex patterns from data and producing outputs that resemble human creativity and adaptability \cite{he2025generative}.} In recent years, bioinformatics has sustained an exceptional transformation, largely enabled by advances in GenAI \cite{das2025generative}. Traditional biological data analysis has mainly relied on rule-based methods, statistical models, and specialized algorithms developed for specific tasks \cite{libbrecht2015machine, glaab2012using}. However, the large amount and complexity of modern omics data, ranging from genomic sequences to protein structures to single-cell transcriptomes, present significant challenges to the traditional approaches \cite{angermueller2016deep}. Recent advances in generative models, particularly transformer-based and foundation model approaches, are reconstructing computational biology \cite{zhang2023applications, guo2025foundation}. Following these advancements, this paper presents a comprehensive review of GenAI in bioinformatics that summarizes recent progress in generative modeling, categorizes key applications across genomics, proteomics, and multi-omics, and outlines challenges that hamper reliable deployment, while also highlighting future directions for advancing robust and interpretable bioinformatics modeling. This is essential, because GenAI is increasingly being established across all levels of biological research that include new innovations in data interpretation, molecular design, and systems-level analysis.

\subsection{Strengths of GenAI for Biological Data}
GenAI models \x{outperform traditional approaches in capturing} contextual relationships from large, unlabeled datasets and are particularly effective in biological tasks where data are often noisy or unannotated \cite{ruan2026large, mardikoraem2023generative}. By training on large-scale collections of sequences, molecular graphs, or gene expression profiles, the models can learn high-dimensional patterns and perform downstream inference tasks with minimal supervision \cite{mardikoraem2023generative}. Compared to traditional models, generative approaches provide enhanced flexibility, enabling zero-shot, few-shot, and transfer learning \cite{meier2021language}. Furthermore, \x{GenAI models now capture structural, spatial, and functional properties} of biological systems, supporting diverse biological modeling tasks \cite{ni2023generative, huang2024qust}. These models advance applications from gene controlling element prediction to protein function annotation, drug target interaction modeling, and cellular attribute generation from single-cell data. The ability to generate biologically consistent outputs, integrate heterogeneous modalities, and support natural language interaction has substantially increased the scope of bioinformatics research \cite{ibrahim2025generative}.

\subsection{Gaps in Current Research}
Existing review papers \cite{li2024progress, guo2025foundation, sarumi2024large, asim2025dna, asim2025protein} have explored the growing role of large language models (LLMs) and foundation models in bioinformatics, focusing on diverse areas such as protein structure prediction, drug discovery, gene expression analysis, and regulatory sequence annotation; however, only a few existing reviews have provided a thorough evaluation of generative artificial intelligence in bioinformatics. Among them, four studies \cite{li2024progress, guo2025foundation, sarumi2024large, asim2025dna} have explored multimodal integration, protein modeling, and applications on drug discovery, while only two \cite{li2024progress, guo2025foundation} discussed instruction tuning in detail. Although multi-agent learning systems, automation, and conversational interfaces represent important advances toward intelligent bioinformatics systems, the limited exploration and integration in current research restrict the full potential of GenAI-driven discovery and analysis. Existing studies \cite{li2024progress, guo2025foundation, sarumi2024large, asim2025dna, asim2025protein} have completely overlooked these system-level aspects and have created a significant gap in understanding how these techniques enable seamless integration, real-time interaction, and scalable deployment of GenAI systems within bioinformatics workflows. To address these gaps, our study aims to cover eleven topics, including multi-domain coverage (MDC), generative model focus (GMF), model architecture diversity (MAD), instruction tuning and prompting (ITP), tokenization and embedding design (TED), multi-agent learning systems (MALS), computational GenAI workflow automation (CGWA), conversational interfaces (CI), multi-modal integration (MMI), protein design and structure biology (PDSB), and drug discovery application (DDA), thereby providing a broad and integrative understanding of how GenAI is reshaping bioinformatics research and practice.

\subsection{Purpose and Research Questions(RQs)}
The objective of the review is to critically evaluate the evolving role of GenAI in bioinformatics, its applications, model architectures, and impact across specific biological domains. This review is inspired by six fundamental research questions (RQs) \x{to evaluate GenAI methodologies, assess domain-specific performance, and identify emerging challenges.}

\vspace{0.5em}
\noindent \textbf{RQ1:} How are GenAI models applied in bioinformatics? Specifically, what novel tasks have they enabled, and how do they compare to traditional methods? \\
\textit{Rationale:} To understand how GenAI models are applied in bioinformatics and how they advance beyond traditional approaches.

\vspace{0.3em}
\noindent \textbf{RQ2:} Which models and architectures are most effective? For instance, how do general-purpose LLMs perform compared to domain-specific models on biological sequence data? \\
\textit{Rationale:}  To identify which model types and architectures perform best for biological data.

\vspace{0.3em}
\noindent \textbf{RQ3:} Which bioinformatics domains benefit most from GenAI? \\
\textit{Rationale:}  To determine which areas of bioinformatics gain the most from GenAI adoption.

\vspace{0.3em}
\noindent \textbf{RQ4:} How have these models improved tasks such as protein structure/function prediction, multi-omics data integration, and code development in bioinformatics? What benchmarks or use cases demonstrate their impact? \\
\textit{Rationale:} To evaluate how GenAI improves key biological tasks and the evidence supporting these advances.

\vspace{0.3em}
\noindent \textbf{RQ5:} What are the current limitations, challenges, and future directions of generative models in terms of data requirements, interpretability, and integration of biological knowledge? \\
\textit{Rationale:} To identify barriers to scalability, transparency, and biological integration in GenAI models, while outlining potential future directions for advancement.

\vspace{0.3em}
\noindent \textbf{RQ6:} How do molecular, cellular, and textual or knowledge-based datasets contribute individually and collectively to the training and generalization of GenAI models? \\
\textit{Rationale:} To understand the way different biological data modalities shape the learning, integration, and generalization capabilities of GenAI models.

\begin{figure*}[ht!]
\centering
\includegraphics[scale=0.17]{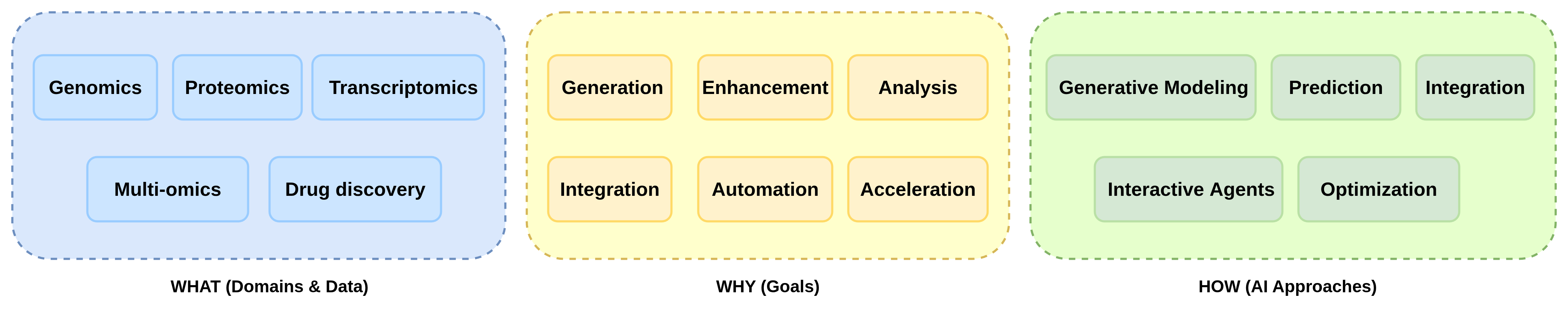}
\caption{Overview of the categorization framework of the review based on three research question dimensions: WHAT describes the bioinformatics domains and biological data modalities where GenAI is applied, WHY explains the goals for applying GenAI, and HOW outlines the AI approaches and tasks.}
\label{fig:wwh}
\end{figure*}

The framework shown in Figure \ref{fig:wwh} organizes the review by the three guiding questions WHAT, WHY, and HOW, clarifying the domains of application, underlying motivations, and computational strategies in GenAI-enabled bioinformatics.

\subsection{Contributions}
This paper presents six primary contributions to the study of GenAI in bioinformatics:
\begin{itemize}
    \item A systematic literature review presents a comprehensive analysis of GenAI applications in genomic sequence modeling, protein representation and design, drug discovery, and single-cell analysis, highlighting the new capabilities enabled by these models. 
    \item A comparison of existing reviews is provided, showing their areas of attention and revealing gaps in current research.
    \item Six RQs are formulated to examine the applications, model architectures, domain-specific performance, and potential challenges to demonstrate the growing impact across diverse biological tasks. 
    \item A detailed discussion of the task-specific improvements and benchmarks in recent studies to assess the effectiveness of generative models in bioinformatics.
    \item Proposed future research directions focusing on the development of biologically grounded GenAI frameworks, including the use of LLMs as reasoning modules and grounding outputs in verifiable tools to improve reliability. New methods for multi-modal biological integration are also suggested to develop advanced bioinformatics-grounded GenAI models for scientific discovery.
    \item A detailed analysis of the key datasets, classified into molecular, cellular, and textual modalities, was conducted to identify and summarize their roles and resources being used for GenAI and LLM applications in bioinformatics.
    
\end{itemize}

\subsection{Paper Organization}
\begin{figure*}[ht!]
\centering
\includegraphics[scale=0.36]{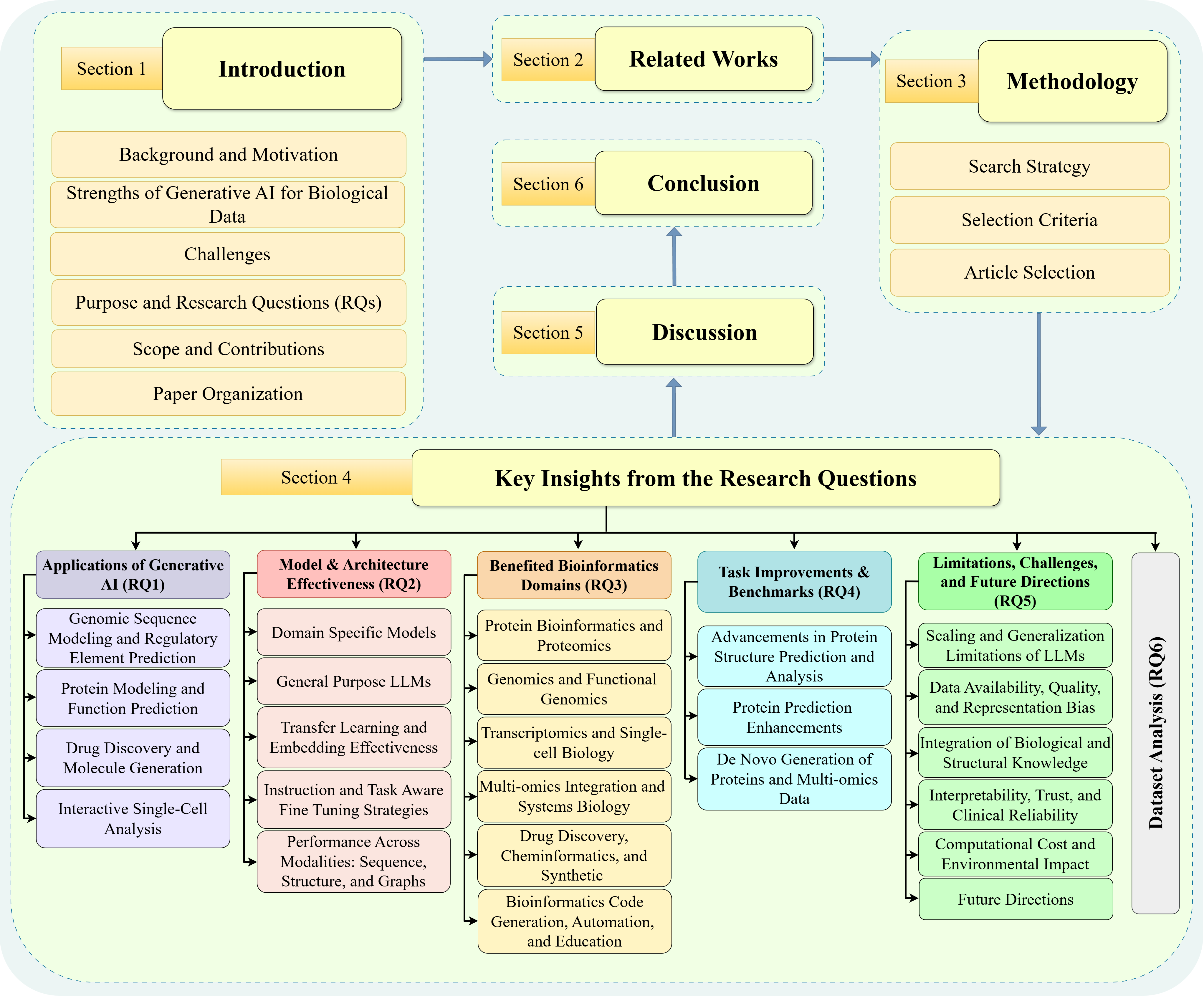}
\caption{A visual outline of the sections, indicating the flow from introduction and methodology through RQ answers, discussion, limitations, and future directions.}
\label{fig:org}
\end{figure*}

The paper is divided into 6 sections. The remainder of the study is structured as follows:
Section \ref{RW} discusses recent related works, Section \ref{meth} outlines the methodology followed in this review process, and Section \ref{RQs} highlights the core RQs and their corresponding answers that frame this study. Section \ref{discussion} provides a detailed discussion and offers insights and observations outlined from the reviewed literature. Finally, Section \ref{conclusion} concludes the paper with the main findings and reflects on the growing role of GenAI within bioinformatics. Figure \ref{fig:org} overall represents the organization of the review.

\section{Related Works}
\label{RW}
The integration of GenAI into bioinformatics has led to significant changes in how biological data are processed and understood. Existing research \cite{li2024progress, guo2025foundation, sarumi2024large, asim2025dna, asim2025protein} covers several areas, including the development of foundational models, system-level analyses, and domain-specific applications in genomics and proteomics. This section reviews these studies, focusing on their methods, areas of application, and the key limitations that remain in the field.

Several researchers have examined generative approaches within diverse bioinformatics domains. Sarumi et al. \cite{sarumi2024large} outlined this revolution, describing how transformer-based models such as Bidirectional Encoder Representations from Transformers (BERT) and Generative Pre-trained Transformer (GPT) have penetrated key areas, including protein structure prediction, drug discovery, gene expression analysis, and regulatory sequence annotation. They focused on mapping model innovations into practice and prioritizing interpretability by focusing on attention-based mechanisms. However, it lacks inter-modal demonstrations, a unified benchmark or strategy for evaluating LLM performance, and theoretical or system-level insights, focusing instead on task-level demonstrations.

In another paper, Li et al. \cite{li2024progress} provided an analysis that followed the historical evolution and architectural variety of foundation models (FMs) in bioinformatics. In contrast, their review provided a structural breakdown of FMs into discriminative and generative types, listing specific examples \x{such as} BioBERT, ESM, ProtGPT2, and ESM3. They also discussed methodological pipelines, including zero-shot, few-shot, and fine-tuning approaches, which can be used to transfer general-domain knowledge to biological problems. However, the review focuses too heavily on architectural classification while giving little attention to cross-domain validation, interpretability, and benchmarking reproducibility. It also neglects the practical impact of generative models such as ESM3 or ProtGPT2 on real biological tasks, limiting its translational applicability, and emphasizes models over the feasibility and reproducibility of evaluation standards.

At the same time, Guo et al. \cite{guo2025foundation} showed the immense impact of multimodal pre-trained foundation models across genomics, proteomics, transcriptomics, drug discovery, and single-cell analysis. Unlike previous narrative reviews, their study classified FMs into four groups, which included language, vision, graph, and multimodal, and aligned them systematically with post-biological tasks. The value of this review is its panoramic taxonomy and the clear classification of model architectures and applications, providing a system-level perspective of the manner in which multimodal \x{pretraining} mediates the rich and poor data environments. However, the review focuses more on scope rather than depth, which does not cover a study of interpretability, biological rationale, and consistency of validation, and thus is more of a review of work than an analytical assessment of cross-domain knowledge.

\begin{table*}[ht!]
\centering
\scriptsize
\caption{Comparison with state-of-the-art studies, \x{showing  which of the eleven topics each review addresses} and illustrating how the current review expands beyond previous work.}
\begin{tabular}{l c ccccccccccc}
\toprule
     &      & \multicolumn{11}{c}{\textbf{Topic Covered}} \\
\cmidrule(lr){3-13}
Paper & Year & MDC & GMF & MAD & ITP & TED & MALS & CGWA & CI & MMI & PDSB & DDA \\
\midrule
\x{Guo et al.} \cite{guo2025foundation}     & \x{2025} & \cmark & \cmark & \cmark & \cmark & \cmark & \xmark & \xmark & \xmark & \cmark & \cmark & \cmark \\
\x{Li et al.} \cite{li2024progress}     & \x{2024} & \cmark & \cmark & \cmark & \cmark & \cmark & \xmark & \xmark & \xmark & \cmark & \cmark & \cmark \\
\x{Sarumi et al.} \cite{sarumi2024large}     & \x{2024} & \cmark & \cmark & \cmark & \xmark & \cmark & \xmark & \xmark & \xmark & \cmark & \cmark & \cmark \\
\x{Asim et al.} \cite{asim2025dna}     & \x{2025} & \cmark & \xmark & \cmark & \xmark & \cmark & \xmark & \xmark & \xmark & \xmark & \xmark & \xmark \\
\x{Asim et al.} \cite{asim2025protein}     & \x{2025} & \cmark & \cmark & \cmark & \xmark & \cmark & \xmark & \xmark & \xmark & \cmark & \cmark & \cmark \\

Ours  & 2025 & \cmark & \cmark & \cmark & \cmark & \cmark & \cmark & \cmark & \cmark & \cmark & \cmark & \cmark \\
\bottomrule
\end{tabular}
\label{tab:survey_comparison_full}
\end{table*}

Asim et al. \cite{asim2025dna} presented DNA sequence analysis, listing 44 tasks with specified learning objectives such as classification, regression, or clustering, and methodically mapping 67 language models and 39 embedding techniques to 140 manually curated datasets. This careful mapping offers concrete advice to both biologists and computational researchers. However, their work scope is on genomics only, \x{limiting} its relevance to proteomics or multi-omics research. Finally, Asim et al. \cite{asim2025protein} focused on protein sequence analysis, categorizing 63 tasks into eleven biological goals, including function prediction, subcellular localization, mutation analysis, and drug binding. Their review covers 627 datasets, 68 biological databases, and 25 embedding methods, and provides a taxonomy of AI patterns such as binary, multi-class, multi-label classification, regression, and clustering applied to protein analysis. However, their study is proteome-centered and barely touches upon new examples such as generative \x{pretraining} or multi-modal biological representation. 

Although recent studies \cite{li2024progress, guo2025foundation, sarumi2024large, asim2025dna, asim2025protein} have explored several aspects of GenAI in bioinformatics, they have ignored some of the scope of methodological and system-level focused areas. To address these areas, our study systematically covers eleven topics that include MDC, GMF, MAD, ITP, TED, MALS, CGWA, CI, MMI, PDSB \& DDA. Most of the previous studies focused on selected areas: all of them \cite{li2024progress, guo2025foundation, sarumi2024large, asim2025dna, asim2025protein} discussed MDC and MAD, four studies \cite{li2024progress, guo2025foundation, sarumi2024large, asim2025dna} explored GMF, MMI, PDSB, and DDA, and two \cite{li2024progress, guo2025foundation} addressed ITP; however, they largely overlooked three core topics, including MALS, CGWA, and CI. In addition, existing reviews hardly integrate contexts from protein design, structural biology, and drug discovery, which result in a partial understanding of the field. Our study addresses all eleven topics with the key research questions in depth, mainly less explored areas, and therefore offers a clearer understanding of the field. Table \ref{tab:survey_comparison_full} highlights gaps in prior literature across key methodology and application in GenAI for bioinformatics, addressing eleven core topics that provide the reasoning of our study.

\section{Methodology}
\label{meth}
Through the review process, a detailed methodology has been followed in which the relevant literature on GenAI applications in bioinformatics was identified, chosen, and reviewed. To ensure both completeness and progressive relevance, a structured search has been conducted across reputable academic databases, using carefully chosen search terms that describe the intersection of generative modeling and computational biology. Clear inclusion and exclusion criteria were applied to gather studies of high relevance and quality, providing a strong basis for an effective and reliable analysis.

\begin{table*}[ht!]
\centering
\scriptsize
\caption{Complete list of the 82 studies included in this review, categorized according to their primary bioinformatics domain.}
\label{tab:included_studies}
\begin{tabularx}{\textwidth}{
>{\raggedright\arraybackslash}p{3cm}
>{\centering\arraybackslash}p{3.3cm}
>{\centering\arraybackslash}p{2cm} |
>{\raggedright\arraybackslash}p{3cm}
>{\centering\arraybackslash}p{3.3cm}
>{\centering\arraybackslash}p{2cm}}
\toprule
\textbf{Study} & \textbf{Domain} & \textbf{Ref.} &
\textbf{Study} & \textbf{Domain} & \textbf{Ref.} \\
\midrule

Ji et al.  & Genomics & \cite{ji2021dnabert} & Cui et al. & Transcriptomics & \cite{cui2024scgpt} \\
Lyu et al.  & Genomics & \cite{lyu2024gp} & Yang et al. & Transcriptomics & \cite{yang2022scbert} \\
Jiani et al.  & Genomics & \cite{ma2024bingo} & Xu et al. & Transcriptomics & \cite{xu2022scican} \\
Chang et al.  & Genomics & \cite{chang2024gene} & Yang et al. & Transcriptomics & \cite{yang2026large} \\
Jin et al.  & Genomics & \cite{jin2024genegpt} & Riffle et al. & Transcriptomics & \cite{riffle2025olaf} \\
Harrigan et al.  & Genomics & \cite{harrigan2024improvements} & Mathur et al. & Transcriptomics & \cite{mathur2025pyevocell} \\
Dalla et al.  & Genomics & \cite{dalla2025nucleotide} & Fu et al. & Transcriptomics & \cite{fu2025foundation} \\
Li et al.  & Genomics & \cite{li2025large} & Wan et al. & Transcriptomics & \cite{wan2023integrating} \\
Sanabria et al.  & Genomics & \cite{sanabria2024dna} & Sidorenko et al. & Transcriptomics & \cite{sidorenko2024precious2gpt} \\
Zhou et al. & Genomics & \cite{zhou2024dnabert} & Shree et al. & Transcriptomics & \cite{shree2023scdreamer} \\
Shao et al. & Genomics & \cite{shao2024long} & Abir et al. & Transcriptomics & \cite{abir2025biollmnet} \\
Zhu et al. & Genomics & \cite{zhu2025enhancing} & Shulgina et al. & Transcriptomics & \cite{shulgina2024rna} \\
Bendani et al. & Genomics & \cite{bendani2025decision} & Penic et al. & Transcriptomics & \cite{penic2025rinalmo} \\
Meier et al. & Proteomics & \cite{meier2021language} & Chen et al. & Transcriptomics & \cite{chen2022interpretable} \\
Bo et al. & Proteomics & \cite{chen2024xtrimopglm} & Yin et al. & Transcriptomics & \cite{yin2025ernie} \\
Zhang et al. & Proteomics & \cite{zhang2022t4sefinder} & Ken et al. & Transcriptomics & \cite{chen2024self} \\
Liu et al. & Proteomics & \cite{liu2025phosf3c} & Wu et al. & Transcriptomics & \cite{wu2026generative} \\
Rehana et al. & Proteomics & \cite{rehana2024evaluating} & Hossain et al. & Transcriptomics & \cite{hossain2025interpretable} \\
Vitale et al. & Proteomics & \cite{vitale2024evaluating} & La et al. & Transcriptomics & \cite{la2025gl4sda} \\
Andrew et al. & Proteomics & \cite{dickson2024fine} & Green et al. & Transcriptomics & \cite{green2025litsumm} \\
Kaminski et al. & Proteomics & \cite{kaminski2023plm} & Sadia et al. & Transcriptomics & \cite{sadia2026crossllm} \\
McWhite et al. & Proteomics & \cite{mcwhite2023leveraging} & Shen et al. & Transcriptomics & \cite{shen2024accurate} \\
Becker et al. & Proteomics & \cite{becker2024learnmsa2} & De et al. & Transcriptomics & \cite{de2026redac} \\
Badkul et al. & Proteomics & \cite{badkul2024trustaffinity} & Xiao et al. & Transcriptomics & \cite{xiao2024rna} \\
Capela et al. & Proteomics & \cite{capela2025comparative} & Shahgir et al. & Transcriptomics & \cite{shahgir2024rna} \\
Wei et al. & Proteomics & \cite{liu2025drbioright} & Weissenow et al. & Structural Bioinformatics & \cite{weissenow2022protein} \\
Smith et al. & Proteomics & \cite{smith2025funcfetch} & Nallapareddy et al. & Structural Bioinformatics & \cite{nallapareddy2023cathe} \\
Ferruz et al. & Proteomics & \cite{ferruz2022protgpt2} & Chao et al. & Structural Bioinformatics & \cite{wang2024protchatgpt} \\
Madani et al. & Proteomics & \cite{madani2023large} & Pantolini et al. & Structural Bioinformatics & \cite{pantolini2024embedding} \\
Nijkamp et al. & Proteomics & \cite{nijkamp2023progen2} & Draizen et al. & Structural Bioinformatics & \cite{draizen2024deep} \\
Brandes et al. & Proteomics & \cite{brandes2022proteinbert} & Lin et al. & Structural Bioinformatics & \cite{lin2023evolutionary} \\
Noelia et al. & Proteomics & \cite{ferruz2022controllable} & Gurusinghe et al. & Structural Bioinformatics & \cite{gurusinghe2025probass} \\
Jia-Ying et al. & Proteomics & \cite{chen2025evaluating} & Pickard et al. & Computational Biology & \cite{pickard2024language} \\
Wang et al. & Proteomics & \cite{wang2025prot2chat} & Haibo et al. & Computational Biology & \cite{jin2024promptmrg} \\
Dan et al. & Proteomics & \cite{liu2025plm} & Mitchener et al. & Computational Biology & \cite{mitchener2025bixbench} \\
Uludougan et al. & Drug Discovery & \cite{uludougan2022exploiting} & Taylor et al. & Computational Biology & \cite{taylor2022galactica} \\
Ye et al. & Drug Discovery & \cite{ye2025drugassist} & Nair et al. & Biomedical Informatics & \cite{nair2025prediction} \\
Moayedpour et al. & Drug Discovery & \cite{moayedpour2024representations} & Luo et al. & Biomedical Informatics & \cite{luo2022biogpt} \\
Born et al. & Drug Discovery & \cite{born2023regression} & Bolton et al. & Biomedical Informatics & \cite{bolton2024biomedlm} \\
Piccolo et al. & Bioinformatics Benchmark & \cite{piccolo2023evaluating} & Karkera et al. & Microbiome Informatics & \cite{karkera2023leveraging} \\
Tang et al. & Bioinformatics Benchmark & \cite{tang2024biocoder} & Mehandru et al. & Agentic System & \cite{mehandru2025bioagents} \\

\bottomrule
\end{tabularx}
\end{table*}

\subsection{Search Strategy}
\textbf{Search Sources:}
The primary sources of literature include peer-reviewed publications from leading journals in bioinformatics, computational biology, and AI. Specifically, ScienceDirect , Oxford Academic , NeurIPS , SpringerNature , BMC , arXiv, and Google Scholar were utilized. These databases were selected based on their relevance, credibility, and broad indexing of GenAI research applied to bioinformatics. \x{The literature search and study were conducted between May 13, 2025 and November 20, 2025 to ensure recent coverage of the rapidly evolving field of generative models in bioinformatics. A final update search was performed on June 11, 2026 to capture newly published studies.} Focusing on recent years ensures that the review captures the most relevant and state-of-the-art (SOTA) developments in the field.

\textbf{Search Terms:}
A targeted list of search terms has been developed to capture the intersection of GenAI and bioinformatics. These include:

\begin{itemize}
    \item GenAI in bioinformatics
    \item Large language models in bioinformatics
    \item Protein language models
    \item Generative deep-learning methods in bioinformatics
    \item Transformer models in genomics
    \item Synthetic biology with GenAI
    \item Multi-omics generative framework
    \item Language models fine-tuned on bioinformatics data
    \item LLM in protein design and function prediction
\end{itemize}

\x{The following Boolean search strings were used across the platforms:}

\x{\textbf{ScienceDirect:} ("protein language model" OR "genomic language model") AND ("generative AI" OR "large language model") AND ("bioinformatics" OR "computational biology")}

\x{\textbf{Oxford Academic:} ("generative deep learning" OR "foundation model") AND ("protein design" OR "genomics" OR "transcriptomics")}

\x{\textbf{ArXiv / NeurIPS:} ("LLM" AND "bioinformatics") AND ("generative model" OR "transformer")}

\x{\textbf{Nature:} ("protein language model" AND "bioinformatics" AND "generative AI")}

\x{\textbf{BMC:} ("large language model" OR "generative AI") AND ("multi-omics" OR "protein function" OR "gene expression")}

\subsection{Selection Criteria}
\begin{figure*}[ht!]
\centering
\includegraphics[scale=0.5]{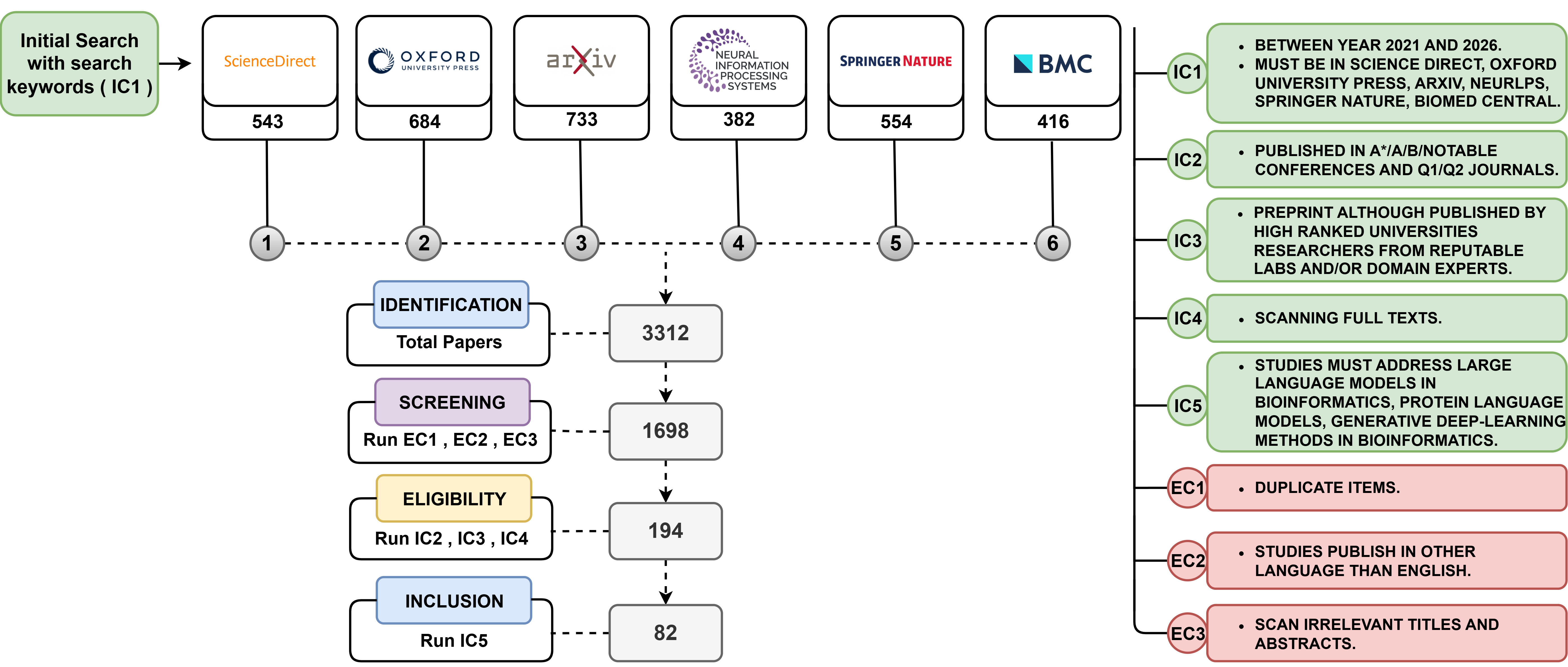}
\caption{Inclusion and exclusion \x{criteria} used for selecting articles based on predefined criteria such as publication year, domain relevance, and methodological quality.}
\label{fig:icec}
\end{figure*}

\x{All retrieved records were exported into EndNote. Duplicate records were identified using DOI, title, and author matching and subsequently removed before screening. After duplicate removal, 3312 unique records remained. After duplicate removal, titles and abstracts were screened to assess relevance to GenAI applications in bioinformatics. A quality assessment of the included studies was performed using a predefined checklist considering study design, dataset quality, and model validation strategy. Three independent reviewers screened all records. Potentially eligible studies underwent full-text review against the predefined inclusion and exclusion criteria. Disagreements between reviewers were resolved through discussion. No formal inter-rater agreement statistic was calculated. The final list of selected articles has been produced and critically discussed through careful evaluation to meet the inclusion criteria (IC) and exclusion criteria (EC) applied in this review, as shown in Figure \ref{fig:icec}.}

\x{In this review, GenAI is defined broadly to include generative models, foundation models, bioinformatics LLMs, and agentic systems that derive their capabilities from large-scale self-supervised or generative pretraining \cite{bommasani2021opportunities, dip2026large}. Following recent foundation-model literature, we consider a model within scope if it learns transferable representations from large unlabeled biological datasets and supports generation, reasoning, simulation, or adaptation across multiple downstream tasks. Therefore, protein language models, genomic foundation models, RNA foundation models, multimodal foundation models, and LLM-based biological agents are included. In contrast, conventional supervised predictors and feature-engineered machine-learning pipelines are excluded unless they serve as comparative baselines. This framework reflects the current evolution of GenAI in bioinformatics, where generative pretraining and foundation-model adaptation increasingly underpin both generative and predictive biological applications.}

\subsection{Article Selection}
\begin{figure}[ht!]
\centering
\includegraphics[scale=0.2]{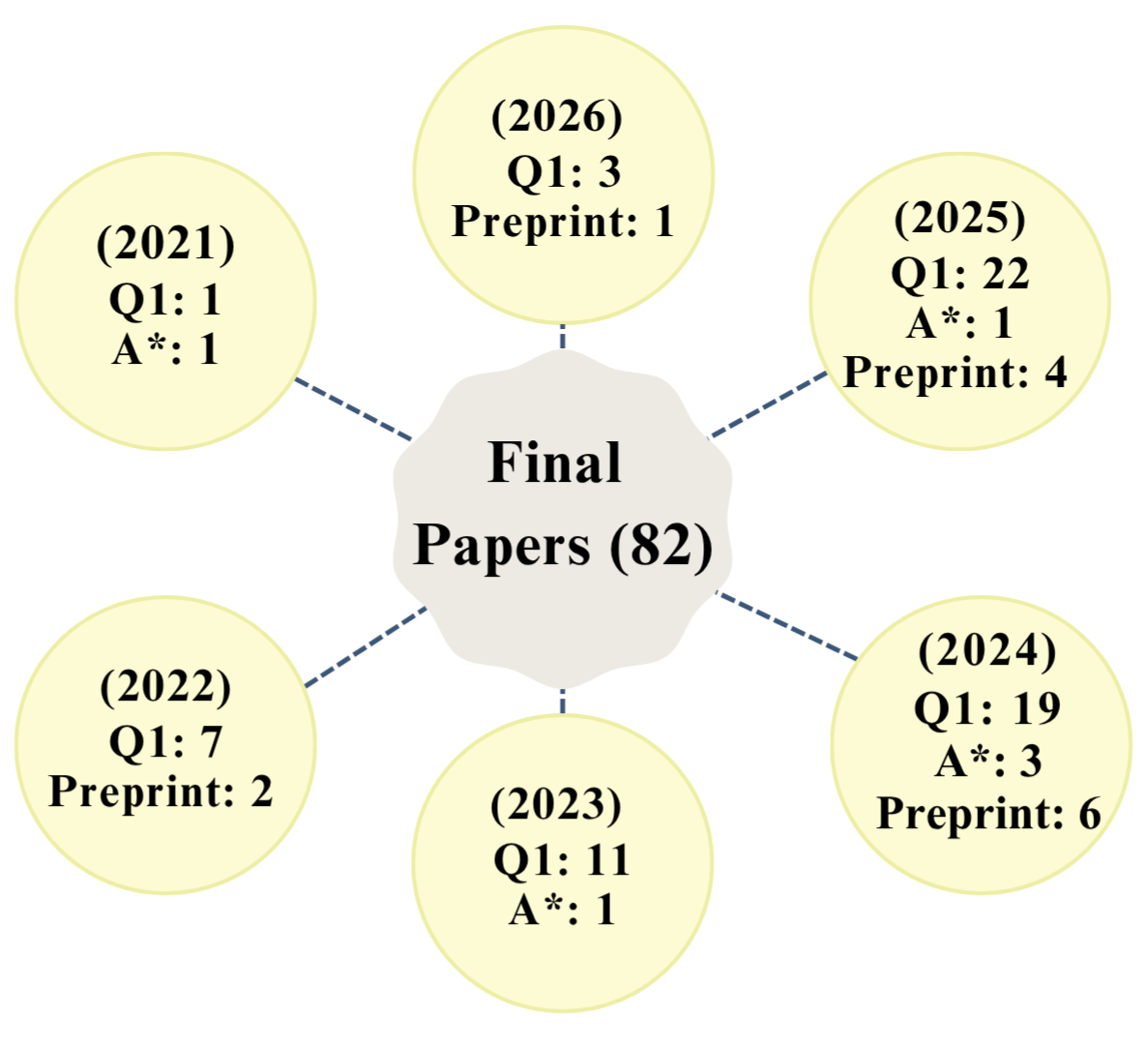}
\caption{Year-wise distribution of selected articles between 2021 and 2026, highlighting the growing trend of GenAI research in bioinformatics.}
\label{fig: paper}
\end{figure}

\begin{figure}[ht!]
\centering
\includegraphics[scale=0.29]{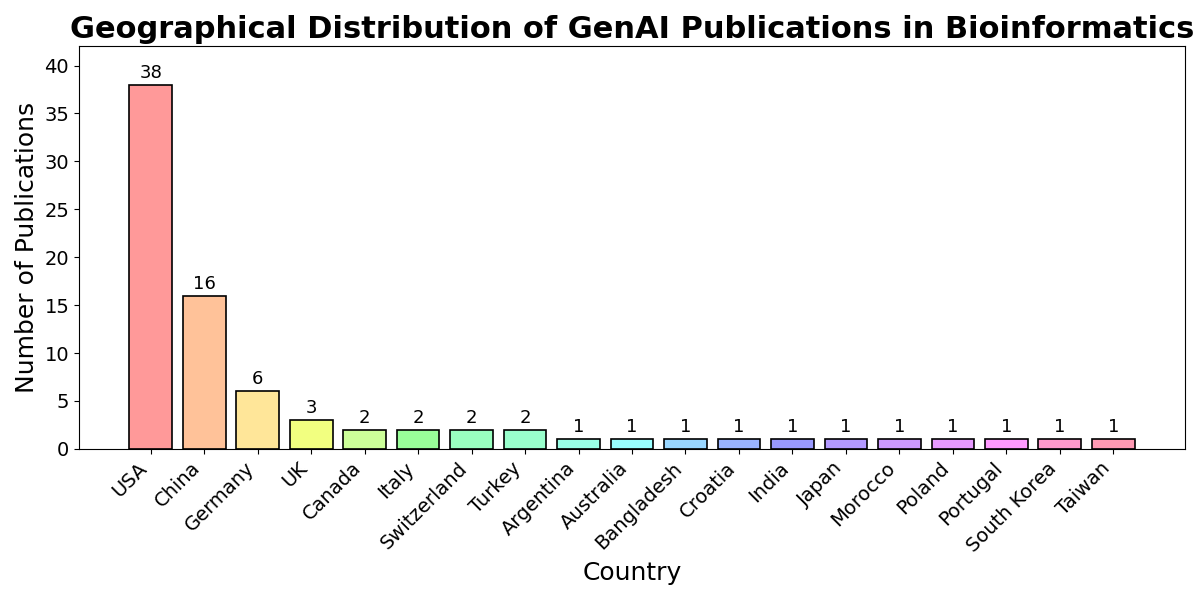}
\caption{Geographic distribution of reviewed publications, illustrating the global research contributions from major countries and regions in the field of GenAI for bioinformatics.}
\label{fig:genai_dis}
\end{figure}

 \x{A total of 3,312 records were retrieved from Google Scholar and publisher platforms, including ScienceDirect (n=543), Oxford Academic (n=684), arXiv (n=733), NeurIPS (n=382), Springer Nature (n=554), and BMC (n=416). Based on the predefined criteria, a total of 82 research articles are selected, meeting all inclusion requirements and maintaining the defined exclusion parameters. Table \ref{tab:included_studies} provides the full set of articles selected for inclusion in this review.} Among the collected studies, \x{63} were published in Q1 journals, \x{6} appeared in A* conference proceedings, and \x{13} were identified as preprints, covering publications from 2021 to \x{2026}. The highest number of journals \x{has} been collected in 2024, followed in sequence by 2025, 2023, 2022, \x{and early-2026}, whereas 2021 has the lowest count. The classification of the journals in this review was determined using the SCImago Journal Rank (SJR) database. Figure \ref{fig: paper} shows the total number of studies, along with the distribution of the publications for the specified year ranges. These selected studies comprehensively reflect the applications of GenAI and LLMs in bioinformatics. Moreover, Figure \ref{fig:genai_dis} presents the geographical distribution of publications, indicating that the United States and China contribute the majority of the publications.
 
\section{Key Insights from the Research Questions}\label{RQs}
This section presents an extensive overview of key findings of six RQs, addressing the applications of GenAI models, the corresponding performance of general-purpose and domain-specific architectures, and benefits across protein, genomics, and multi-omics research. The overview also covers the task-specific improvements, such as protein structure prediction and data integration, summarizes the datasets used for training and grounding the models, and discusses the current limitations, challenges, and future directions for advancing the field.

\subsection{Applications of GenAI in Bioinformatics (RQ1)}
\x{GenAI models have introduced powerful and versatile capabilities} and enabled novel applications such as \textit{de novo} protein design \cite{korendovych2020novo}, sequence generation, and automatic functional annotation. These developments represent significant advances over rule-based or statistical methods through the learning of complex patterns in bioinformatics and the generation of biologically meaningful outputs. 

\subsubsection{Genomic Sequence Modeling and Regulatory Element Prediction}
GenAI models have transformed the framework of genomic sequence modeling to enable advanced interpretation and prediction of regulatory elements from DNA sequences directly. Among the most referenced and notable studies in this direction is DNABERT \cite{ji2021dnabert}, a BERT-derived Transformer model that treated genomic sequences as a language by k-mer tokenization. DNABERT enabled high-accuracy predictions on a variety of tasks, such as promoter prediction, transcription factor binding site prediction, and splice site prediction. It outperformed traditional Convolutional Neural Network (CNN) and Recurrent Neural Network (RNN) models by a wide margin through encoding sequential dependencies without requiring large labeled datasets, thereby generalizing to a variety of bioinformatics tasks. Similarly, DNABERT-2 \cite{zhou2024dnabert} enables sequence-only prediction of colorectal cancer enhancers by learning contextual DNA representations through Byte Pair Encoding-based genomic language modeling. Extending this thread of research, another work introduced GROVER \cite{sanabria2024dna}, a foundational language model pretrained on the human genome through byte pair encoding. GROVER performed better than DNABERT-2 \cite{zhou2024dnabert} and other comparable models in genome element recognition tasks and protein-DNA interaction prediction. The more advanced vocabulary encoding enabled finer-grained sequence understanding, validating the impact of architectural innovations on genomic modeling. Pushing generalizability even further, Nucleotide Transformers \cite{dalla2025nucleotide} were trained on vast, unlabeled DNA datasets to produce general-purpose embeddings. These were fine-tuned for molecular attribute prediction and variant ranking. Compared to traditional supervised models such as BPNet, these models achieved competitive or enhanced accuracy at greatly reduced training costs, highlighting their efficiency and adaptability across genomic applications.

\x{While DNABERT \cite{ji2021dnabert} and GROVER \cite{sanabria2024dna} demonstrate the effectiveness of transformer-based pretraining for genomic sequence analysis, they primarily focus on learning sequence representations rather than modeling chromatin architecture and long-range regulatory interactions. Their limited sequence context restricts the capture of regulatory relationships involving distal enhancers, chromatin loops, and topologically associating domains (TADs). In contrast, Enformer \cite{avsec2021effective} explicitly models long-range genomic dependencies, improving gene expression and noncoding variant effect prediction, while Sei \cite{chen2022sequence} provides a comprehensive framework for mapping sequences to regulatory activities through large-scale chromatin profile prediction. These studies highlight that effective chromatin state modeling requires integrating long-range regulatory information beyond sequence-level representation learning, an area that remains underexplored in current genomic language models.}

Meanwhile, TAPEBert \cite{zhang2022t4sefinder}, a protein language model designed for protein tasks, was reconfigured for genome-scale prediction, such as classifying Type-IV Secreted Effectors. The improved performance and efficiency over position-specific scoring matrix–based methods show that shared language modeling architectures enhance the connection between genomic and proteomic representations. GP-GPT \cite{lyu2024gp} enabled advanced mapping of gene-trait relationships and biomedical question answering. By embedding genome-wide associations within a generative framework, this approach exceeded rule-based systems by providing flexible reasoning and a broader understanding of biological context. Another application, LLM2geneset \cite{zhu2025enhancing}, used LLMs to create gene set databases based on specific experimental conditions. In contrast to traditional Over-Representation Analysis (ORA) software with fixed gene sets, LLM2geneset added domain relevance and task relevance, extending the usefulness of refinement analyses in genomics. Breaking new ground beyond prediction, Precious2GPT (P2GPT) \cite{sidorenko2024precious2gpt} introduced generative modeling for artificially generated omics data generation, which helped benchmark analysis pipelines, affect differential expression, and augment small datasets. Compared to classical deep learning methods such as conditional GANs or component-wise methods such as CDiffusion and MoPT, P2GPT captured biological authenticity more effectively by learning conditional and complex relationships in omics data.

To support advanced analytical reasoning in genomics, BioAgents \cite{mehandru2025bioagents} employed multiple small language models to provide on-demand code generation, workflow specification, and biological interpretation. The multi-agent assessment system improved reproducibility and usability of computational genomics, showing how generative models can be employed as intelligent assistants for real-time bioinformatics problem solving. The use of LLMs has also extended to single-cell genomics, with mLLMCelltype \cite{yang2026large} being a multi-LLM consensus model for annotating cell types in single cell RNA (scRNA)-seq data. By integrating marker gene expression and tissue-specific context, the approach improves annotation accuracy by nearly 15\% over standard approaches. It also introduced a model uncertainty evaluation and reasoning transparency, an essential feature of sensitive biomedical analysis. While Bingo \cite{ma2024bingo} demonstrated a hybrid approach that combined the ESM-2 protein language model with graph neural networks (GNN) to predict essential genes in organisms from sequence alone. It represented an example of how structural and domain-specific information extracted from generative models can replace time-consuming experimental annotations in functional genomics. Moving a step further, PLM-interact \cite{liu2025plm} extends the ESM-2 from single-protein representation to paired-protein interaction learning. It enables protein–protein interaction (PPI) prediction, mutation-effect prediction on interactions, and virus–human PPI prediction from sequence alone. Compared with traditional frozen-embedding classifiers, PLM-interact jointly encodes protein pairs and shows stronger cross-species PPI performance.

\begin{figure*}[ht!]
\centering
\includegraphics[scale=0.09]{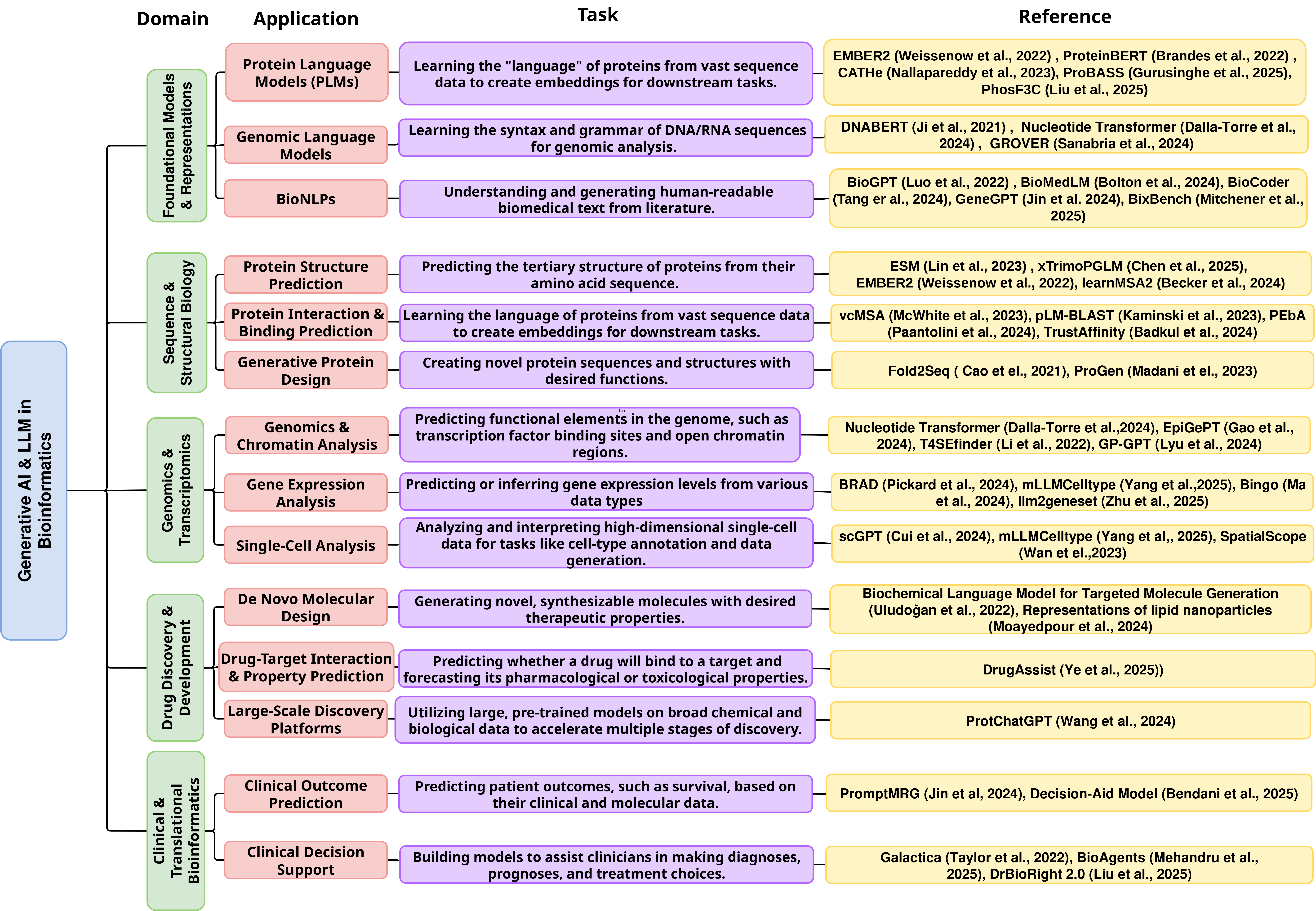}
\caption{Hierarchical taxonomy of the classification of GenAI applications within the bioinformatics fields of genomics, proteomics, transcriptomics, and drug discovery based on model type and methodological approach.}
\label{fig:taxonomy}
\end{figure*}

In general, GenAI has significantly advanced genomic sequence modeling and regulatory element prediction by enabling context-aware learning from unlabeled genomic data, enabling dynamic and task-specific gene set generation, improving interpretability in analyses, and producing biologically realistic synthetic data for benchmarking and enhancement. Figure \ref{fig:taxonomy} presents the taxonomy of recent GenAI models in bioinformatics, organized by application domains and \x{highlighting representative model architectures for each domain.}

\subsubsection{Protein Representation, Functional Annotation, and Structure Prediction}
The capabilities of protein models ESM-1v, multiple sequence alignment (MSA) Transformer, and ProtT5 have greatly improved with zero-shot prediction, which no longer requires MSAs or specific training tasks \cite{meier2021language}. This is a core component of protein representation learning. These models provide a generalized approach across entire families of proteins, which allows for unsupervised assessment of mutations in proteins and is useful for understanding diseases as well as for protein engineering. To begin with, EMBER2 \cite{weissenow2022protein} extended this by using ProtT5 embeddings to predict distances between \x{residues} and even 3D structures, \x{enabling MSA-free structure prediction and faster protein modeling pipelines}. Furthermore, ESMFold, which eliminated the need for MSAs \cite{lin2023evolutionary}, allows a prediction 60 times faster than AlphaFold2 \cite{skolnick2021alphafold}. Large-scale structural interpretation derived from metagenomic sequences (i.e., 617M proteins) can now be performed within reasonable time frames for the first time.

Extending protein function modeling to the next level, PhosF3C \cite{liu2025phosf3c} illustrated its ability to effectively predict PTM sites, such as phosphorylation, methylation, and crotonylation, by employing layer-wise relevance propagation to provide a high degree of explainability. Tools such as pLM-BLAST \cite{kaminski2023plm} and transformer-based soft alignment methods enable remote homology detection in an unsupervised manner, specifically between low-similarity proteins, by replacing the traditional substitution matrices with high-dimensional embedding similarities, profoundly impacting homology detection and evolutionary analysis \cite{harrigan2024improvements}. GenAI is also transforming protein–protein and protein-ligand interaction prediction. \x{Additionally}, TrustAffinity \cite{badkul2024trustaffinity} is another model that uses ESMFold-derived embeddings and graph representations to predict more accurate protein-ligand affinities than conventional docking tools, especially in out-of-distribution settings, and therefore is an important drug discovery tool. From an interactive perspective, applications such as ProtChatGPT \cite{wang2024protchatgpt} and DrBioRight 2.0 \cite{liu2025drbioright} brought natural language interfaces to protein bioinformatics. ProtChatGPT supports conversational protein Q\&A, design analysis, and drug target evaluation. Meanwhile, DrBioRight 2.0 largely supports protein-centered cancer analysis tasks such as survival modeling and pathway correlation that were previously dependent on domain-expert interactions. Similarly, Prot2Chat \cite{wang2025prot2chat} integrates protein sequence and structural embeddings with LLMs to enable natural-language protein question answering, structure-aware sequence design, and richer functional annotation that produces human readable, context-aware explanations that outperform traditional sequence-only approaches.

ProtGPT2 \cite{ferruz2022protgpt2}, ProGen \cite{madani2020progen}, and ProGen2 \cite{nijkamp2023progen2} are a few models that have already shown capacity for designing new, diverse protein sequences with natural folds and even enzyme activity. While Madani et al. \cite{madani2023large} demonstrated that lysozyme designed by ProGen maintained activity, outperforming traditional coevolutionary methods such as bmDCA, whose designs lacked measurable function. In another study, researchers went further by using ProGen2 \cite{nijkamp2023progen2} for the exploration of protein fitness spaces and modeling feasible sequences without additional fine-tuning, outside the limitations of evolutionary sequence reconstruction and sparse MSA-based approaches. XTrimoPGLM \cite{chen2024xtrimopglm} and its structural submodule xT-Fold also include the trend towards unified, multimodal protein models through combining understanding and generation tasks in one pretraining framework. Another study, ProteinBERT \cite{brandes2022proteinbert}, combines language modeling with Gene Ontology prediction to enable efficient protein function prediction and achieves strong performance across multiple protein analysis tasks, including structure prediction, post-translational modification detection, and biophysical property estimation. These models provide SOTA structure and function prediction, while enabling controlled and tunable sequence generation. DeepUrfold \cite{draizen2024deep} employs deep generative modeling to explore protein fold space, revealing hidden relationships that traditional clustering methods often fail to capture. Taken together, these studies demonstrate that protein language models have evolved from sequence representation tools to versatile frameworks for protein function prediction, structural analysis, and controllable protein design, expanding the capabilities of bioinformatics beyond traditional feature-engineered and alignment-based approaches.

\subsubsection{RNA Foundation Models for Structure, Function, Interaction, and Design}
\x{GenAI has shifted Ribonucleic Acid (RNA) bioinformatics from developing separate models for individual tasks to building foundation models that learn general RNA representations from large-scale sequence data. This transition is well demonstrated by the models RNA-FM \cite{chen2022interpretable}, RiNALMo \cite{penic2025rinalmo}, and ERNIE-RNA \cite{yin2025ernie}. Unlike traditional approaches that depend on thermodynamic assumptions, dynamic programming, handcrafted features, or sequence alignment, these models are pretrained on millions of unlabeled RNA sequences and learn structural and functional patterns directly from data. As a result, they can be adapted to a wide range of downstream tasks, including secondary structure prediction, non-coding RNA (ncRNA) classification, splice-site prediction, RNA contact prediction, 3D structure-related analysis, and expression prediction. Compared with classical methods such as RNAfold, RNAstructure, Mfold, and CONTRAfold, as well as task-specific deep learning models named SPOT-RNA, UFold, MXfold2, and UTR-LM, GenAI RNA foundation models generally show better generalization across diverse RNA families and biological settings. Their main contribution, however, extends beyond improvements in predictive performance. By providing a shared representation that can be transferred across multiple tasks, these models reduce the need for developing separate specialized systems and establish a more unified framework for RNA analysis.}

\x{Another important application area is splicing and regulatory sequence modeling, where generative AI supports both prediction and sequence design. SpliceBERT \cite{chen2024self} demonstrates the value of self-supervised pretraining on large collections of vertebrate RNA sequences by providing a unified framework for branchpoint prediction, splice-site prediction, and zero-shot assessment of variant effects on splicing. Compared with conventional splicing models such as HAL, SPANR, MMSplice, SpliceAI, Pangolin, Branchpointer, and LaBranchoR, SpliceBERT benefits from transferable sequence representations that improve performance across multiple splicing-related tasks. Moving beyond prediction, TrASPr+BOS \cite{wu2026generative} combines transformer-based alternative splicing prediction with Bayesian optimization to generate RNA sequences that produce desired tissue-specific splicing patterns. Further REDAC \cite{de2026redac} applies GenAI in RNA-seq by using Gemma and Large Language Model Meta AI (LLaMA) to let users perform Differential Expression Analysis (DEA), generate plots, produce R programming language code, and interpret pathway enrichment results through natural language queries. Compared with traditional Graphical User Interface (GUI) tools and unconstrained LLM code generation, REDAC improves usability and reproducibility by using predefined Empirical Analysis of Digital Gene Expression Data in R (edgeR) workflows, JavaScript Object Notation (JSON)-constrained outputs, Retrieval-Augmented Generation (RAG) based on PubMed, and automated reports, reducing hallucination-related failures.}

\x{Generative AI has also extended RNA bioinformatics beyond sequence analysis to interaction prediction, controllable RNA design, and knowledge-driven biological interpretation.  BioLLMNet \cite{abir2025biollmnet}, and CrossLLM-Mamba \cite{sadia2026crossllm} combine pretrained language models for RNA, proteins, and small molecules to predict RNA--protein, RNA--RNA, and RNA--small molecule interactions. By learning representations directly from biological sequences and molecular data, these frameworks reduce the reliance on handcrafted structural and physicochemical features that are commonly used in traditional interaction prediction methods. Moving a step further, RNA-DCGen \cite{shahgir2024rna} enables controlled RNA sequence generation by designing sequences that satisfy predefined structural constraints, including secondary structures and distance maps, while maintaining conserved regions. In RNA structure modeling, RhoFold+ \cite{shen2024accurate} integrates RNA-FM representations with geometric learning strategies to improve \textit{de novo} RNA 3D structure prediction, outperforming several template-based, sampling-based, and deep learning approaches. Beyond predictive and generative tasks, GenAI is increasingly being applied to biological knowledge discovery and interpretation. RNA-GPT \cite{xiao2024rna}, BIOGEN \cite{hossain2025interpretable}, GL4SDA \cite{la2025gl4sda}, and LitSumm \cite{green2025litsumm} support applications such as RNA-focused question answering, transcriptomic interpretation with biological evidence, snoRNA disease association prediction, and automated ncRNA literature summarization.}


\subsubsection{Drug Discovery and Molecule Generation}
Over the past few years, GenAI has been a significant breakthrough in molecular design and drug discovery, greatly influencing how new therapeutic molecules are created, optimized, and experimentally validated. Traditional approaches are heavily dependent on structural binding simulations, molecular fingerprints, or supervised learning within the constraints of available interaction datasets. In contrast, generative models, particularly those based on LLMs and PLMs, hold the promise of \textit{de novo} molecule creation and optimization based on learned biochemical and contextual patterns, even in the absence of existing target molecule interaction data.

One promising direction frames molecule generation as a translation task in which biochemical language models map protein sequences to the corresponding chemical structures. This approach enables the design of novel target-specific molecules for proteins that lack known ligands, which is infeasible with traditional docking or supervised methods \cite{uludougan2022exploiting}. By learning latent representations that link molecular structures with amino acid sequences, these models make it possible to scale \textit{de novo} drug discovery from raw biological input. Enhancing this capability is DrugAssist \cite{ye2025drugassist}, an LLM-based system that supports conversational molecular optimization. It supports both single-property and multi-property optimization under realistic range-based limitations, which better reflect the balances required in pharmaceutical development. Another unique feature of DrugAssist is that it supports iterative refinement through expert feedback, allowing domain experts to interactively guide molecular generation and optimization, breaking from the static, black box behavior of traditional AI pipelines. 

GenAI is also making drug discovery more interactive with systems such as ProtChatGPT \cite{wang2024protchatgpt}, which bridges the gap between protein analysis and drug target prediction with a natural language interconnection. ProtChatGPT allows users to query protein properties, assess binding potential, and explore design strategies in an interactive and accessible format. The multi-level protein–language alignment model of ProtChatGPT provides protein understanding and drug discovery insights without requiring scripting or extensive domain expertise. Beyond small molecules, GenAI is also gaining momentum in the modeling of drug delivery systems. A prominent example is the use of fine-tuned MegaMolBART \cite{moayedpour2024representations} embeddings to predict lipid nanoparticle (LNP) gene delivery performance, a key parameter in messenger RNA (mRNA)-based therapeutics. These LLM-derived molecular representations outperform traditional graph or fingerprint-based methods by capturing more comprehensive structural and chemical relationships, providing an effective framework for optimizing next-generation biopharmaceuticals. Respectively, Top-DTI \cite{talo2025top} uses LLM embeddings together with topological feature vectors to perform drug–target interaction (DTI) prediction, improving on ligand-based similarity and classic ML feature-vector approaches by learning complementary sequence and topological information and dynamically fusing them for enhancing predictive performance and robustness. The outcome is a generative model based pipeline that advances practical tasks in drug discovery, where traditional methods either require expensive 3D binding or may fail when known ligands are limited.

GenAI is transforming the drug discovery process from search-heavy, rigid, algorithm-based approaches to more fluid and systematic ones through the integration of molecule generation. By integrating a unified system for optimization, interaction prediction, and delivery modeling, these innovations expand the scope of computational capabilities in the development of therapeutics.

\subsubsection{Interactive Single-Cell Analysis}
These latest advancements in GenAI have significantly \x{advanced} single-cell analysis by \x{offering effective solutions to challenges} such as high dimensionality to batch effects and integration of omics modalities. By prelearning latent biological representations over a range of modalities, scGPT \cite{cui2024scgpt} enables transfer learning on a range of tasks such as cell type annotation, modification response prediction, and gene regulatory network (GRN) inference. Compared to traditional task-specific models, which require task-by-task independent pipelines, scGPT enables a unified framework based on shared biological structure, with improved performance and generalizability across datasets. To integrate with this, mLLMCelltype \cite{yang2026large} introduces a multi-LLM consensus method for cell type annotation of scRNA-seq data, a task involving domain expertise and manual curation. Importantly, mLLMCelltype also introduced uncertainty quantification and opened reasoning chains, elements that are commonly lacking from black-box annotation tools. While models such as scGPT aim for general-purpose modeling, frameworks such as OLAF \cite{riffle2025olaf} and BRAD \cite{pickard2024language} mark the growing importance of agentic and interactive GenAI in single-cell workflows. OLAF enables natural language-based execution of complete data analysis pipelines, including clustering, marker gene detection, and visualization, bridging LLM reasoning with live code execution. Similarly, BRAD provides a mechanism for the discovery of biomarkers using LLMs via linking to third-party software and databases, and allows LLMs to be applied in dynamic workflows, not only static question answering.

In cell pathway inference, PyEvoCell \cite{mathur2025pyevocell} \x{combines LLMs with dynamic biological modeling to interpret single-cell progression, proposing biologically feasible transitions and annotating downstream differential gene expression (DGE) and gene set enrichment analysis (GSEA) results.} This improved interpretability offers a new interface to pathway analysis, which has previously depended on deterministic or heuristic methods. GenAI also accelerates the advancement of spatial transcriptomics (ST). SpatialScope \cite{wan2023integrating} advances the field by enhancing low-resolution sequencing-based data, including Visium, to pseudo single-cell resolution and imputing complete transcriptomes from high-resolution, image-based ST data. The model outperforms traditional deconvolution methods such as RCTD and Cell2location, as well as alignment-based approaches including Tangram and CytoSPACE, by predicting transcriptome-wide pseudo-cell profiles rather than merely assigning cells or estimating proportions. In particular, it improves dropout imputation and enables fine-grained spatial mapping in complex tissues.

\begin{figure*}[ht!]
\centering
\includegraphics[scale=0.57]{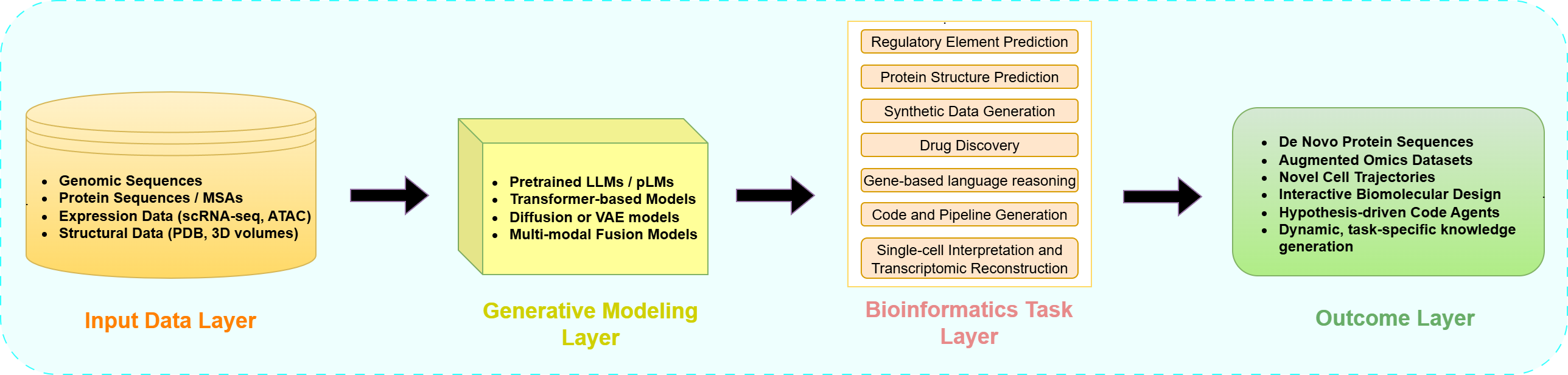}
\caption{End-to-end workflow demonstrating the integration of GenAI in bioinformatics, from data collection and preprocessing to model development, training, and evaluation.}
\label{fig:rq1}
\end{figure*}

Multi-omics integration has been another critical challenge in single-cell analysis. Generative models such as sciCAN \cite{xu2022scican}, which employs a combination of variational autoencoders (VAEs) and generative adversarial networks (GANs), facilitate unsupervised integration of scRNA-seq and single-cell Assay for Transposase-Accessible Chromatin (scATAC)-seq through learning a shared latent space that aligns the modalities while preserving biological variability. As compared to five standard integration methods, sciCAN performs better in modality transfer and batch removal, particularly in registering heterogeneous omics layers. Similarly, scDREAMER \cite{shree2023scdreamer} employs deep generative modeling for the integration of heterogeneous scRNA-seq datasets to generate batch-invariant embeddings that can preserve cellular diversity and biological signals between datasets. Compared with existing batch correction methods that focus on classical batches, scDREAMER is more robust to imbalanced cell type distributions and higher-order nested batch effects, thus allowing for more reliable comparisons in cell atlas-scale applications. These techniques collectively demonstrate how GenAI is transforming the single-cell analysis workflow, transitioning from static rules-based pipelines to dynamic, explainable, multimodal systems that can keep up with biological complexity.

\begin{table*}[ht!]
\centering
\renewcommand{\arraystretch}{1.3} 
\setlength{\tabcolsep}{6pt}       
\scriptsize
\caption{Comprehensive overview of recent GenAI models in bioinformatics, categorized by application domains such as genomic sequence modeling, protein representation and design, RNA sequence modeling, drug discovery, and single-cell analysis. The table summarizes key models along with their comparative baselines and core methodological innovations.}
\label{tab:genai_bioinfo}
\begin{tabular*}{\textwidth}{@{\hspace{2pt}}l@{\hspace{6pt}} l@{\hspace{6pt}} c@{\hspace{6pt}} l@{\hspace{6pt}} p{6.8cm}}
\hline
\textbf{Type} & \textbf{Method} & \textbf{Year} & \textbf{Baseline} & \textbf{Key Technique} \\
\hline

\multirow{8}{*}{Genomic Sequence Modeling} 
& DNABERT \cite{ji2021dnabert} & 2021 & CNN, RNN & k-mer tokenized BERT \\
& TAPEBert \cite{zhang2022t4sefinder} & 2022 & PSSM methods & Genome-scale protein LM repurposing \\
& DNABERT-2 \cite{zhou2024dnabert} & 2024 & DNABERT & Enhanced transformer variant \\
& GROVER \cite{sanabria2024dna} & 2024 & DNABERT-2 & Byte-pair encoded genome model \\
& P2GPT \cite{sidorenko2024precious2gpt} & 2024 & GANs, CDiffusion & Synthetic omics data generation \\
& GP-GPT \cite{lyu2024gp} & 2024 & Rule-based methods & Genome–phenotype LLM reasoning \\
& Nucleotide Transformers \cite{dalla2025nucleotide} & 2025 & BPNet & Foundation embeddings from unlabeled DNA \\
& LLM2geneset \cite{zhu2025enhancing} & 2025 & ORA tools & Context-specific gene set generation \\
\hline

\multirow{10}{*}{Protein Modeling} 
& ESM-1v, ProtT5 \cite{meier2021language} & 2021 & MSA models & Zero-shot sequence understanding \\
& EMBER2 \cite{weissenow2022protein} & 2022 & MSAs & Structure prediction from ProtT5 embeddings \\
& ESMFold \cite{lin2023evolutionary} & 2023 & AlphaFold2 & Fast MSA-free folding model \\
& pLM-BLAST \cite{kaminski2023plm} & 2023 & BLAST, HMMER & PLM similarity for remote homology \\
& TrustAffinity \cite{badkul2024trustaffinity} & 2024 & Docking tools & GIN with ESMFold based affinity scoring \\
& ProtChatGPT \cite{wang2024protchatgpt} & 2024 & Manual Q\&A & Conversational protein insights \\
& PhosF3C \cite{liu2025phosf3c} & 2025 & PTM tools & Layer-wise relevance for PTM sites \\
& DrBioRight 2.0 \cite{liu2025drbioright} & 2025 & Manual analysis & Protein-based cancer model generation \\
\hline

\multirow{4}{*}{Protein Design} 
& ProtGPT2 \cite{ferruz2022protgpt2} & 2022 & bmDCA & \textit{De novo} protein generation \\
& ProGen2 \cite{nijkamp2023progen2} & 2023 & ProGen & Fitness-guided protein generation \\
& Madani et al. \cite{madani2023large} & 2023 & bmDCA & Activity-validated protein sequences \\
& xTrimoPGLM \cite{chen2024xtrimopglm} & 2024 & Uni-modal models & Multimodal pretraining for structure \& sequence \\
\hline

\multirow{10}{*}{\x{RNA-Specific Modeling}}
& \x{RNA-FM \cite{chen2022interpretable}} 
& \x{2022} 
& \x{RNAfold, Mfold} 
& \x{Self-supervised ncRNA foundation model} \\
& \x{SpliceBERT \cite{chen2024self}} 
& \x{2024} 
& \x{SpliceAI, MMSplice} 
& \x{Multi-species MLM for splice-site modeling} \\
& \x{RhoFold+ \cite{shen2024accurate}} 
& \x{2024} 
& \x{FARFAR2, AlphaFold3} 
& \x{RNA-FM-guided full-atom 3D structure prediction} \\
& \x{RNA-GPT \cite{xiao2024rna}} 
& \x{2024} 
& \x{RNA-FM, RiNALMo} 
& \x{Multimodal conversational RNA analysis} \\
& \x{RNA-DCGen \cite{shahgir2024rna}} 
& \x{2024} 
& \x{gRNAde, RNA-FrameFlow} 
& \x{Gradient-guided conditional RNA sequence design} \\
& \x{RiNALMo \cite{penic2025rinalmo}} 
& \x{2025} 
& \x{SPOT-RNA, UFold} 
& \x{Large-scale ncRNA language model} \\
& \x{ERNIE-RNA \cite{yin2025ernie}} 
& \x{2025} 
& \x{RNA-FM, RiNALMo} 
& \x{Zero-shot structure prediction} \\
& \x{BioLLMNet \cite{abir2025biollmnet}} 
& \x{2025} 
& \x{RPISeq, RSAPred} 
& \x{Gated BioLLM fusion for RNA interaction prediction} \\
& \x{CrossLLM-Mamba \cite{sadia2026crossllm}} 
& \x{2026} 
& \x{BioLLMNet, IntaRNA} 
& \x{Bidirectional Mamba for RNA interaction crosstalk} \\
& \x{REDAC~\cite{de2026redac}}
& \x{2026} 
& \x{Gemma and LLaMA} 
& \x{Dual LLMs, automated report generation} \\
\hline

\multirow{4}{*}{Drug Discovery} 
& Protein-to-Drug Generator \cite{uludougan2022exploiting} & 2022 & Docking tools & Protein-to-molecule translation \\
& ProtChatGPT \cite{wang2024protchatgpt} & 2024 & Scripted tools & Q\&A for drug–protein design \\
& MegaMolBART \cite{moayedpour2024representations} & 2024 & GNNs, fingerprints & LNP delivery prediction from embeddings \\
& DrugAssist \cite{ye2025drugassist} & 2025 & Optim. tools & Interactive molecule optimization via LLM \\
\hline

\multirow{5}{*}{Single-Cell Analysis} 
& scGPT \cite{cui2024scgpt} & 2024 & Task-specific pipelines & Transformer for multi-omics learning \\
& BRAD \cite{pickard2024language} & 2024 & Manual biomarker tools & Agentic biomarker discovery \\
& mLLMCelltype \cite{yang2026large} & 2026 & Manual annotation & Multi-LLM consensus cell typing \\
& OLAF \cite{riffle2025olaf} & 2025 & Coding tools & Natural language pipeline execution \\
& PyEvoCell \cite{mathur2025pyevocell} & 2025 & PAGA, Monocle & Pathway prediction with LLM explanations \\
\hline

\multirow{2}{*}{Multi-omics Integration} 
& sciCAN \cite{xu2022scican} & 2022 & Seurat, Harmony & VAE and GAN joint latent space for scRNA \& scATAC \\
& scDREAMER \cite{shree2023scdreamer} & 2023 & Batch correction tools & Batch-invariant deep embedding generation \\
\hline
\end{tabular*}
\end{table*}

GenAI has progressed from specialized applications to multipurpose toolkits in genomics, protein modeling, drug discovery, and single-cell analysis.  These models outperform traditional statistical, rule-based, and narrowly scoped deep learning methods by capturing contextual, high-dimensional biological patterns that generalize to new, unseen tasks and generate biologically meaningful outputs. Cross-domain applications, such as protein language models (PLMs) for genomics, and the creation of interactive, multimodal systems, signal the emergence of integrated, explainable, and workflow-aware AI in bioinformatics for accelerated discovery and translation.

Table~\ref{tab:genai_bioinfo} provides a comprehensive overview of recent GenAI models in bioinformatics, categorized by application domains such as genomic sequence modeling, protein representation and design, drug discovery, and single-cell analysis. It summarizes key models, their comparative baselines, and the core methodological innovations that define current trends in GenAI for computational biology. Figure \ref{fig:rq1} \x{illustrates the end-to-end workflow of applying GenAI in bioinformatics, from data collection and preprocessing to model training, evaluation, and application-specific outputs.}

\subsection{Effective Models for Biological Sequences (RQ2)}
Recent advances in GenAI highlight the need to identify the models and architectures that are most effective for bioinformatics applications. While researchers use increasingly general-purpose LLMs, as well as domain-specific protein, DNA, or RNA language models. \x{This section examines these factors in depth and illustrates how domain-specialized models tend to outperform general models on certain task categories, though this advantage varies considerably across tasks and evaluation settings.} 

\subsubsection{The Edge of Specialization: Domain-Specific Models in Bioinformatics}
The increasing complexity and specificity of biological information have encouraged the appearance and application of domain-specific language models, which consistently outperform general-purpose LLMs on various sets of bioinformatics tasks. In contrast to general LLMs trained on diverse textual datasets, domain-specific models are specialized to learn the contextual meaning and patterns embedded in biological sequences, molecular interactions, and cellular networks. Applying k-mer based tokenization and biological sequence semantics, DNABERT \cite{ji2021dnabert} achieved SOTA promoter and transcription factor binding site (TFBS) prediction performance, by a wide margin outperforming general architectures such as DeepSEA, DanQ, and DeePromoter \x{by a wide margin, as these models lack domain-specific pretraining}. It reported mean and median accuracy values of 91.8\% and 91.9\%, respectively, on TFBS prediction tasks. In the proteome as well, ESM-1v and ESM-2, the large masked PLMs outperformed general LLMs such as UniRep and ProtBERT-BFD in zero-shot prediction of mutational effects \cite{meier2021language}. A single model achieved a Spearman correlation of approximately 0.48 on the full set and 0.46 on the test set, while an ensemble of five models improved this to around 0.51 and 0.48, respectively. Fine-tuning through augmented MSA further increased the correlation to about 0.54, placing these models on par with supervised predictors, particularly in low-identity and low-resource settings. Domain-specific architectures excel in sequence modeling as well as in structural and functional prediction. In Pfam family classification tasks, ProtT5-XL-U50 and ESM-2, when combined with ensemble CNN or similar transfer learning classifiers, achieve the lowest error rates of 7.28\% for ProtT5-XL-U50 and CNN-E, 7.67\% for ESM-2 and CNN-E, thus exceeding traditional supervised DL models without pretrained embeddings \cite{vitale2024evaluating}.

Several studies highlight the benefits of fine-tuning backbones with domain-specific information. TrustAffinity~\cite{badkul2024trustaffinity}, which integrates structure-informed ESMFold with a GNN-based encoder for the ligand, demonstrated strong predictive performance with a Root Mean Square Error (RMSE) of 0.384, Mean Absolute Error of 0.312, Pearson’s $r$ of 0.856, and Spearman’s $\rho$ of 0.820, thereby significantly outperforming general LLMs as well as conventional deep learning baselines for protein--ligand binding affinity prediction. Similarly, MegaMolBART \cite{moayedpour2024representations}, a fine-tuned biochemical LLM, achieved the highest \x{Area Under the Curve} (AUC) of 0.981 for lipid nanoparticle classification, surpassing graph-based models such as GROVER and hand-engineered molecular descriptors. Collectively, these findings underscore the advantage of task-aware fine-tuning and domain-specific embeddings in enabling the acquisition of in-depth molecular properties. Prot2Chat \cite{wang2025prot2chat} highlights the advantage of domain specialization by integrating protein structure and sequence embeddings with a fine-tuned LLaMA3 backbone, achieving Bilingual Evaluation Understudy (BLEU)-2 and Recall-Oriented Understudy for Gisting Evaluation (ROUGE)-L scores of 35.85 and 50.51 on the Mol-Instructions dataset and surpassing larger general-purpose models that confirms biologically aligned multimodal fusion is more effective than scaling model size alone. Similarly, fine-tuned PLM-interact \cite{liu2025plm} showed usefulness for cross-species PPI prediction, mutation-aware interaction analysis, and virus–host interaction prediction, showing value for disease biology, variant-effect interpretation, and host–pathogen studies. For mutation-effect prediction, PLM-interact reached Area Under the Precision-Recall (AUPR) 0.612 and Area Under the Receiver Operating Characteristic Curve (AUROC) 0.794, while zero-shot models were near random.

In single-cell analysis, models such as scGPT \cite{cui2024scgpt}, deployed with a masked multihead attention mechanism customized for the non-sequential nature of gene expression data, have been shown to generalize across tasks including cell type annotation, multi-omic integration, and treatment response prediction. In its original study, scGPT was pretrained on over 33 million cells and demonstrated strong downstream performance, achieving nontrivial correlation metrics in biological response benchmarks when fine-tuned. In comparison, transformer-based models such as scBERT \cite{yang2022scbert}, pretrained on large-scale scRNA-seq corpora, achieved annotation accuracies of approximately 0.84 with favorable macro F1 scores, while GEARS \cite{roohani2022gears}, designed to predict transcriptional responses to combined gene modifications, achieved roughly 40\% higher precision in classifying genetic interaction subtypes and doubled the detection rate of the strongest interactions relative to prior methods. Nevertheless, in zero-shot reconstruction settings, \x{scGPT does not consistently exceed trivial baseline predictors} \cite{kedzierska2025zero}.

Autoregressive Transformer-based models, such as ProGen \cite{madani2023large} and ProGen2 \cite{nijkamp2023progen2}, have advanced domain specialization in the design of generative protein sequences. Scaled up to 6.4B parameters and trained on more than 1 \x{billion} protein sequences spanning genomic, metagenomic, and immune repertoires, these models demonstrated SOTA performance in modeling evolutionary sequence distributions, generating novel viable sequences, and predicting protein fitness without task-specific fine-tuning. Similarly, BioGPT \cite{luo2022biogpt}, pretrained on large-scale biomedical literature, reported F1 scores of 44.98\% on BioCreative V Chemical–Disease Relation dataset (BC5CDR) \cite{wei2016assessing}, 38.42\% on Knowledge Discovery Drug–Target Interaction dataset (KD-DTI) \cite{hou2022discovering}, and 40.76\% on Drug–Drug Interaction (DDI) \cite{herrero2013ddi} relation extraction, along with 78.2\% accuracy on PubMedQA\cite{jin2019pubmedqa}, thereby surpassing earlier biomedical Natural Language Processing (NLP) models in most tasks \cite{luo2022biogpt}. BioMedLM, comprising 2.7B parameters and trained solely on PubMed abstracts and full texts, achieved 57.3\% accuracy on MedMCQA and 69.0\% on the Medical Genetics section of the Massive Multitask Language Understanding (MMLU) benchmark, thereby \x{matching or surpassing substantially larger general-purpose LLMs, even when those models were fine-tuned} \cite{bolton2024biomedlm}. In immunotherapy biomarker prediction, a Random Forest ensemble integrating histological and clinical features attained an accuracy of 82\% and an AUC of 0.91 on the validation set, demonstrating strong discriminative ability in classifying tumor mutation burden levels in triple-negative breast cancer \cite{bendani2025decision}. 

\begin{figure*}[ht!]
\centering
\includegraphics[scale=0.79]{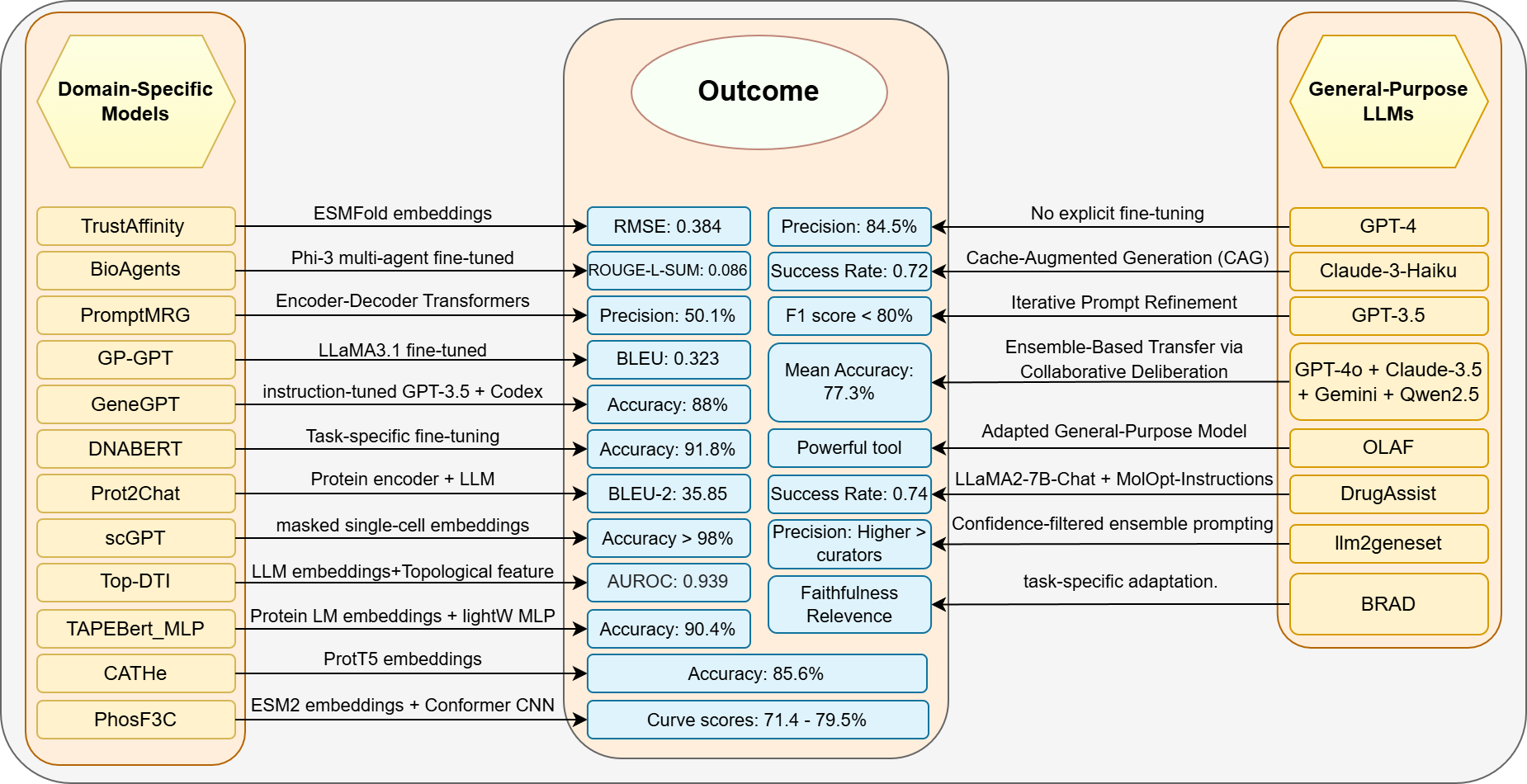}
\caption{Comparison between domain-specific and general-purpose language models, with focus on training strategies, and performance.}
\label{fig:rq2}
\end{figure*}

\x{Together, these domain-specific models offer improved performance on specific benchmarks; however, direct comparisons are constrained by heterogeneous evaluation protocols, datasets, and metrics across studies, making broad generalizations difficult.} By utilizing the explicit inclusion of prior knowledge into model formulations, tokenization schemes, and pretraining data, these domain-specialized LLMs illustrate an efficient bridging of linguistic representation and biological interpretation. Figure~\ref{fig:rq2} illustrates the comparative overview of GenAI models, contrasting domain-specific bioinformatics language models with general-purpose LLMs.

\subsubsection{General Purpose LLMs in Bioinformatics: Strengths and Shortcomings}

General-purpose LLMs such as GPT-3.5, GPT-4o-mini, and GPT-4o have demonstrated notable capabilities in various bioinformatics applications when supported by appropriate prompting strategies, fine-tuning, or integration within modular frameworks \cite{zhu2025enhancing}. These models benefit from their large scale, strong generalization, and \x{advanced} reasoning abilities. Their main advantages are observed in tasks such as semantic synthesis, literature summarization, and flexible reasoning within biomedical texts.

In information retrieval and question answering, GPT-3.5 achieved 97.3\% accuracy in gene-related queries within seven attempts and outperforms Google Bard \cite{piccolo2023evaluating}. Similarly, GPT-4o and Claude Sonnet-3.5 produced higher-quality outputs for biomedical question answering and gene function extraction compared to earlier models, particularly when combined with prompt engineering techniques such as ensembling and confidence filtering. The llm2geneset system illustrated this improvement, where 33-94\% of LLM-generated gene sets were significantly enriched in human-curated references (Bonferroni-adjusted $P<0.01$). GPT-4o achieved 69\% enrichment that closely matches human curators (70\%), and reached 95\% HUGO Gene Nomenclature Committee (HGNC) compliance, outperforming GPT-3.5 and GPT-4o-mini (88\%) \cite{zhu2025enhancing}. These results indicate that general-purpose LLMs can achieve performance comparable to traditional \x{Over-representation} Analysis (ORA) methods for information extraction, particularly when used interactively or in ensemble configurations.

However, their effectiveness declines in highly specialized multi-step analytical tasks. On the BixBench benchmark, GPT-4o achieved only $\sim$9\% accuracy in open-ended evaluations, while Claude 3.5 Sonnet reached $\sim$17\%; both of them performed at random levels in multiple-choice settings, which shows inadequate performance for real-world analyses \cite{mitchener2025bixbench}. \x{Nevertheless, direct comparison remains challenging because the reported results are based on heterogeneous tasks and evaluation criteria, and the observed performance gaps may be influenced by methodological differences rather than fundamental model limitations.} Domain-augmented models such as BioAgents \cite{mehandru2025bioagents}, built on Phi-3 with orthogonal fine-tuning and self-aware methods, significantly outperformed GPT-4o. BioAgents achieved ROUGE-1 = 0.121, ROUGE-2 = 0.012, ROUGE-L = 0.071, and ROUGE-L-SUM = 0.086, clearly \x{outperforming} GPT-4o baselines. The Galactica \cite{taylor2022galactica} model demonstrates stronger performance than general-purpose LLMs, such as GPT-3 by 68.2\% versus 49.0\% accuracy on biological reasoning tasks. Specifically, Galactica 120B achieves superior accuracy in domain-specific benchmarks and knowledge probes, highlighting the importance of curated scientific corpora over massive but unfiltered training data.

Performance can be enhanced when these models are integrated into modular systems. OLAF \cite{riffle2025olaf} and BRAD \cite{pickard2024language} extended the capabilities of GPT-3.5 and GPT-4 beyond conventional Retrieval-Augmented Generation (RAG) by incorporating agentic execution and external bioinformatics tools such as Scanpy and PubMed. For BRAD, the RAG pipeline produced higher faithfulness and relevance scores than both standard LLM and Enhanced RAG (ERAG) settings. Similarly, Claude-3-Haiku, using cache-augmented generation (CAG) via Amazon Bedrock, achieved consistently strong performance, \x{although its performance varied with dataset complexity, indicating the need for adaptive prompting strategies} \cite{nair2025prediction}. Further quantitative evaluations reinforce these observations. GPT-4o and GPT-4 presented strong precision and recall (Precision = 0.95, Recall = 0.94) without fine-tuning, with performance enabled by prompt design and temperature tuning \cite{smith2025funcfetch}. GPT-4 also achieved competitive retrieval metrics: Hit Ratio (HR) of 23.32\% @10, 24.39\% @15, and 24.59\% @20 for $N=10$; and 53.90\% @10, 58.38\% @15, and 59.55\% @20 for $N=20$, demonstrating substantial improvements with larger retrieval scopes \cite{chang2024gene}. 

Domain-specialized fine-tuning further improves molecular design tasks. DrugAssist \cite{ye2025drugassist}, developed on LLaMA2-7B-Chat with MolOpt-Instructions, surpassed GPT-3.5 and BioMedGPT-LM on 16 molecular optimization tasks. It achieved a valid molecule generation rate of 0.98 and a success rate of up to 0.80 across Quantitative Estimate of Drug-likeness (QED+), Solubility+, and Blood-Brain Barrier Penetration+ (BBBP+) tasks, significantly higher than GPT-3.5-turbo (valid ratio 0.94--0.98; success ratio $\leq$0.16) and BioMedGPT-LM (valid ratio $\leq$0.46; success ratio $\leq$0.18). Fine-tuned domain-specific models generally outperformed general-purpose LLMs, with BioLinkBERT and fine-tuned GPT-3 achieving the highest F1 scores of 0.804 and 0.810, respectively \cite{karkera2023leveraging}.

Overall, general-purpose LLMs demonstrate strong performance in flexible reasoning, summarization, and retrieval tasks, often matching or surpassing traditional pipelines and earlier domain-specific models. Their main weaknesses \x{emerge} in fine-grained biological analyses, for instance, protein structure prediction, gene expression modeling, and sequence alignment, where specialized models retain a clear advantage. Nevertheless, their versatility, linguistic competence, and accessibility make them valuable tools. 

\begin{figure*}[ht!]
\centering
\includegraphics[scale=0.75]{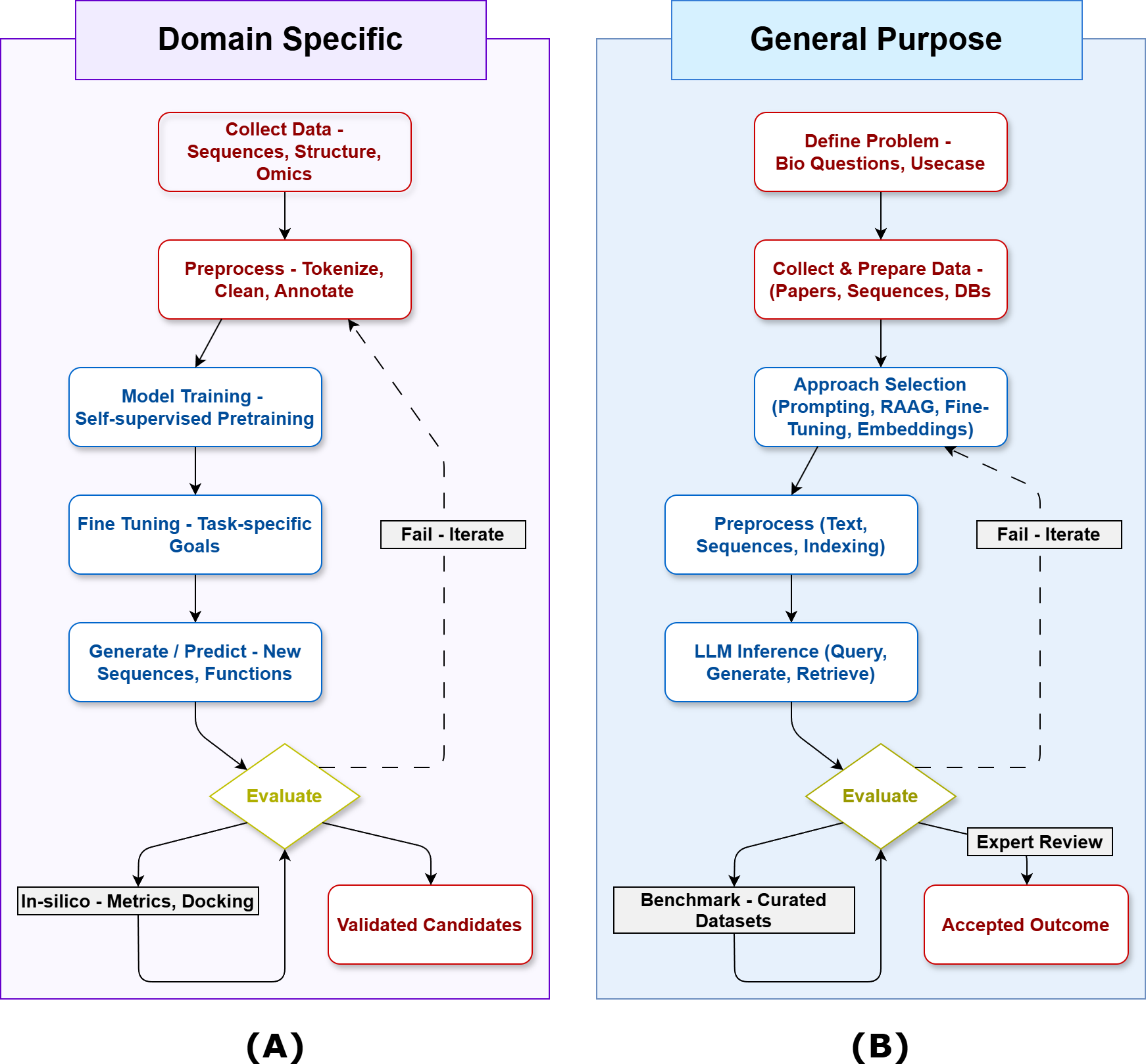}
\caption{Comparative workflow demonstrating the variation in the data processing, optimization, and fine-tuning between (A) the domain-specific models and (B) the general-purpose language models in bioinformatics.}
\label{fig:ds_gp}
\end{figure*}

Figure \ref{fig:ds_gp} illustrates this contrast by outlining the respective workflows of domain-specific and general-purpose GenAI models in bioinformatics, emphasizing differences in data preparation, model development, and evaluation strategies.

\subsubsection{Transfer Learning and Embedding Effectiveness}

One of the most consistent findings across applications of GenAI in bioinformatics is the \x{often substantial, though task-dependent} impact of pretraining strategies and embedding quality on downstream performance. Recent advancements demonstrate that transfer learning and embedding quality play a decisive role in determining downstream performance across genomics, proteomics, and transcriptomics. Comparative analyses consistently indicate that domain-specific pretraining objectives, biologically meaningful tokenization, and multimodal representations yield substantially higher predictive accuracy than general-purpose LLMs.

Tokenization strategies have a critical impact on model performance. For example, DNABERT \cite{ji2021dnabert} employed a k-mer tokenization strategy that successfully captured sequence motifs missed by generic tokenizers, resulting in over 92\% accuracy in promoter classification, splice site detection, and transcription factor binding prediction. GROVER \cite{sanabria2024dna}, trained using byte-pair encoding, further improved promoter prediction accuracy to nearly 99\%, outperforming fixed k-mer and TF-IDF schemes. These results highlight that even the choice of tokenization during pretraining can significantly influence genomic predictive performance. Another study \cite{shulgina2024rna} developed two types of RNA-specific models: a GNN incorporating structural information and a compact GPT-like language model (LM) using sequence-only data. For the RNA-LM, overlapping triplet nucleotide tokenization with rotary positional embeddings proved most effective.

PLMs further illustrate the benefits of transfer learning. Embedding-based models such as ProtT5, ProstT5, and ESM-2 consistently surpass classical sequence alignment tools, including Basic Local Alignment Search Tool (BLAST), Foldseek, and DALI~\cite{chen2025evaluating, pantolini2024embedding}. On the CAFA3 \cite{zhou2019cafa} benchmark, ESM-2 (650M) achieved Fmax scores of 0.542 for biological process, 0.619 for molecular function, and 0.693 for cellular component prediction, while ProtT5 reported 0.536, 0.575, and 0.674 for the same tasks. These values exceed ProtBERT and similarity-based methods by a clear margin. Fine-tuned embeddings combined with simple classifiers such as k-nearest neighbors have also outperformed raw model predictions in functional annotation tasks \cite{dickson2024fine}. Moreover, tools such as learnMSA2 \cite{becker2024learnmsa2} and pLM-BLAST \cite{kaminski2023plm} extend these principles to sequence alignment, enabling accurate matching in low-identity regions where traditional methods fail.

Pretrained embeddings enable shallow models to achieve strong performance, sometimes surpassing deeper baselines. CATHe \cite{nallapareddy2023cathe}, which integrates ProtT5 embeddings with a lightweight neural network, reached 85.6\% accuracy across 1,773 CATH superfamilies and 98.2\% on the largest 50 families. Structural validation using AlphaFold2 confirmed over 90\% of predictions on Pfam domains. Similarly, T4SEfinder \cite{zhang2022t4sefinder} combined a protein-specific language model with a multilayer perceptron, achieving 96.5\% accuracy and an \x{Matthews correlation coefficient} (MCC) of 0.838 in type IV effector classification, outperforming CNN and Bidirectional Long Short-Term Memory (BiLSTM) baselines.

Fusion-based methods further enhance predictive power by integrating complementary representations. PhosF3C \cite{liu2025phosf3c} combined ESM-2 embeddings with convolutional and transformer branches, achieving an AUC of 0.795 for serine phosphorylation site prediction and outperforming MusiteDeep and DeepPSP. Bingo \cite{ma2024bingo} bridged ESM-2 embeddings with graph attention networks, reporting up to 30\% higher AUC compared to CNN and BiLSTM models in cross-species essential gene prediction. Structure-informed models such as ESMFold~\cite{lin2023evolutionary} and AlphaFold2~\cite{jumper2021highly} further demonstrate how large-scale pretrained embeddings generalize from sequences to 3D structures, establishing new performance standards in protein engineering and drug discovery.

Transfer learning principles are increasingly applied to multi-omics data. scGPT~\cite{cui2024scgpt}, pretrained on large-scale single-cell datasets, generates latent embeddings that outperform scBERT, GEARS, and scGLUE in cell type annotation and cellular response prediction. vcMSA~\cite{kaminski2023plm} also demonstrates the utility of protein embeddings for MSA, maintaining robust performance in low-identity regions. These results collectively emphasize that domain-specific embeddings provide a unified and scalable approach across diverse biological modalities. Overall, embedding quality frequently exerts a stronger influence on task accuracy than architectural complexity. Domain-specific pretraining enables lightweight or hybrid models to outperform deeper architectures and traditional bioinformatics pipelines. As embedding fusion and multimodal integration continue to advance, transfer learning is expected to remain a central strategy for enhancing the effectiveness of GenAI in bioinformatics. Table \ref{tab:transfer_comparison} presents a comparative overview of key models and their reported performances.

\begin{table*}[ht!]
\centering
\scriptsize
\caption{Performance comparison of domain-specialized models across diverse biological tasks.}
\begin{tabular}{l c c c c c c c}
\toprule
\textbf{Model} & \textbf{Domain/Task} & \textbf{Dataset} & \textbf{Metric} & \textbf{Result} & \textbf{Eval. Setting} & \textbf{Val. Type}\\
\midrule
ESM-1v \cite{meier2021language} & Proteomics & DMS & Spearman $\rho$ & 0.51 (ensemble) & Zero-shot & Experimental \\
TrustAffinity \cite{badkul2024trustaffinity} & Proteomics & ChEMBL31 & RMSE & 0.384 & OOD & External \\
ProGen2 \cite{nijkamp2023progen2} & Proteomics & ProteinGym & Fitness prediction & SOTA & Zero-shot & Held-out test\\
ProtT5-XL-U50 \cite{vitale2024evaluating} & Proteomics & Pfam-seed & Error rate & 7.28\% & Remote-homology & Clustered split\\
ESM-2 \cite{vitale2024evaluating} & Proteomics & Pfam-seed & Error rate & 7.67\% & Remote-homology & Clustered split \\
DNABERT \cite{ji2021dnabert} & Genomics & ENCODE ChIP & Accuracy & 91.8--91.9\% & TFBS Prediction & Held-out \\
MegaMolBART \cite{moayedpour2024representations} & Chemoinformatics & LNP & AUC & 0.981 & Binary classify & Test split\\
SpliceBERT \cite{chen2024self} &	Transcriptomics & Mercer's & Average Precision & 0.745 & Branchpoint & Nested 10-fold\\
TrASPr+BOS \cite{wu2026generative} & Transcriptomics & GTEx & Pearson Correlation & 0.82 & Percent Spliced & Chromosome \\
BioLLMNet \cite{abir2025biollmnet} & Transcriptomics & RPI1460 & Accuracy & 92.30\% & Interaction & 5-fold CV\\
RNA-DCGen \cite{shahgir2024rna} & Transcriptomics & bpRNA & F1-score & 0.4 & Structure design & F-generation \\
GL4SDA \cite{la2025gl4sda} & Transcriptomics & RNADisease v4.0 & Accuracy & 84.60\% & Link prediction & 10-fold CV \\
RNA-GPT \cite{xiao2024rna} & Transcriptomics & RNA-QA (AIS) & F1-score & 0.8494 & RNA QA & Held-out test \\
scGPT \cite{cui2024scgpt} & Transcriptomics & Multiple Sclerosis & Accuracy & 84\% & Cell annotation & Ref-query split\\
scBERT \cite{yang2022scbert} & Transcriptomics & Zheng68K & Accuracy & 84\% & Cell annotation & Test split\\
GEARS \cite{roohani2022gears} & Transcriptomics & Perturb-Seq & Precision & +40\% & Genetic interactions & Leave-one-out\\
REDAC~\cite{de2026redac} & Transcriptomics & RNA-seq & Hallucination rate & 24.7\% & Reliability & 150 tests \\
RiNALMo \cite{penic2025rinalmo} & RNA bioinformatics & TS0 & F1-score & 0.75 & Structure Prediction & Test split\\
RNA-FM \cite{chen2022interpretable} & RNA Bioinformatics & ArchiveII600 & F1-Score & 0.941 & RNA-seq & Cross-dataset \\
ERNIE-RNA \cite{yin2025ernie} & RNA Bioinformatics & bpRNA-1m & F1-Score & 0.873 & Structure prediction & Test split \\
BioGPT \cite{luo2022biogpt} & Biomedical NLP & PubMedQA & Accuracy & 78.2\% & QA & Held-out test \\
BioMedLM \cite{bolton2024biomedlm} & Biomedical NLP & MedMCQA & Accuracy & 57.3\% & Medical-QA & Test split\\
\bottomrule
\end{tabular}
\label{tab:transfer_comparison}
\end{table*}

\subsubsection{Instruction and Task-Aware Fine-Tuning Strategies}

Large-scale pretraining provides a strong foundation for applying GenAI in bioinformatics; however, the practical effectiveness of these models increasingly relies on instruction-aware and task-specific fine-tuning. Approaches such as instruction tuning, parameter-efficient adaptation (e.g., LoRA, QLoRA), and domain-optimized prompting have become essential for aligning general-purpose architectures with specialized biological objectives. These strategies enhance model performance while ensuring reliability, explainability, and adaptability across diverse biomedical tasks.

Several recent studies highlight the benefits of targeted fine-tuning. DrugAssist \cite{ye2025drugassist} fine-tuned LLaMA2-7B-Chat with the MolOpt-Instructions dataset, achieving a solubility success rate of 0.74, a Blood-Brain Barrier Penetration (BBBP) success rate of 0.80, and a validity rate near 0.99. These results exceeded those of GPT-3.5-turbo and BioMedGPT-LM in multi-property molecular optimization, indicating the advantage of task-aware adaptation. Similarly, Top-DTI \cite{talo2025top} uses ProtT5 and MoLFormer embeddings plus persistent-homology (Betti curves and persistence landscapes) features fed into a GNN with a learnable fusion module, and this strategy outperforms prior LLM-based and other SOTA methods: on the BioSNAP \cite{zitnik2018biosnap} random split, Top-DTI achieves AUROC of 0.939 and Area Under the Precision–Recall Curve (AUPRC) of 0.941, where DrugLAMP\cite{luo2024accurate}, which mainly depends on LLM-based  
drug and target embeddings, attains AUROC of 0.917 and AUPRC of 0.922. GP-GPT \cite{lyu2024gp}, built on LLaMA3.1 and trained on biomedical knowledge bases such as ng LLaMA-based architectures fine-tuned with curated resources such as OMIM  \cite{amberger2015omim}, UniProt \cite{boutet2007uniprotkb}, and DisGeNET \cite{pilault2020extractive}, outperformed GPT-4 and BioGPT in gene–feature mapping, with a BLEU score of 0.323, F1 of 45.9\%, and accuracy of 53.3\%. In protein–ligand interaction prediction, TrustAffinity \cite{badkul2024trustaffinity} integrated ESMFold embeddings with a graph-based encoder and Gaussian process uncertainty, achieving Pearson correlations above 0.9 and outperforming BACPI in out-of-distribution (OOD) benchmarks. Collectively, these examples demonstrate that instruction-aware fine-tuning consistently yields superior task performance compared to both general-purpose and domain-pretrained baselines.

In addition to single-model adaptation, multi-agent and consensus-based instruction strategies have shown competitive or superior results. BioAgents \cite{mehandru2025bioagents}, built on Phi-3, coordinated multiple specialized agents for bioinformatics tasks and achieved ROUGE-1 scores comparable to GPT-4o on BioStars QA pairs, while improving workflow efficiency. Similarly, multi-LLM consensus frameworks for large-scale cell type annotation achieved a mean accuracy of 77.3\% across 50 datasets~\cite{yang2026large}, outperforming single-model approaches such as GPTCelltype. PromptMRG \cite{jin2024promptmrg} further demonstrated that task-specific prompting can reach SOTA performance in medical report generation without full model retraining, highlighting the efficiency of prompt-supported strategies.

Instruction tuning has also proven valuable in enrichment analysis and cross-domain applications. ProBASS \cite{gurusinghe2025probass} leverages two powerful protein language models (PLMs), ESM-2 for sequence representation learning and ESM-IF1 for structure-aware embeddings, and subsequently employs a CatBoost regression model to fine-tune predictions of $\Delta\Delta G_{\mathrm{bind}}$ values. Similarly, llm2geneset \cite{zhu2025enhancing} utilized ensemble prompting with confidence filtering, enabling GPT-4o to outperform classical enrichment methods such as ORA, with 69\% of generated gene sets significantly enriched in Gene Ontology Biological Processes and achieving strong results for Kyoto Encyclopedia of Genes and Genomes (KEGG) pathways. GeneGPT \cite{jin2024genegpt}, which combines GPT-3.5-turbo with Codex and structured API calls, achieved SOTA results on eight of nine GeneTuring benchmarks, with an average score of 0.83. The results were better than those of domain-sensitive models, including BioMedLM \cite{bolton2024biomedlm} and BioGPT \cite{karkera2023leveraging}, which highlight the competitive advantages of well-designed instruction-aware solutions.

In summary, instruction and task-conscious fine-tuning approaches have become one of the enablers of solving complex bioinformatics challenges with GenAI. Comparative analyses across tasks demonstrate that these methods demonstrate superior performance to both general-purpose and domain-specialized baselines in metrics such as accuracy, F1 score, BLEU, ROUGE, and Pearson correlation. \x{Their ability to deliver high accuracy while maintaining transparency and scalability makes them essential bridges between large-scale pretraining and the fine-grained demands of biomedical applications.}

\subsubsection{Performance Across Modalities: Sequence, Structure, and Graphs}

One of the most important trends in bioinformatics today is the move towards multimodal generative models which combine sequences, structures, and graphs. Compared to single-modality approaches, these frameworks integrate complementary biological signals and provide results with enhanced accuracy, robustness, and cross-task generalization. Models of this type are now actively used in protein-ligand binding, PPI modeling, essential gene prediction, and ST.

TrustAffinity \cite{badkul2024trustaffinity} represents a good example of this trend. \x{It integrates ESMFold embeddings with graph-based molecular representations, achieving Pearson correlations above 0.9} in protein-ligand affinity prediction and significantly lowering prediction error than in the BACPI baseline. Also, its inference time is orders of magnitude faster than conventional docking, and we have seen both efficiency and accuracy improvements.

Similarly, Bingo \cite{ma2024bingo} integrates ESM-2 sequence embeddings with graph attention networks to predict cross-species essential genes. In \textit{C. elegans}, it achieved an AUC of 0.90 and an AUPR of 0.91, and maintained highly predictive performance in human HepG2 cells. \x{These results surpass CNN, BiLSTM, and transformer-only baselines}, especially in sparse and noisy data scenarios, and highlights the benefit of multimodal fusion when dealing with challenging biological situations. Structure-centric generative approaches such as AlphaFold2 \cite{jumper2021highly} and ESMFold \cite{lin2023evolutionary} have also demonstrated that large-scale pretraining on sequences can generalize effectively to 3D structure prediction, thereby enabling downstream applications in drug discovery and protein engineering.

Latent-space generative models offer complementary advantages by encoding structural information into learned representations. DeepUrfold \cite{draizen2024deep}, for example, employs a 3D convolutional variational autoencoder to encode protein domain volumes, which produces structural embeddings of protein domains that can be used to identify evolutionary and functional relationships that cannot be identified by sequence alignment alone. While its clustering metrics are lower than those of curated resources such as CATHe \cite{nallapareddy2023cathe}, this approach offers alternative perspectives on protein structural organization, enriching the diversity of analytical strategies.

The benefits of multimodality are not limited to the study of proteins. SpatialScope \cite{wan2023integrating} integrates spatial and molecular features to impute single-cell expression profiles, reducing cell type misclassification by more than 50\% compared \x{with} Tangram. Similarly, scGPT \cite{cui2024scgpt} fuses single-cell RNA-seq, ATAC-seq, and epigenetic modalities to learn unified cellular representations. \x{This multimodal integration improves cell type annotation accuracy and enables prediction of uncertain cellular responses}, which shows that multimodal generative frameworks are adaptable in both transcriptomic and epigenomic spaces.

Altogether, these studies indicate that multimodal models are much better predictors than unimodal variants, in terms of accuracy and generalizability. Key performance metrics such as Pearson correlation, AUC, AUPR, and classification accuracy consistently favor models that integrate sequence, structural, and graph information. Moreover, the use of domain-specific embeddings and biologically informed pretraining enables these architectures to show improved performance over general-purpose LLMs on fine-grained biological reasoning tasks. As evaluation now depends more on clear and measurable benchmarks, multimodal fusion has become a significant approach for improving GenAI applications in bioinformatics.

\x{Models compared over sequence, structure, and graph modalities demonstrate that domain-specialized generative models show superiority over general-purpose LLMs on tasks requiring fine-grained biological reasoning, though the degree of advantage depends heavily on task type, dataset, and evaluation metric.} Specialized architectures, domain-specific pretraining, and biology-aware tokenization drive improved performance, stability, and clarity, specifically in low-resource or difficult settings. High-quality \x{pretrained} embeddings are important for downstream performance improvements when fused across modalities, while instruction or task-aware fine-tuning can help even general-purpose LLMs to attain evenness with domain-tuned frameworks. \x{It should be noted that performance comparisons across the studies are not always directly comparable, as models differ in evaluation datasets, metrics, and fine-tuning strategies. Several reported improvements reflect results from the original papers rather than controlled head-to-head evaluations.}

\begin{figure*}[ht!]
\centering
\includegraphics[scale=0.13]{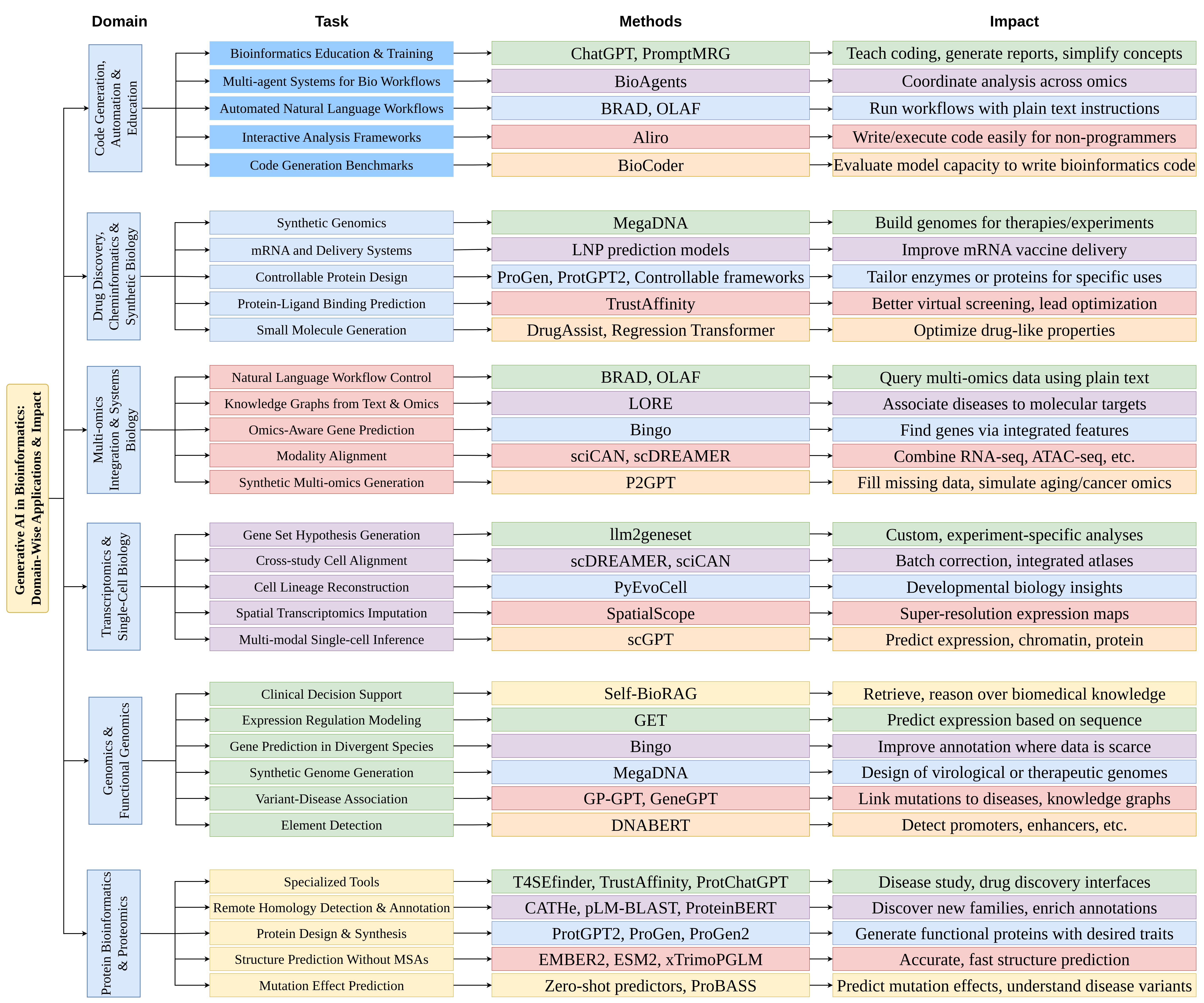}
\caption{Domain-wise representation of bioinformatics applications, highlighting the generative methods in each domain and their corresponding biological impacts.}
\label{fig:rq3_flow_tree}
\end{figure*}

\subsection{Benefiting Bioinformatics Domains (RQ3)}

GenAI has rapidly reshaped the profile of bioinformatics, enabling novel approaches for modeling, interpreting, and engineering biological systems. In contrast to conventional techniques, generative models tap into the enormous size of biological data and intricate patterns inherent within to generate insights, predictions, and even completely novel biological sequences and structures. These functionalities are not limited to one domain; however, they extend across protein bioinformatics, genomics, transcriptomics, multi-omics integration, synthetic biology, and even code generation.  Figure \ref{fig:rq3_flow_tree} shows a flow tree chart that graphically structures bioinformatics domains and their respective GenAI-enabled tasks, aiding in visualizing how generative models assist in each domain.

\subsubsection{Protein Bioinformatics and Proteomics}
GenAI has developed protein bioinformatics with powerful tools for the analysis, prediction, and design of proteins that outshine those of traditional methods. One of the most significant applications is mutation effect prediction, where zero-shot models such as ESM-1b \cite{meier2021language} have shown that the potential functional impact of mutations can be inferred directly from protein sequence data in an unsupervised manner which provides valuable insights into disease-associated variants across diverse protein families. In structural bioinformatics, EMBER2 \cite{weissenow2022protein} demonstrated that protein language model embeddings had the capacity to make precise predictions of inter-residue distances without requiring MSAs, such that predictions could be made quicker and at lower cost. At a larger scale, ESM-2 \cite{chen2025evaluating} and xTrimoPGLM \cite{chen2024xtrimopglm} demonstrated that PLMs can implicitly learn atomic level structure features directly from sequences and make predictions consistent with physics based methods, while achieving results several orders of magnitude faster. Similarly, PLM-interact \cite{liu2025plm} is useful for cross-species PPI prediction, mutation-aware interaction analysis, and virus–host interaction prediction, showing value for disease biology, variant-effect interpretation, and host–pathogen studies. For mutation-effect prediction, PLM-interact reached AUPR 0.612 and AUROC 0.794, while zero-shot models were near random.

\begin{table*}[ht!]
\centering
\renewcommand{\arraystretch}{1.3} 
\setlength{\tabcolsep}{6pt}       
\scriptsize
\caption{Overview of GenAI models across major bioinformatics domains, including protein modeling, genomics, transcriptomics, multi-omics integration, drug discovery, and automation. The table outlines each model’s core applications and key benefits.}
\label{tab:rq3_table}
\begin{tabularx}{\textwidth}{
@{\hspace{1pt}}
>{\raggedright\arraybackslash}p{3.8cm} 
@{\hspace{6pt}}
>{\arraybackslash}m{3cm} 
@{\hspace{6pt}}
c 
@{\hspace{6pt}}
m{5cm} 
@{\hspace{6pt}}
X 
}
\hline
\textbf{Type} & \textbf{Method} & \textbf{Year} & \textbf{Primary Applications} & \textbf{Key Benefits} \\
\hline

\multirow{8}{=}{\textbf{Protein Bioinformatics}}
& EMBER2 \cite{weissenow2022protein} & 2022 & Inter-residue distance prediction & Structure prediction without MSAs \\
& ProtGPT2 \cite{ferruz2022protgpt2} & 2022 & Protein sequence generation & \textit{De novo} generation of realistic sequences \\
& T4SEfinder \cite{zhang2022t4sefinder} & 2022 & Effector protein detection & Embedding-based pathogen analysis \\
& ProGen \cite{madani2023large} & 2023 & Protein design & Explores functional non-natural proteins \\
& CATHe \cite{nallapareddy2023cathe} & 2023 & Remote homology detection & Detects distant evolutionary relationships \\
& ProtChatGPT \cite{wang2024protchatgpt} & 2024 & Protein Q\&A and exploration & Conversational interface for structure-function \\
& ESM-2 \cite{chen2025evaluating} & 2025 & Atomic-level structure prediction & Learns 3D structure directly from sequence \\

\hline

\multirow{5}{=}{\textbf{Genomics}}
& GP-GPT \cite{lyu2024gp} & 2024 & Gene–phenotype association & Multi-layer biological integration \\
& MegaDNA \cite{shao2024long} & 2024 & Synthetic genome generation & Supports gene therapy and virology \\
& Bingo \cite{ma2024bingo} & 2024 & Eukaryotic gene prediction & Useful for low-resource annotations \\
& GET \cite{fu2025foundation} & 2025 & Transcriptional modeling & Links DNA sequence to expression \\
& LORE \cite{li2025large} & 2025 & Disease–gene extraction & Graph embedding of clinical-genomic links \\
\hline

\multirow{7}{=}{\textbf{Transcriptomics}}
& scDREAMER \cite{shree2023scdreamer} & 2023 & Cross-species atlas alignment & Removes batch effects, preserves signals \\
& sciCAN \cite{shree2023scdreamer} & 2023 & Gene with chromatin integration & Unified, noise-resistant embedding \\
& SpatialScope \cite{wan2023integrating} & 2023 & Super-res spatial transcriptomics & Reveals cell–cell interactions \\
& scGPT \cite{cui2024scgpt} & 2024 & Joint prediction (gene, chromatin, protein) & Captures cell regulation and cellular response \\
& ChatGPT \cite{wang2024protchatgpt} & 2024 & Learning and coding support & Bioinformatics tasks and education \\
& PyEvoCell \cite{mathur2025pyevocell} & 2025 & Lineage and trajectory mapping & Models differentiation with priors \\
& llm2geneset \cite{zhu2025enhancing} & 2025 & Gene set generation & Boosts pathway analysis \\
\hline

\multirow{11}{=}{\textbf{\x{RNA Biology}}}
& \x{RNA-FM \cite{chen2022interpretable}}
& \x{2022}
& \x{3D modeling, RNA--protein interaction}
& \x{Generalizes across unseen RNA types} \\
& \x{SpliceBERT \cite{chen2024self}}
& \x{2024}
& \x{Splice-site prediction, branchpoint}
& \x{Cross-species splicing without labeled data} \\
& \x{RhoFold+ \cite{shen2024accurate}}
& \x{2024}
& \x{RNA 3D structure}
& \x{Full-atom prediction surpassing AlphaFold3} \\
& \x{RNA-GPT \cite{xiao2024rna}}
& \x{2024}
& \x{RNA annotation, disease association}
& \x{Literature-grounded sequence interpretation} \\
& \x{RNA-DCGen \cite{shahgir2024rna}}
& \x{2024}
& \x{Conditional RNA sequence design}
& \x{Multi-constraint generative RNA design} \\
& \x{RiNALMo \cite{penic2025rinalmo}}
& \x{2025}
& \x{Structure prediction, ncRNA classification}
& \x{Strong generalization to unseen RNA families} \\
& \x{ERNIE-RNA \cite{yin2025ernie}}
& \x{2025}
& \x{RNA-protein binding}
& \x{Zero-shot structure prediction via attention} \\
& \x{BioLLMNet \cite{abir2025biollmnet}}
& \x{2025}
& \x{RNA--molecule interaction}
& \x{Unified interaction prediction} \\
& \x{GL4SDA \cite{la2025gl4sda}}
& \x{2025}
& \x{snoRNA--disease association prediction}
& \x{Improve ncRNA discovery} \\
& \x{LitSumm \cite{green2025litsumm}}
& \x{2025}
& \x{ncRNA literature summarization}
& \x{Scales curation to 177,500 papers} \\
& \x{CrossLLM-Mamba \cite{sadia2026crossllm}}
& \x{2026}
& \x{RNA--molecule affinity}
& \x{Dynamic state-space interaction modeling} \\
\hline
\multirow{5}{=}{\textbf{Multi-omics Integration}}
& sciCAN \cite{xu2022scican} & 2022 & scRNA \& scATAC alignment & Captures shared signals across omics \\
& scDREAMER \cite{shree2023scdreamer} & 2023 & Unified cell atlases & Shows subtle biological variation \\
& Bingo \cite{ma2024bingo} & 2024 & Cross-omics gene prioritization & Finds conserved functional markers \\
& P2GPT \cite{sidorenko2024precious2gpt} & 2024 & Multi-omics data generation & Conditioned on age, tissue, disease \\
& BRAD \cite{pickard2024language} & 2024 & Natural Language-based omics querying & Enables no-code workflows \\
\hline
\multirow{7}{=}{\textbf{Drug Discovery \& SynBio}}
& ProtGPT2 \cite{ferruz2022protgpt2} & 2022 & Enzyme/protein generation & Constrained \textit{de novo} design \\
& ProGen2 \cite{nijkamp2023progen2} & 2023 & Functional protein synthesis & Matches nature’s catalytic performance \\
& Regression Transformer \cite{born2023regression} & 2023 & Molecule generation \& prediction & Efficient exploration of chemical space \\
& LNP Framework \cite{moayedpour2024representations} & 2024 & mRNA delivery performance & Aids vaccine LNP design \\
& MegaDNA \cite{shao2024long} & 2024 & Genome synthesis & Useful in synthetic setups \\
& TrustAffinity \cite{badkul2024trustaffinity} & 2024 & Ligand affinity prediction & Generalizes to novel ligands \\
& DrugAssist \cite{ye2025drugassist} & 2025 & Molecular optimization & Assists in lead compound design \\
\hline
\multirow{5}{=}{\textbf{Code Generation \& Automation}}
& Aliro \cite{cui2024scgpt} & 2024 & LLM-powered coding & Helps bio learners with coding \\
& PromptMRG \cite{jin2024promptmrg} & 2024 & Diagnostic report generation & Combines input and domain knowledge \\
& BRAD \cite{pickard2024language} & 2024 & Natural Language-to-pipeline automation & Builds workflows from prompts \\
& OLAF \cite{riffle2025olaf} & 2025 & Natural Language-executed pipelines & Enables reproducible no-code analysis \\
& BioAgents \cite{mehandru2025bioagents} & 2025 & Multi-agent omics workflows & Boosts reproducibility and throughput \\
\hline
\end{tabularx}
\end{table*}

\x{The capacity for novel protein design has also been transformed}. Generative models such as ProtGPT2 \cite{ferruz2022protgpt2}, ProGen \cite{madani2023large}, and ProGen2 \cite{nijkamp2023progen2}  have demonstrated the capability of producing biologically realistic and functional protein sequences, venturing into areas of sequence space outside of natural proteins. From this idea, designable structure frameworks \cite{ferruz2022controllable} then created property conditioned generation, where researchers could direct the generative process towards specific functional or biophysical properties, which is a major advancement in synthetic biology and enzyme engineering. Remote homology detection and functional annotation have also benefited greatly from GenAI. CATHe \cite{nallapareddy2023cathe} and pLM-BLAST \cite{kaminski2023plm} utilize high dimensional embeddings to detect distant evolutionary relationships and classify proteins into previously unknown families. These approaches surpass traditional alignment based approaches in sensitivity, range, augment databases, and enable evolutionary functional analysis.

Specialized software has also emerged, such as T4SEfinder \cite{zhang2022t4sefinder}, a pre-trained embedding based method to identify secreted effector proteins in bacterial genomes, towards deciphering pathogen–host interaction. Similarly, TrustAffinity \cite{badkul2024trustaffinity} calculates reliable binding affinities among proteins and ligands, facilitating therapeutic development pipelines. \x{Similarly}, hybrid methods such as ProtChatGPT \cite{wang2024protchatgpt} integrate multilevel representations such as sequence, structure, and function into question answering interfaces for protein analysis and drug target identification. However, the Prot2Chat \cite{wang2025prot2chat} framework is specifically designed for protein sequence and structure understanding, enabling the model to interpret complex biological information and produce meaningful natural-language explanations, primarily focuses on protein-level tasks.

Table \ref{tab:rq3_table} summarizes important GenAI models, their fundamental tasks, and applications within bioinformatics domains. Collectively, these developments illustrate how GenAI naturally integrates sequence, structure, and functional knowledge natively to allow researchers to annotate, design, and explore proteins at scales and depths previously unimaginable. From predicting fine-grained mutation effects to designing entirely new enzymes, GenAI has become a pillar of strength in modern protein bioinformatics and proteomics.

\subsubsection{Genomics and Functional Genomics}
In genomics and functional genomics, GenAI has created new possibilities for interpreting the regulatory grammar of DNA and analyzing its relation to traits. In contrast to conventional models that depend on hand-designed features or shallow learning, generative methods handle DNA such as a language, discovering motifs and functional elements directly from sequence data. DNABERT \cite{ji2021dnabert}, for instance, interprets genomic sequences as k-mers and achieves SOTA in promoter, splice site, and regulatory motif identification with impressive generalizability. Its applicability across species and attention-based visualization also indicates possibilities beyond genomics, into transcriptomics and synthetic biology.

One other promising direction is analyzing the relationship between genetic variation and disease characteristics. Models such as GP-GPT \cite{lyu2024gp} and GeneGPT \cite{jin2024genegpt} illustrate how LLMs can combine gene, protein, and trait information to forecast disease associations, \x{answer complex biomedical questions, and construct knowledge graphs capturing multi-layered biological relationships}. This functionality is especially significant in medical genetics and precision medicine, where variant interpretation continues to be a considerable challenge.

Generative models have been used on unexplored or extremely divergent genomes as well. MegaDNA \cite{shao2024long}, for instance, is capable of producing realistic bacteriophage genomes with regulatory and coding \x{sequences}, which will aid in advances in synthetic virology and the design of gene therapies. Similarly, Bingo \cite{ma2024bingo} uses language models alongside GNN to foresee necessary genes in eukaryotic and parasitic genomes, complementing annotation in organisms where experimental information is limited.

At a higher level of regulation, models such as GET\cite{fu2025foundation} evaluate transcriptional programs by predicting gene expression profiles from sequence and chromatin data, and reveal both universal and cell-type-specific transcription factor and regulatory element networks. Such models are especially useful for linking DNA sequence context to downstream functional readouts, even across cell types and conditions.

\x{At the translational level}, hybrid systems such as Self-BioRAG \cite{jeong2024improving} even go further, integrating retrieval and reasoning to augment medical decision-making, illustrating how genomic knowledge can be directly applied in clinical contexts. Collectively, these generative strategies have taken genomics from the domain of static annotation to dynamic interpretation of sequences, prediction of regulatory interactions, and even synthesis of new genomes.

\subsubsection{RNA, Transcriptomics, and Single-cell Biology}
Single-cell gene expression has been among the most challenging and data-hungry domains of bioinformatics and a domain where GenAI has made crystal-clear advances. Rather than only predicting cell types or clustering gene expression profiles, new models are able to capture the full richness of single-cell states and regulatory dynamics. Particularly, scGPT \cite{cui2024scgpt} is emblematic of this movement by jointly predicting single-cell gene expression, chromatin accessibility, and surface protein abundance. The model can infer regulatory networks and predict cellular responses to changes, making it a general platform for exploring heterogeneity during development, disease, and therapeutic processes.

Generative approaches have also helped the spatial dimension of transcriptomics. One such example is SpatialScope \cite{wan2023integrating} that super-resolves sparse spatially resolved transcriptomic data to single-cell resolution, fills in missing genes and determines cell–cell interactions in tissue contexts. It allows researchers to relate molecular expression to tissue organization and function more accurately. On a complementary dimension, PyEvoCell \cite{mathur2025pyevocell} supports cellular lineage and trajectory reconstruction using computational bioinformatics methods and suggests probable differentiation pathways based on existing data.

\x{Individual cell datasets from diverse experiments, species, and modalities were previously integrated through extensive manual curation}; however, generative models do so automatically to a significant degree. Software such as scDREAMER \cite{shree2023scdreamer} aligns millions of cells into common atlases with retained biological signals and enables comparisons across species and conditions. Similarly, sciCAN \cite{shree2023scdreamer} uses adversarial training to align gene expression and chromatin accessibility profiles to provide outputs resistant to batch effects and technical noise and yet present integrated perspectives of cellular states. Even single-cell data hypothesis creation and subsequent interpretation have been enhanced. Such as llm2geneset \cite{zhu2025enhancing} facilitate dynamic gene set creation for experimental context, improved differential expression analyses, and pathway discovery related to the context. This comes in handy with experiments that are beyond the scope of static gene set databases.

\x{RNA bioinformatics has become one of the most active application areas of GenAI, with advances in structure prediction, functional annotation, splicing analysis, sequence design, and interaction modeling. Foundation models such as RNA-FM~\cite{chen2022interpretable} and RiNALMo~\cite{penic2025rinalmo}, pretrained on tens of millions of unlabeled RNA sequences, have shown that self-supervised learning can capture structural and functional information useful for diverse tasks, including secondary structure prediction, ncRNA classification, translation efficiency estimation, and splice-site prediction. ERNIE-RNA~\cite{yin2025ernie} further improves this framework by incorporating RNA-specific base-pairing information, achieving strong performance in zero-shot secondary structure prediction, RNA contact prediction, RNA--protein interaction prediction, and MRL estimation. GenAI has also advanced RNA 3D structure prediction. RhoFold+~\cite{shen2024accurate} combines RNA-FM representations with geometric learning to generate full-atom RNA 3D structures.}

\x{A major development in RNA bioinformatics is the shift from predictive splicing models to frameworks that support both prediction and sequence design. SpliceBERT~\cite{chen2024self} shows that large-scale pretraining across multiple vertebrate species improves splice-site prediction, branchpoint identification, and variant-effect analysis compared with conventional splicing tools. Extending this direction, TrASPr+BOS~\cite{wu2026generative} combines splicing prediction with generative optimization, enabling the design of RNA sequences with desired tissue-specific splicing patterns. Beyond regulatory modeling, GenAI has substantially advanced RNA interaction prediction. BioLLMNet~\cite{abir2025biollmnet} and CrossLLM-Mamba~\cite{sadia2026crossllm} integrate representations from RNA, protein, and molecular language models to predict diverse biomolecular interactions within a unified framework, reducing the dependence on handcrafted features and task-specific pipelines. Generative capabilities have also expanded RNA engineering. RNA-DCGen~\cite{shahgir2024rna} enables the design of RNA sequences conditioned on structural constraints while preserving biologically important regions, illustrating how foundation models can move beyond prediction toward controllable RNA generation.}

\x{Beyond sequence and structure analysis, GenAI is increasingly being used for biological knowledge extraction and interpretation. A common trend across recent studies is the integration of RNA representations with external knowledge sources, enabling tasks that were difficult to achieve with traditional RNA prediction models alone. RNA-GPT~\cite{xiao2024rna} combines RNA sequence information with scientific literature to support functional annotation, disease association discovery, and mechanistic interpretation. Similarly, LitSumm~\cite{green2025litsumm} automates large-scale ncRNA literature curation, improving the accessibility of RNA knowledge and supporting database enrichment. At the disease level, GL4SDA~\cite{la2025gl4sda} shows that language model-derived disease representations can improve snoRNA--disease association prediction compared with conventional identifier-based approaches. BIOGEN~\cite{hossain2025interpretable} further demonstrates the value of retrieval-augmented GenAI for interpreting RNA-seq data, uncovering biologically meaningful patterns that are often missed by traditional enrichment analyses. Similarly, REDAC~\cite{de2026redac} highlights the growing impact of GenAI in transcriptomics by enabling natural language-driven bulk RNA-seq differential expression analysis and pathway enrichment interpretation. By integrating LLM-based assistance with standardized RNA-seq workflows, visualization modules, biological interpretation, and reproducible report generation, REDAC makes advanced transcriptomic analysis more accessible to both non-programming biologists and experienced bioinformaticians.}

\x{These studies suggest that the contribution of GenAI extends beyond prediction toward knowledge integration, automated literature analysis, and biologically informed interpretation, providing a more comprehensive framework for RNA-focused discovery.} These advances are part of a broader transformation in single-cell analysis, from disconnected scrutiny of expression matrices towards an entire, context-perceiving study of cellular behavior. Through careful linking of modalities, addition of spatial and temporal context, and provision for hypothesis-supported probing, GenAI has united single-cell, RNA, and transcriptomics research into a more coherent and interpretable science.

\subsubsection{Multi-omics Integration and Systems Biology}
\x{Biological processes rarely occur in isolation; understanding them requires integrating information across multiple molecular layers.} Knowing about them necessitates the integration of information across more than one molecular layer. \x{GenAI has shown significant potential in addressing this complexity}, allowing researchers to integrate genomics, transcriptomics, epigenomics, and proteomics data into coherent, biologically meaningful models. For instance, P2GPT \cite{sidorenko2024precious2gpt} integrates transcriptome and methylome data, generating synthetic multi-omics data that is conditioned on age, tissue type, or disease status, a highly valuable capability in aging and cancer research.

At the level of individual cells, modality coordination has long posed a significant challenge; however, recent models have considerably eased its resolution. For instance, sciCAN \cite{xu2022scican},  employs adversarial networks to align scATAC-seq and scRNA-seq data, capturing shared biological signals and suppressing technical noise. At the same time, scDREAMER \cite{shree2023scdreamer} scales up to millions of cells, constructing integrated atlases that represent subtle biological differences among species, tissues, and conditions.

It is also now feasible to construct knowledge frameworks spanning multiple biological areas. Such tools as Bingo \cite{ma2024bingo} combine GNN with language models and predict genes of high importance by identifying conserved functional features that bridge across omics layers. Similarly, LORE \cite{li2025large} extracts disease–gene associations from text and embeds them into knowledge graphs, offering a systematic way of associating clinical observable characteristics with molecular mechanisms.

In the implementation space, software such as BRAD \cite{pickard2024language} and OLAF \cite{riffle2025olaf} bring multi-omics analysis within reach through the integration of natural language interfaces with powerful bioinformatics workflows. The tools allow researchers to pose biological questions in plain language and receive answers based on multi-layered data without writing a single line of code.

The impressive aspect of this field lies both in the ability to integrate heterogeneous data types and in how GenAI models can reveal and characterize the relationships between them. Instead of bounding omics datasets to parallel tracks, these approaches interweave them into a narrative of molecular system functioning, adaptation, and disease, providing more profound insight into the complexities of life.

\subsubsection{Drug Discovery, Cheminformatics, and Synthetic Biology}
The intersection of GenAI and molecular design has witnessed some of the most innovative and effective bioinformatics applications. In drug discovery, generative models have been used to optimize molecular properties, predict protein–ligand interactions, and suggest new compounds. For instance, DrugAssist \cite{ye2025drugassist} presented an interactive, instruction-based system that helps medicinal chemists optimize molecules for desired pharmacological or chemical properties. Similarly, TrustAffinity \cite{badkul2024trustaffinity} accurately predicts binding affinities even for out-of-distribution protein–ligand pairs, aiding virtual screening and lead optimization workflows. The Top-DTI study \cite{talo2025top} demonstrates that topological descriptors extracted from molecular images and protein contact outlines meaningfully complement language model sequence embeddings to improve DTI prediction, and the authors highlight the method’s potential for virtual screening and structural genomics while recommending integration of multi-omic data.

GenAI also advances the design of functional enzymes and proteins. Models such as ProGen \cite{madani2023large} and ProGen2 \cite{nijkamp2023progen2} have shown the capacity to produce synthetic proteins and enzymes with designed activities, even matching the functional performances of their natural alternatives. These approaches are not restricted to only random sequence generation. Methods such as controllable protein design frameworks \cite{ferruz2022controllable} enable researchers to apply functional or structural constraints during generation, opening the door for therapeutic or industrial applications with targeted specificity. Similarly, ProtGPT2 \cite{ferruz2022protgpt2} has been employed to venture into unexplored areas of sequence space, yielding structurally plausible, however, novel proteins that extend the known collection of biomolecules.

Beyond proteins, generative models are reshaping the way that researchers explore and engineer chemical space. For instance, the Regression Transformer \cite{born2023regression} predicts properties and generates molecules simultaneously, weakening the distinction between predictive and generative tasks. This bidirectional functionality enables iterative exploration of molecular domains, supporting hypothesis-supported and discovery-oriented research in cheminformatics. The reach of GenAI further extends to emerging areas such as mRNA therapeutics and nanoparticle delivery. \x{Models such} as the LNP prediction framework \cite{moayedpour2024representations} forecast the performance of lipid nanoparticles for delivering mRNA vaccines, providing guidance for the design of delivery systems for RNA therapies. In contrast, synthetic genomics and virology have benefited from tools, such as MegaDNA \cite{shao2024long}, which can generate synthetic genomes that preserve realistic coding and regulatory features, and support applications ranging from gene therapy to synthetic biology research. 

GenAI in this field is notable for its broad range of applications, including small molecules, proteins, genomes, and nanoparticles, and also for the increasing interactiveness and interpretability of its software. With features that combine creativity and accuracy, these models allow researchers to design biological molecules with greater control and confidence than ever before.

\subsubsection{Bioinformatics Code Generation, Automation, and Education}
Apart from direct biological applications, GenAI has been significantly used for the computational and teaching workflows of bioinformatics. As bioinformatics workflows typically consist of complex scripts, language models have become a valuable assistance in coding, automation, and the demonstration of tools. For instance, BioCoder \cite{tang2024biocoder} provides a benchmark against which to judge the degree to which LMs can be used to generate functional code for application domains, and to what degree they are gaining the power to handle complex bioinformatics logic, dependencies, and data structures. Aliro \cite{cui2024scgpt}, respectively, places LLMs inside an interactive framework through which users can author and execute code easily, accelerating analysis and minimizing the entry point for beginners.

Workflow-level automation has also become more viable. Tools such as BRAD \cite{pickard2024language} and OLAF\cite{riffle2025olaf} have embedded LLMs into more comprehensive frameworks to enable natural language-based data pipelines for processing and analysis. These tools simplify the running of complex processes such as single-cell RNA-seq analysis, gene annotation, and literature mining, providing reproducible and configurable workflows with no need for in-depth programming skills. Furthermore, BioAgents \cite{mehandru2025bioagents} demonstrates the strength of multi-agent schemes to integrate genomics and transcriptomics workflows' analyses, once more becoming more efficient and reproducible.

GenAI is also revolutionizing bioinformatics education and training. Experiments to compare models, for instance, ChatGPT \cite{wang2024protchatgpt} have established that the tools can solve a majority of undergraduate bioinformatics programming questions, especially when reinforced with iterative feedback. \x{Not only does this help learners understand programming concepts, it also supports instructors in designing more engaging teaching materials}. Novel applications such as PromptMRG \cite{jin2024promptmrg} highlight the capabilities of language models to generate diagnostic medical reports by combining domain knowledge and input information, an extension that can easily be adopted for producing summaries and documentation in research. By bridging biological data and human interpretation, these tools \x{facilitate a more inclusive and collaborative research and learning environment.}

Integration of GenAI into coding, automation of workflows, and education has not just enhanced efficiency; however, also made complex analyses accessible to all, transforming bioinformatics more accessible for researchers and clinicians.

\subsection{Task-Specific Improvements and Benchmarks (RQ4)}

\begin{figure}[ht!]
\centering
\includegraphics[scale=0.45]{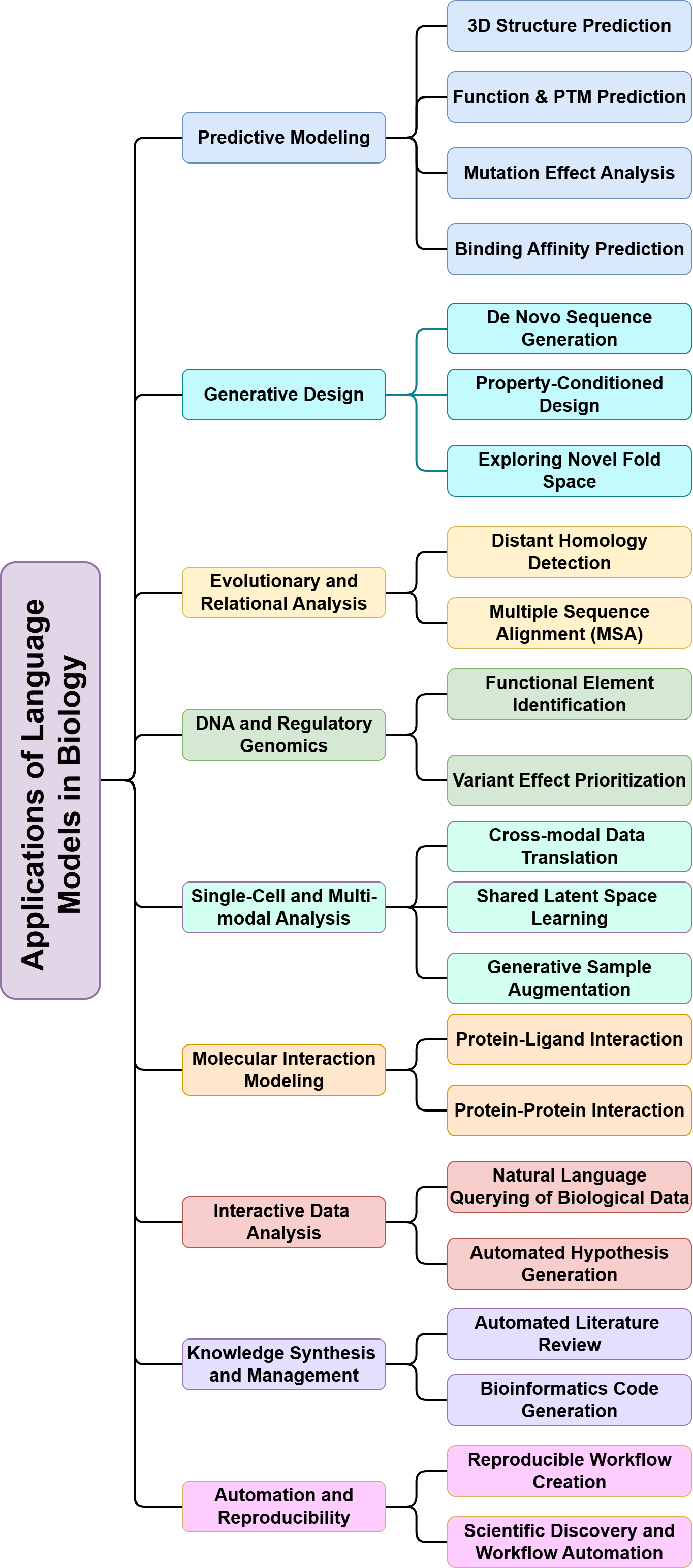}
\caption{Application centric overview of language models in bioinformatics, highlighting their roles in protein representation, function annotation, molecular interaction prediction, and cell-level analysis.}
\label{fig:rq4}
\end{figure}

The use of generative models has led to major breakthroughs in bioinformatics \cite{ji2021dnabert,luo2022biogpt,chen2024xtrimopglm,ferruz2022protgpt2,liu2025drbioright}. This section overviews how these models have improved biological research, focusing on three main applications: Protein structure prediction, improving the prediction of protein function and biophysical properties and capability to perform \textit{de novo} generation of functional proteins, genomes, and synthetic multi-omics data. Figure \ref{fig:rq4} illustrates the application-centric scenarios of language models in bioinformatics, mapping their utilization across key domains.

\subsubsection{Advancements in Protein Structure Prediction and Analysis}
The three-dimensional structure of a protein is crucial for its biological function, yet experimental determination is often slow and resource-intensive \cite{weissenow2022protein,becker2024learnmsa2,chen2024xtrimopglm}. AlphaFold2 \cite{skolnick2021alphafold} has transformed the field by achieving high accuracy in protein structure prediction. Still, these models depend on co-evolutionary information from MSAs (MSAs), which can create a computational bottleneck \cite{mcwhite2023leveraging}. The accuracy of these methods is limited when predicting structures for proteins that have only a small number of known. GenAI, particularly PLMs, shows that atomic-level structural information can be learned directly from protein sequences. This allows for quick, alignment-free structure prediction on an innovative scale \cite{lin2023evolutionary}. Similarly, ProteinBERT \cite{brandes2022proteinbert} improved downstream predictions for diverse tasks, such as secondary structure (74\% accuracy), stability (Spearman 0.76), and remote homology detection, demonstrating significant gains from gene oncology-based pretraining.

Large-scale transformers trained on hundreds of millions of protein sequences have shown that implicit rules of folding and remain interactions can be captured through language modeling. ESM-2 and ESMFold exemplify this shift, predicting high-resolution structures using only learned embeddings \cite{lin2023evolutionary}. ESMFold achieves accuracy comparable to alignment-based pipelines while operating orders of magnitude faster, enabling the creation of the ESM Metagenomic Atlas, which now contains more than 600 million predicted structures and has dramatically expanded the known structural proteome. Other alignment-free methods highlight similar advantages. EMBER2 \cite{weissenow2022protein}, for instance, combines ProtT5 embeddings with a lightweight convolutional network to predict inter-remains distances, achieving Matthews correlation coefficients near 0.30 and template modeling scores of 0.50 on CASP12–14 benchmarks, while running far faster than MSA-based approaches.

Hybrid models that integrate both sequence and structural representations have significantly advanced the prediction of mutational effects and the analysis of molecular interactions \cite{weissenow2022protein}. Recent generative models such as Prot2Chat \cite{wang2025prot2chat} further demonstrate that integrating protein sequence, structure, and textual representations enables accurate and interpretable prediction and description of protein functions. General purpose models, such as GPT-4 demonstrated strong zero-shot performance on structured prediction tasks, particularly on simpler datasets such as LLL, whereas fine-tuned BERT-based models achieved superior precision and recall on more complex PPI datasets and established state-of-the-art performance for automated literature mining \cite{rehana2024evaluating}.

Generative models are also transforming structural classification by moving beyond fixed hierarchical frameworks, for instance, CATH and SCOP, toward continuous, embedding-based representations of protein space. CATHe \cite{nallapareddy2023cathe}, trained on ProtT5 embeddings, detects remote homologues across 1,773 CATH superfamilies with 85.6\% accuracy and an F1 score of 72.4\%, while achieving 98.2\% accuracy and 95.5\% F1 on the 50 largest superfamilies. Applied at scale, CATHe expanded the CATH database with 4.62 million new domain annotations. Similarly, T4SEfinder \cite{zhang2022t4sefinder} integrates TAPEBert embeddings with a multilayer perceptron to classify bacterial type IV secreted effectors, achieving 96.5\% accuracy and a Matthews correlation coefficient of 0.838 when tested on SecReT4 v2.0 \cite{bi2013secret4} and UniProt \cite{boutet2007uniprotkb} datasets, thus enabling accurate genome-scale effector identification.

The evaluation of these models relies on well-established benchmarks and diverse metrics. UniProt \cite{boutet2007uniprotkb} and the Protein Data Bank (PDB) remain the most widely used datasets for training and validation \cite{zhang2022t4sefinder,lyu2024gp,chen2024xtrimopglm,lin2023evolutionary}, while additional resources such as SKEMPI, Pfam \cite{mistry2021pfam}, and SecReT4 \cite{bi2013secret4} provide specialised testing grounds for mutation effects, homology detection, and effector classification \cite{nallapareddy2023cathe,zhang2022t4sefinder}. The consistent use of these benchmarks facilitates fair comparison and confirms that PLMs increasingly rival or surpass alignment-based baselines while delivering far greater scalability \cite{capela2025comparative}.

These advances illustrate how GenAI has shifted protein structure prediction and analysis from alignment-dependent pipelines to adaptable, alignment-free, and hybrid approaches. They enable large-scale generation of structures exemplified by the ESM Atlas \cite{lin2023evolutionary}, efficient prediction and fine-grained exploration of the protein fold topology through models, for instance, CATHe and T4SEfinder \cite{nallapareddy2023cathe,zhang2022t4sefinder}. Beyond accelerating structural discovery, these methods provide a richer and more dynamic view of protein structure–function relationships and establish a foundation for downstream applications in drug discovery, enzyme engineering, and systems biology.

\subsubsection{Enhancing Protein Function, Interaction, and Property Prediction}

One of the central goals in bioinformatics is to understand how the linear sequence of amino acids encodes the complex functional, structural, and physicochemical properties of proteins \cite{mcwhite2023leveraging, harrigan2024improvements}. Traditional approaches rely heavily on sequence alignment and similarity-based inference, which perform well for close homologues yet often fail when sequence identity is low \cite{mcwhite2023leveraging}. PLMs have redefined this approach by treating proteins as a language and learning context-aware representations of amino acids. This enables accurate predictions of protein properties, functions, and interactions, even when evolutionary similarity is weak, thereby surpassing the limitations of alignment-based methods \cite{kaminski2023plm}. Recent hybrid approaches, such as Top-DTI \cite{talo2025top} demonstrate that combining LLM embeddings with topological descriptors can greatly improve the accuracy, robustness, and generalization of protein–ligand interaction prediction, achieving SOTA performance and reporting higher AUROC/AUPRC in both random-split and cold-split evaluations across standard benchmarks while enabling biologically validated discoveries.

The earliest and most established application of PLMs is the prediction of protein properties directly from primary sequences. Foundational models such as DNABERT \cite{ji2021dnabert} and ProtT5, pre-trained on datasets including the human genome, UniRef90 \cite{suzek2007uniref}, and millions of unannotated protein sequences to capture both local motifs and long-range dependencies. These embeddings have proven effective in identifying functional elements such as secondary structures, PTMs, disordered regions, and regulatory domains. Similarly, CATHe \cite{nallapareddy2023cathe} uses ProtT5 embeddings to detect remote homologues with accuracies exceeding $85\%$ in more than 1,700 superfamilies and more than $98\%$ for the largest 50 families, significantly extending annotation coverage in the CATH database. Such results demonstrate the importance of unsupervised representation learning for capturing distant evolutionary relationships.

The shift from qualitative annotation to quantitative prediction has been particularly evident in mutation effect analysis \cite{ferruz2022controllable}. Generative models such as ESM-1v \cite{meier2021language} predict the functional impact of amino acid substitutions in a zero-shot manner, using embeddings trained across evolutionarily diverse protein families. On 41 deep mutational scanning datasets, ESM-1v achieved average Spearman rank correlations above 0.48, exceeding supervised methods without requiring retraining \cite{meier2021language,nijkamp2023progen2}. More advanced architectures integrate sequence and structural modalities to refine predictions of protein–protein interactions. Similarly, hybrid frameworks such as TrustAffinity \cite{badkul2024trustaffinity} demonstrate that integrating learned sequence embeddings with structural coordinates consistently improves accuracy for both single and double mutation scenarios.

Structural bioinformatics has experienced transformative advances through generative pretraining. Large-scale PLMs such as ESM-2 \cite{lin2023evolutionary}, with over 15 billion parameters, have shown that structural features can be contained directly from sequences without reliance on MSAs. This led to the development of ESMFold \cite{badkul2024trustaffinity,nijkamp2023progen2}, a generative structure prediction model that achieves near AlphaFold2-level resolution while operating orders of significance faster by bypassing advanced coupling analysis. ESMFold enabled the creation of the ESM Metagenomic Atlas, a database of more than 600 million predicted structures, accelerating functional annotation of metagenomic proteins that had remained structurally structurally unannotated \cite{lin2023evolutionary}. Models such as EMBER2 \cite{weissenow2022protein} also highlight the potential of embedding-based structural prediction: using ProtT5 embeddings and a CNN. EMBER2 predicted inter-residue distances and demonstrated competitive accuracy to alignment-based methods, with Matthews correlation coefficients around 0.30 for all contacts and 0.25 for long-range contacts.

These advances are validated by strict benchmarking across multiple datasets and metrics. Protein property prediction is commonly assessed on datasets such as UniRef90 \cite{suzek2007uniref}, Pfam \cite{mistry2021pfam}, and various secondary structure and remote homology benchmarks \cite{zhang2022t4sefinder,lyu2024gp,chen2024xtrimopglm,lin2023evolutionary}. Mutation effect prediction makes use of deep mutational scanning datasets and SKEMPI, with evaluation based on Spearman correlation and $\Delta\Delta G$ prediction accuracy. Structure prediction models are validated against UniProt and PDB-derived benchmarks, using metrics such as template modelling scores, MCC for contact prediction, and global distance test scores. Such standardised benchmarks have been critical in demonstrating that PLMs consistently outperform traditional alignment-based and supervised baselines across tasks \cite{zhang2022t4sefinder,lyu2024gp,lin2023evolutionary}.

Generative models have shifted the study of proteins from reliance on sequence similarity to integrating the contextual and structural representations learned through large-scale pretraining. Two major insights are revealed: first, domain-specific PLMs trained on biological sequences consistently outperform general-purpose AI models in tasks ranging from mutation scanning to structure prediction \cite{ji2021dnabert, weissenow2022protein}; second, accuracy improves when multiple modalities, particularly sequence embeddings and structural information, are integrated within a unified framework \cite{zhang2022t4sefinder,weissenow2022protein}. These developments not only advance the ability to predict protein function, interaction, and properties but also lay the foundation for rational protein design, enabling the active engineering of proteins with desired functions rather than passive annotation of existing ones.

\subsubsection{De Novo Generation of Proteins and Multi-omics Data}

GenAI has transformed the creation of novel biological entities and synthetic datasets, marking a conceptual shift from descriptive modeling to active design \cite{ferruz2022protgpt2,chen2024xtrimopglm,shao2024long}. Protein engineering and molecule discovery relied on modifying existing structures or completely screening chemical libraries; however, generative models now enable the \textit{de novo} synthesis of proteins, molecules, genomes, and even multi-omics data, thereby opening unexplored biological spaces. Early protein generation efforts demonstrated the feasibility of autoregressive models for sequence-level creativity. For instance, ProtGPT2 \cite{ferruz2022protgpt2} produced amino acid sequences predicted to fold into stable globular structures, while ProGen \cite{madani2020progen} successfully generated artificial lysozyme enzymes that exhibited functional efficiencies comparable to natural equivalents, showing that generative language models can design functional proteins even when sequences differ heavily from natural distributions.

In chemical-bioinformatics, generative frameworks have proven equally transformative. DrugAssist \cite{ye2025drugassist}, a conversational LLM fine-tuned from LLaMA2, enables interactive molecule optimization under property constraints such as solubility and blood–brain barrier permeability. It achieved higher success rates than GPT-3.5 and BioMedGPT across sixteen optimization tasks using the MolOpt-Instructions dataset, reaching solubility and BBBP optimization success rates of 0.74 and 0.80, respectively. Similarly, warm-started encoder–decoder models pre-trained on BindingDB, MOSES, and PDB demonstrated validity as high as 99.1 percent and superior scaffold-level novelty compared with baseline generative approaches \cite{uludougan2022exploiting}. These results highlight the importance of pretraining on large-scale ligand–target benchmarks.

GenAI has also scaled beyond single biomolecules to entire genomes and transcriptomes. MegaDNA \cite{shao2024long}, a long-context transformer trained on viral genomes, successfully generated complete bacteriophage genomes of nearly 96 kilobases, which contained realistic open reading frames, regulatory motifs, and annotated proteins. This shows that generative models can reproduce the hierarchical structure of entire genetic systems, offering unprecedented opportunities in synthetic virology and genome engineering. Similarly, GP-GPT \cite{lyu2024gp} focuses on gene–feature mapping by applying LLaMA-based architectures fine-tuned with curated resources such as OMIM  \cite{amberger2015omim}, UniProt \cite{boutet2007uniprotkb}, and DisGeNET \cite{pilault2020extractive}. It outperformed GPT-4 and BioGPT in relation extraction tasks, achieving BLEU-1 scores of 0.141 and gene–phenotype mapping accuracy above 53 percent, thereby establishing itself as a specialized tool for biomedical relation discovery.

Beyond the creation of entities, generative models also integrate multi-omics datasets to address scarcity in domains, for instance, growing research and rare disease studies. P2GPT \cite{sidorenko2024precious2gpt}, which combines a transformer with a conditional diffusion model, generates biologically consistent gene expression and DNA methylation profiles conditioned on tissue type and age. The quality of this generated data was validated by the fact that classifiers trained on it could accurately predict conditioning attributes, confirming its biological accuracy. Similarly, scGPT \cite{cui2024scgpt} extends generative pretraining to single-cell transcriptomics and multi-omics integration, achieving SOTA accuracy in cell type annotation of up to 98\% and macro F1-scores consistently higher than baselines such as scBERT, GEARS, and scGLUE. 

\x{Although generative models have driven substantial progress in bioinformatics, the evidence supporting reported performance improvements is not always directly comparable. Some studies evaluate models against competing methods on the same benchmark datasets under identical experimental settings, whereas others report results from separate studies that use different datasets, evaluation protocols, and validation procedures.} The evaluation of generative models in bioinformatics relies on a growing collection of datasets and benchmarks with metrics ranging from accuracy, F1, Spearman correlation, Matthews correlation coefficient, and BLEU scores to success rates in property optimization and batch correction indices \cite{capela2025comparative,nijkamp2023progen2}. These standards highlight the maturity of the field, where performance is systematically assessed across structural, functional, and generative tasks. Overall, the generative workflow is progressing from proof-of-concept to practical utility in synthetic biology, protein engineering, drug discovery, and omics data augmentation. These models can now design novel proteins by capturing the statistical and structural regularities of biological systems with desired functions, generate entire synthetic genomes, and produce realistic multi-omics datasets to augment experimental data, a development that shows a fundamental shift from passive observation to active creation in modern bioinformatics.
To support the discussion above, \x{Table~\ref{tab:generative_validation} presents the \textit{de novo} generative outputs covered in this section and shows the type of evidence provided by each study, from computational assessments to experimental validation.}

\begin{table*}[ht!]
\centering
\renewcommand{\arraystretch}{1.3}
\setlength{\tabcolsep}{6pt}
\scriptsize
\caption{Validation status of representative \textit{de novo} generative outputs in GenAI-driven bioinformatics, including the generated output, reported evidence, validation level, and remaining gaps toward practical biological or clinical use.}
\label{tab:generative_validation}
\begin{tabularx}{\textwidth}{
@{\hspace{1pt}}
>{\raggedright\arraybackslash}p{2.8cm}
@{\hspace{6pt}}
>{\raggedright\arraybackslash}p{2.8cm}
@{\hspace{6pt}}
>{\raggedright\arraybackslash}p{4.2cm}
@{\hspace{6pt}}
>{\raggedright\arraybackslash}p{2.8cm}
@{\hspace{6pt}}
X
}
\toprule
\textbf{Model} & \textbf{Generated Output} & \textbf{Reported Evidence} & \textbf{Validation Level} & \textbf{Gap to Translational Use} \\
\midrule

ProtGPT2 \cite{ferruz2022protgpt2}
& Novel protein sequences
& Predicted stable globular folds (Structure prediction)
& Plausibility
& No reported wet-lab activity assays \\

ProGen \cite{madani2020progen}
& Artificial lysozyme enzymes
& Experimental enzymatic activity comparable to natural lysozymes
& Experimental
& Direct activity assay reported separately \\

ProGen2 \cite{nijkamp2023progen2}
& Protein fitness landscape exploration
& State-of-the-art zero-shot fitness prediction
& Benchmark-only
& No reported experimental synthesis of generated sequences \\

DrugAssist \cite{ye2025drugassist}
& Optimized small molecules
& Solubility, BBBP
& Plausibility
& Requires pharmacokinetic profiling, toxicity screening, and validation in animal models \\

MegaDNA \cite{shao2024long}
& Synthetic bacteriophage genomes
& Realistic ORFs and regulatory motifs (sequence-level plausibility)
& Plausibility
& No reported functional or infectivity assays \\

GP-GPT \cite{lyu2024gp}
& Gene--phenotype mappings
& BLEU-1 = 0.141; accuracy $>$53\% compared with GPT-4 and BioGPT
& Benchmark comparison
& No experimental validation of generated gene--phenotype associations \\

P2GPT \cite{sidorenko2024precious2gpt}
& Synthetic multi-omics profiles
& Recovery of conditioning attributes (age, tissue, disease)
& Attribute-level
& Does not evaluate preservation of patient-level biological heterogeneity \\

scGPT \cite{cui2024scgpt}
& Single-cell representations and perturbation predictions
& Cell-type annotation accuracy up to 98\%; macro F1 higher than scBERT, GEARS, and scGLUE
& Benchmark-only
& No reported experimental validation of generated cellular profiles \\

DrBioRight 2.0 \cite{liu2025drbioright}
& Cancer omics analysis and biological interpretations
& Retrospective benchmark accuracy
& Retrospective
& Requires prospective evaluation across diverse patient populations \\

PromptMRG \cite{jin2024promptmrg}
& Diagnostic report generation
& Benchmark report-quality metrics
& Retrospective
& Requires integration into clinical workflows and prospective testing \\

\bottomrule
\end{tabularx}
\end{table*}

\subsection{Limitations and Future Directions (RQ5)}
\begin{figure*}[ht!]
\centering
\includegraphics[scale=0.10]{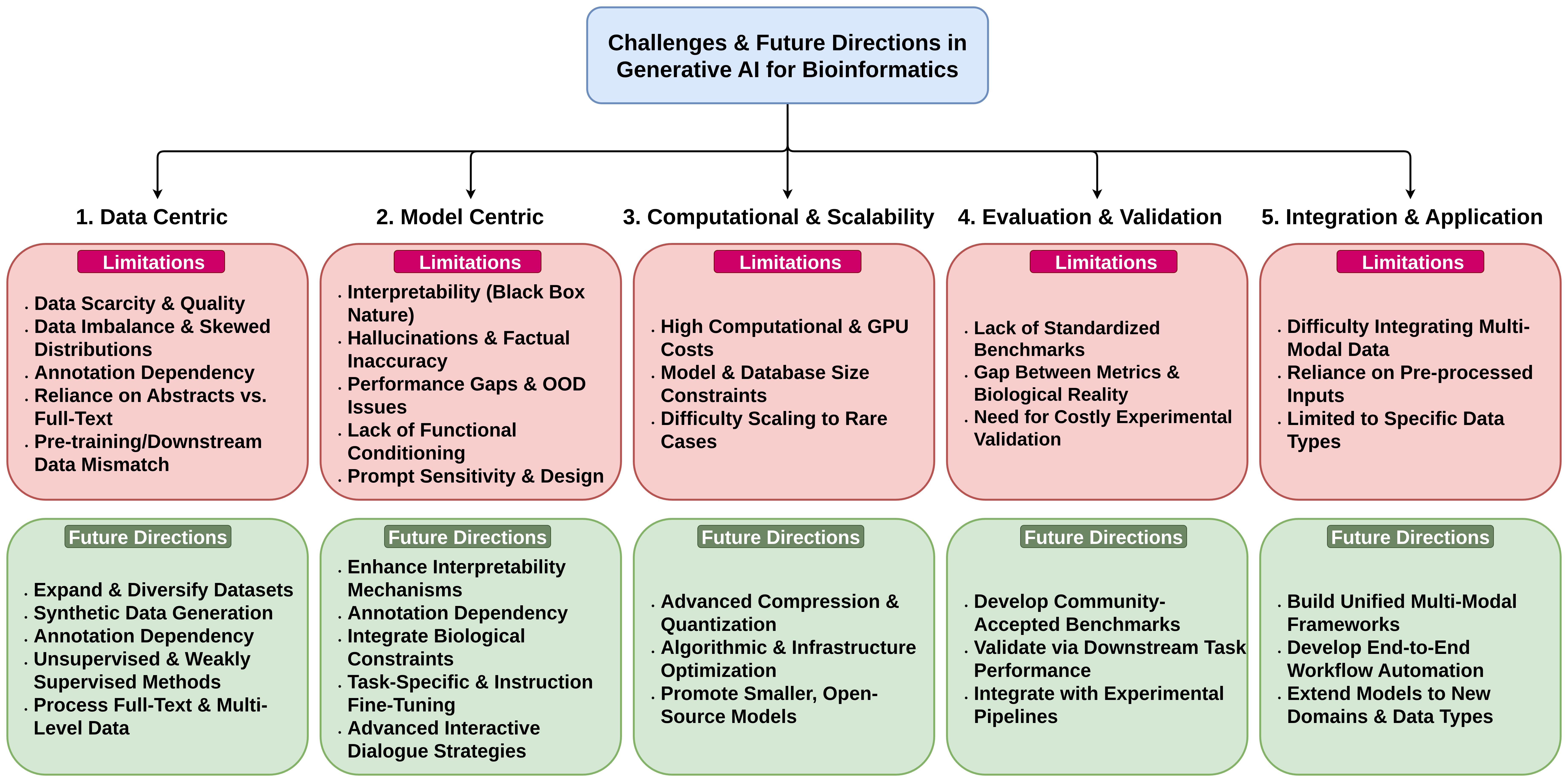}
\caption{Structured overview of open problems and future research opportunities in GenAI for Bioinformatics, highlighting data limitations, interpretability challenges, and new multimodal frameworks.}
\label{fig:rq5}
\end{figure*}

Despite remarkable progress, the integration of GenAI into bioinformatics faces several critical challenges that hamper its full translational impact \cite{chen2024xtrimopglm,chen2025evaluating}. Current models often struggle with biological interpretability, limited benchmark standardization, and generalization across diverse datasets \cite{nallapareddy2023cathe}. Addressing these limitations is essential to ensure reliability, clinical applicability, and sustainable progress in GenAI-based biological research. A structured overview of open problems and research opportunities in GenAI for bioinformatics is presented in Figure \ref{fig:rq5}.

\subsubsection{Scaling and Generalization Limitations of LLMs}
The developments of GenAI in bioinformatics are closely connected to the scaling of models. Models with billions of parameters, such as the 15 \x{Billion}-parameter ESM-2 \cite{chen2025evaluating} and the 100 \x{Billion}-parameter xTrimoPGLM \cite{chen2024xtrimopglm}, show profound abilities in understanding and generating complex biological data. The scaling has played a key role in developing foundation models that \x{learn} the underlying language of proteins and genomes. However, achieving large scale is difficult and often gives smaller improvements as the models grow. It is highlighted in the literature that it is both computationally and financially challenging to train these large models. This need for specialized hardware, for instance, high-end GPUs \cite{chen2024xtrimopglm}, limits access to well-funded institutions and slows down development and replication efforts in the broader community.

Besides resource concerns, the studies indicate that larger models are not necessarily better performers in all bioinformatics operations. One general observation is that large and general-purpose models generally fail to perform as well as smaller domain-specific models.  For example, General-purpose models such as GPT-4 show strong capabilities in workflow management and code generation \cite{tang2024biocoder,piccolo2023evaluating}; however, they are consistently outperformed by domain-specialized models on specific predictive tasks. Research indicates that ESM \cite{chen2025evaluating} and ProtBERT \cite{capela2025comparative} provide more functional annotation useful embeddings than general models do \cite{luo2022biogpt}. This shows that raw computational power cannot replace the detailed, domain-specific knowledge found in well-curated biological datasets, making domain adaptation vital for achieving top performance.

The problem of generalization is usually the nature of biological data, which is large; however, often sparse. Even when trained on billions of sequences, models have difficulty with the biological diversity, which includes rare protein families \cite{penic2025rinalmo}, less-studied organisms, or unique mutation contexts. For instance, ProGen2 \cite{nijkamp2023progen2}, despite its size, was not able to perform well on the narrow protein fitness spaces, which suggests that scale does not necessarily lead to good results. The quality of \x{pretraining} data is crucial for task-specific success \cite{nijkamp2023progen2}. This reliance on high-quality, comprehensive training data remains a major obstacle for strong generalization.

In addition, transformer architectures have a number of technical constraints to their practical use. \x{One major limitation in genomics is the narrow context window, which prevents models such as the Nucleotide Transformer and DNABERT from capturing the long-range regulatory interactions typical in gene regulation, including those organized through TADs and chromatin loops, which can span hundreds of kilobases to megabases \cite{dalla2025nucleotide, ji2021dnabert}. Architectures specifically designed for this challenge, such as Enformer \cite{avsec2021effective} and Sei \cite{chen2022sequence}, demonstrate that substantially extending the receptive field of sequence models enables more accurate modeling of distal regulatory elements and chromatin state dynamics. However, fully integrating 3D genome organization and cell-type-specific chromatin accessibility into generative language model frameworks remains an open and important challenge for the field.} The other common problem is the hallucination of models, models may produce convincing yet inaccurate information \cite{jin2024genegpt}. It is a major risk in the scientific applications, from generating incorrect code in workflows \cite{mehandru2025bioagents} to fabricating enzyme-substrate interactions \x{during} literature mining \cite{smith2025funcfetch}. These models have a structural limitation to scientific discovery and clinical application due to their black box character. It cannot be interpreted clearly to derive new biological hypotheses or rely on the predictions made by a model to make vital decisions. This is very challenging and it affects the applications from genomics \cite{ji2021dnabert} to drug discovery \cite{badkul2024trustaffinity}.

GenAI is a complicated tradeoff between the benefits of scaling and its constraints in bioinformatics. Although larger models have brought about impressive capability, they are costly and have issues with out-of-distribution generalization and fundamental technical complexities that are challenging to address \cite{chen2024xtrimopglm,chen2025evaluating}. Therefore, future progress will focus on developing efficient, data-aware, and reliable model designs that can turn large-scale learning into trustworthy and accessible scientific knowledge.

\subsubsection{Data Availability, Quality, and Representation Bias}

The effectiveness of GenAI in bioinformatics relies heavily on having large, high-quality datasets for \x{pretraining}. This creates a paradox: while models perform well in data-rich environments, many critical biological questions arise in situations with little data. The success of foundational models, for instance, ESM \cite{chen2025evaluating} and the Nucleotide Transformer \cite{dalla2025nucleotide} comes from large, publicly available data sources, for instance, UniProt and the human reference genome \cite{asim2025protein,zhang2022t4sefinder}. However, literature shows that applying these models in areas with limited data, such as rare protein families, less-studied organisms, or specific disease areas, poses a significant challenge \cite{jin2024promptmrg,chen2022interpretable, ye2025drugassist,lyu2024gp}. This information gap minimizes the performance of the model and generalization, which was observed in T4SE \cite{zhang2022t4sefinder} prediction and RNA engineering \cite{shulgina2024rna} research. To counter this, one of the most important uses of GenAI is to solve this problem by producing realistic biological data. An example is P2GPT \cite{sidorenko2024precious2gpt}, which produces multi-omics samples in advancing research, where it is often challenging to collect long-term samples.

Along with availability, the quality and organization of available data are another significant barrier. Biological data can tend to be distributed in unstructured form, such as scientific papers, and therefore, it takes much manual effort to curate and annotate. This is a labor-intensive process that is slow enough to form a bottleneck in training advanced models. As a result, there is a growing trend to use GenAI to develop its own data supply chain. Prot2Chat \cite{wang2025prot2chat} notes a modal gap between biological data and human language, as direct sequence input to LLMs often yields uneven or meaningless responses. Tools such as FuncFetch \cite{smith2025funcfetch} and LORE \cite{li2025large} utilize LLMs to automate the large-scale extraction of structured information, such as enzyme-substrate interactions and gene-disease relationships, directly from thousands of published articles. While these methods significantly speed up the creation of datasets ready for analysis, their reliability depends on the quality and potential inaccuracies of the source literature. This change implies that the issue is no longer how to be manually curated; rather, it is how to be verified automatically and minimize the errors that may follow.

One of the biggest data-related issues is representation bias. The large databases used to train foundational models do not reflect biology neutrally; they are heavily curved towards well-studied organisms, medically important human genes, and areas of research that received significant funding \cite{ji2021dnabert,meier2021language,zhang2022t4sefinder,nallapareddy2023cathe, chen2024self}. As a result, models trained on this data inherit and amplify these biases, performing well on well-represented tasks; however, they struggle with understudied areas. This limitation is clear in the lower accuracy of models on small, imbalanced datasets for complex conditions, for instance, Lupus \cite{nair2025prediction} and their difficulty in generalizing to rare protein families or new biological contexts \cite{ma2024bingo}. This bias has serious significance, potentially limiting the use of advanced tools across much of the biological domain and increasing health imbalances if models are not validated on diverse populations.

The current data distribution presents a two-fold challenge. The vast amount of public biological data has enabled the GenAI revolution in bioinformatics \cite{ferruz2022protgpt2}. However, data availability, quality, and deep representation biases are now growing as key barriers to future advancement \cite{nallapareddy2023cathe, talo2025top}. The ability of the field to progress will depend not only on finding innovative algorithms but also on making collaborative efforts to gather more unbiased, complete, and high-quality datasets, as well as developing methods that remain effective and fair despite these known challenges.

\subsubsection{Benchmark Heterogeneity and Metric Incomparability}
\x{The reviewed studies cover a wide range of tasks, including TFBS prediction \cite{ji2021dnabert}, protein fitness prediction \cite{madani2023large, nijkamp2023progen2}, cell type annotation \cite{cui2024scgpt, yang2026large}, and drug--target interaction prediction \cite{talo2025top}. They also rely on different benchmark datasets, such as CAFA3 \cite{zhou2019cafa}, BioSNAP \cite{zitnik2018biosnap}, PubMedQA \cite{jin2019pubmedqa}, and BixBench \cite{mitchener2025bixbench}, and use diverse evaluation metrics, including Accuracy, F1-score, AUC, ROUGE, and BLEU. As a result, direct performance comparisons across studies are not methodologically appropriate. For example, the ROUGE-L score reported for BioAgents ($0.071$) \cite{mehandru2025bioagents} and the Spearman correlation reported for ESM-1v ($0.48$) \cite{meier2021language} measure performance on fundamentally different tasks and therefore cannot be compared directly. Consequently, claims regarding overall model superiority remain largely qualitative and should be interpreted with caution.}

\subsubsection{Data Leakage and Homologous Splits}
\x{A common concern in genomics and proteomics benchmarking is the presence of similar sequences in both training and test sets. Foundation models such as ESM-2 and ProtT5 are pretrained on large UniProt-scale collections \cite{meier2021language, vitale2024evaluating}, and benchmarks derived from related sources may introduce homology leakage, leading to performance estimates that are higher than what would be expected under true generalization. In several studies, including CATHe~\cite{nallapareddy2023cathe} and T4SEfinder~\cite{zhang2022t4sefinder}, the sequence identity thresholds used for train--test splitting are not clearly reported. As a result, it is difficult to determine whether the reported accuracies of 85.6\% and 96.5\%, respectively, reflect generalization to novel sequences or prediction on highly similar examples. Similar concerns arise in DNABERT~\cite{ji2021dnabert}, where k-mer tokenization naturally produces overlapping sequence patterns across genomic regions. Without homology-controlled data splits, the reported TFBS accuracy of 91.8--91.9\% may partly capture sequence similarity rather than broader biological understanding. Related issues are also present in evaluations of ESM-1v and ESM-2 for mutational effect prediction~\cite{meier2021language}, where the extent of overlap between pretraining data and Deep Mutational Scanning (DMS) test datasets is not always clear. In single-cell analysis, scGPT~\cite{cui2024scgpt} was pretrained on more than 33 million cells, and downstream benchmarks based on overlapping cell atlases may similarly benefit from data proximity.}

\x{Consequently, comparisons between domain-specific and general-purpose models on these benchmarks may partly reflect differences in data overlap rather than genuine differences in model capability. When homology-filtered, organism-held-out, or other strict evaluation protocols are not employed, reported performance improvements should be interpreted cautiously and viewed as upper-bound estimates rather than definitive evidence of robust generalization.}

\subsubsection{Benchmark Saturation and Out-of-Distribution Generalization}
\x{Another important limitation is the growing saturation of several widely used benchmarks, including CAFA3 \cite{zhou2019cafa} and PubMedQA \cite{jin2019pubmedqa}. On these datasets, many architecturally different models achieve very similar performance, making it difficult to identify meaningful differences between approaches. For example, both scGPT \cite{cui2024scgpt} and scBERT \cite{yang2022scbert} report approximately 84\% accuracy for cell type annotation (Table~\ref{tab:transfer_comparison}), suggesting limited practical separation despite architectural differences. In addition, most studies evaluate models under in-distribution settings, which may overestimate their real-world utility. Evidence from scGPT~\cite{kedzierska2025zero} shows that strong benchmark results do not necessarily translate to robust performance in zero-shot settings, while results from BixBench~\cite{mitchener2025bixbench} indicate that leading foundation models can perform poorly on complex multi-step analytical tasks despite strong scores on standard benchmarks. These findings suggest that out-of-distribution and prospective evaluations remain limited and should receive greater attention in future studies.}

\subsubsection{Comparisons Between Fine-Tuned, Prompt-Based, and RAG Systems}
\x{Several comparisons in the literature evaluate models under fundamentally different settings, making it difficult to determine whether performance differences arise from model architecture or from the adaptation strategy used. DrugAssist~\cite{ye2025drugassist}, a fine-tuned LLaMA2-7B model, is compared with zero-shot GPT-3.5-turbo, meaning that the reported improvement reflects the effect of task-specific fine-tuning as well as the underlying model itself. Similarly, GeneGPT~\cite{jin2024genegpt} extends GPT-3.5-turbo with structured National Center for Biotechnology Information API calls and achieves an average GeneTuring score of 0.83, outperforming both BioMedLM \cite{bolton2024biomedlm} and BioGPT \cite{luo2022biogpt} without updating model weights. This result shows that tool integration can, in some cases, provide benefits similar to domain-specific pretraining for retrieval-focused tasks. BioAgents~\cite{mehandru2025bioagents}, based on Phi-3 with orthogonal fine-tuning, reports better ROUGE scores than zero-shot GPT-4o; however, the comparison combines differences in model size, fine-tuning strategy, and multi-agent design. Similar challenges appear in modular RAG systems. BRAD~\cite{pickard2024language} and OLAF~\cite{riffle2025olaf} enhance GPT-3.5 and GPT-4 with agentic workflows and external bioinformatics tools, leading to higher faithfulness and relevance than the base models alone.}

\x{Similarly, Claude-3-Haiku with cache-augmented generation~\cite{nair2025prediction} and GPT-4o with ensemble prompting and confidence filtering~\cite{zhu2025enhancing} show that prompt-based methods and retrieval augmentation can substantially improve performance on text-based tasks. These observations suggest that differences between general-purpose and domain-specific models are often influenced by fine-tuning, retrieval support, and tool integration rather than architecture alone. At the same time, when domain-specific models achieve better results, the individual effects of pretraining data, tokenization methods, and task-specific fine-tuning are rarely examined separately. As a result, controlled studies that isolate these factors remain an important gap in the current literature.}

\subsubsection{Integration of Biological and Structural Knowledge}
A key limitation of standard language models in bioinformatics is their sequence simplicity. These models treat DNA, RNA, and protein sequences as simple one-dimensional strings, ignoring the complex 3D structures and biological context that affect function \cite{pantolini2024embedding,chen2024xtrimopglm}. The biological environment is mostly multi-dimensional. A protein's function depends on its folded shape, while a gene's expression is affected by a constantly changing mix of regulatory elements and cellular states \cite{ji2021dnabert,weissenow2022protein}. Recent studies show a clear and growing trend towards overcoming this issue by creating models that clearly include this essential context. Researchers are moving from just processing sequences to more comprehensive biological reasoning. The simplest method of integrating this is by providing structural information to PLMs. Older techniques relied on post-event analysis; however, more recent models have used structure as a central factor. More advanced models, for instance, Bingo \cite{ma2024bingo} utilizes a protein language model to capture sequence features and a Graph Neural Network (GNN) to analyze the protein's contact map, which serves as a proxy for its 3D structure. This combined method leads to frameworks, for example, ProtChatGPT \cite{wang2024protchatgpt}. This framework creates a rich representation by merging embeddings from the primary sequence, secondary structural elements, and tertiary atomic coordinates. At the same time, large-scale models such as ESMFold \cite{chen2025evaluating} have shown that, when sufficient parameter size and data are used, structural information can be derived from sequence data alone. 


This idea of contextual integration applies to systems biology with models, for instance, GET \cite{fu2025foundation}. GET predicts gene expression by incorporating cell-type-specific chromatin accessibility data, rather than relying solely on DNA sequences. This allows it to infer complex, context-dependent regulatory networks that static DNA-based models cannot capture. For models such as Top-DTI \cite{talo2025top}, a key challenge remains the effective alignment of multimodal information, as using language or topological features alone leads to suboptimal results.

The direction of GenAI in bioinformatics is clearly moving from models focused solely on sequences to those that are deeply integrated and aware of context \cite{fu2025foundation}. The most effective models integrate language modeling with insights from structural biology, functional genomics, and multi-omics \cite{ji2021dnabert,weissenow2022protein,chen2025evaluating}. This combination enables models to create more precise and biologically relevant representations, bridging the gap between one-dimensional sequences and actual biological function. Developing these multi-modal architectures is resource-intensive and requires access to high-quality, well-annotated, and diverse datasets \cite{chen2024xtrimopglm,chen2025evaluating}. Biological knowledge combined with generative models will be fundamental in the construction of systems that can predict, understand, and reason complex biological processes.

\subsubsection{Interpretability, Trust, and Clinical Reliability}

Although LLMs are transforming bioinformatics through prediction and generation, their black box nature poses a major challenge \cite{ji2021dnabert,meier2021language,zhang2022t4sefinder}. Scientific progress depends on both accurate predictions and explainability. Clinical decisions require clear and verifiable reasoning. Many studies acknowledge this limitation. While models such as DNABERT \cite{ji2021dnabert}, TrASPr+BOS \cite{wu2026generative}, and TrustAffinity \cite{badkul2024trustaffinity} perform well, their complex architectures offer little insight into the biological principles underlying their results. This lack of clarity is an important obstacle because it stops researchers from developing new biological hypotheses and challenges the trust needed for high-stakes applications.

In response, more research is focusing on techniques that analyze these complex models. A key method involves examining attention mechanisms. These mechanisms can show which parts of an input sequence the model considers most significant. For instance, DNABERT \cite{ji2021dnabert} uses attention-based visualization to find conserved relations in DNA sequences. Meanwhile, the Bingo \cite{ma2024bingo} uses attention maps and GNNExplainer to highlight specific amino acid residues and structural regions crucial for gene essentiality. More sophisticated approaches are also emerging. One example is using Linear Regression Tomography in the PhosF3C \cite{liu2025phosf3c} model to quantitatively assess the importance of learned features for predicting phosphorylation sites. These methods are essential first steps toward making models more transparent and useful for scientific discovery.

Trust also requires that models are both understandable and dependable, especially when faced with new data. A significant advancement in this area is adding uncertainty quantification. The TrustAffinity \cite{badkul2024trustaffinity} illustrates this by predicting protein-ligand binding affinity while also providing a confidence score. This allows users to evaluate how reliable the predictions are. Similarly, the multi-LLM model \cite{yang2026large} for cell type annotation offers uncertainty scores, clearly showing the model's confidence in its classifications. This feature is crucial for responsible use, as it enables models to recognize their limits and avoid overly confident mistakes with difficult inputs.

Combining interpretability and reliability is essential for clinical-grade performance, as errors can have serious consequences. Model hallucination, producing believable yet incorrect information, directly threatens clinical reliability \cite{hossain2025interpretable, green2025litsumm}. This concern arises in applications from conversational molecule optimization \cite{ye2025drugassist} to automated literature review \cite{smith2025funcfetch}. Even when models perform well, integrating them into clinical settings comes with many challenges. Researchers must navigate complex regulations and prove the models' reliability across diverse patient data, as indicated in the study on Lupus classification \cite{nair2025prediction}. Thus, while generative models hold great potential as decision-support tools, a significant gap remains between their current abilities and the high standards of safety, transparency, and trust needed for everyday use in patient care.

\x{A critical distinction that the current literature largely overlooks is the difference between computational plausibility and experimental validation. These are not equivalent, yet benchmark performance is routinely presented as evidence of biological utility. This standard for validation varies significantly depending on the output type. Experimental confirmation is performed through activity assays (wet laboratory experiments) and structural characterization for generated proteins such as those from ProtGPT2 \cite{ferruz2022protgpt2} and ProGen \cite{madani2020progen}. Although this is not the typical use, the only one that has found measurable lysozyme activity is the work of Madani et al. \cite{madani2023large}. For example, molecules generated by DrugAssist \cite{ye2025drugassist} \textit{in silico} property scores (solubility, BBBP) are not sufficient to support any therapeutic claim, and require pharmacokinetic profiling, toxicity screening, and validation in animal models. Recovering known conditioning attributes in the classifier is not a test of preservation of patient-level biological heterogeneity when using the synthetic omics profiles of P2GPT \cite{sidorenko2024precious2gpt}. Retrospective benchmark accuracy is not enough for a prospective, multi-population evidence standard and for clinical deployment of such clinical outputs as DrBioRight 2.0 \cite{liu2025drbioright} and PromptMRG \cite{jin2024promptmrg} have claimed. In all of these output types, a multi-layered validation process from \textit{in silico} benchmarks, held-out biological testing, experimental functional validation, and prospective clinical testing is needed before GenAI outputs can be responsibly applied to biological discovery, therapeutic design or patient care.}

\subsubsection{Computational Cost and Environmental Impact}
The impressive capabilities of GenAI in bioinformatics depend on massive scaling. Models with tens or hundreds of billions of parameters are trained on large datasets to capture the complex language of biology \cite{chen2024xtrimopglm,chen2025evaluating}; however, this requires substantial computational resources. Top-DTI highlights limitations such as dependence on high-quality molecular structures associated with topological feature extraction. The 15 \x{Billion}-parameter ESM-2 \cite{chen2025evaluating} and the 100 \x{Billion}-parameter xTrimoPGLM \cite{chen2024xtrimopglm} require access to large-scale, specialized computing systems. This often means using hundreds of high-end GPUs running for weeks or months. The ability to train or fine-tune SOTA models tends to be concentrated in a few well-funded industrial and academic labs. This situation limits the spread of advanced research and makes independent verification expensive.

An often overlooked consequence of high computational cost is the environmental impact. Training large-scale models consumes significant energy, resulting in a substantial carbon footprint \cite{chen2024xtrimopglm, talo2025top}. This raises important questions about the long-term sustainability of this research \cite{chen2025evaluating}. Although individual studies do not usually measure this impact, the scale of computations suggests significant energy use that should not be ignored. As the bioinformatics community increasingly depends on these powerful, however, resource-heavy methods, it faces an ethical challenge to balance scientific discovery with environmental responsibility \cite{jin2024genegpt,cui2024scgpt}.

To address the challenges of cost and sustainability, the field is increasingly focusing on computational efficiency. One notable strategy involves creating more efficient model architectures. An example is the Nucleotide Transformer v2 \cite{dalla2025nucleotide}, which outperformed its larger predecessor while using significantly fewer parameters. Another important innovation is the use of parameter-efficient fine-tuning (PEFT) techniques, such as the LoRA method seen in the PhosF3C model \cite{liu2025phosf3c}. This technique allows large pre-trained models to adapt to specific tasks with far less computational power than full retraining requires. Additionally, there is growing interest in using smaller, specialized language models. The BioAgents \cite{mehandru2025bioagents} highlights this, built on the smaller Phi-3 model to allow for local, accessible, and less resource-intensive deployment.

This focus on efficiency does not signal the end of large-scale modeling yet represents a move toward a more diverse and sustainable research ecosystem. The future of GenAI in bioinformatics will combine massive, centralized foundation models as shared resources with a range of smaller, optimized, and task-specific models \cite{liu2025phosf3c,mehandru2025bioagents}. This approach will broaden access and support more sustainable innovation. Such a balanced strategy is vital for ensuring that the transformative potential of GenAI is realized fairly and responsibly by the entire scientific community.

\subsubsection{Future Directions: Towards Modular, Efficient, and Bioinformatics-Grounded Models}

As GenAI continues to advance within bioinformatics, there is a clear transition from monolithic, standalone models \x{toward more modular, integrated, and accessible systems} \cite{smith2025funcfetch,ji2021dnabert}. Persistent challenges such as high computational demands, limited cross-domain generalization, and insufficient biological grounding are increasingly shaping the trajectory of innovation in this field. Growing consensus highlights three key priorities: (i) the development of modular and extensible frameworks, (ii) the optimization of computational efficiency, and (iii) the integration of multi-modal biological knowledge.

The expected future of applied bioinformatics AI is centered on modular architectures that manage collections of specialized tools rather than relying on a single, dominant model \cite{luo2022biogpt}. Recent approaches, including BRAD \cite{pickard2024language}, GeneGPT \cite{jin2024genegpt}, and OLAF \cite{riffle2025olaf}, exemplify this conceptual shift. In such systems, a generalist LLM functions as a reasoning engine, interpreting natural language queries and delegating specific tasks to established bioinformatics tools such as BLAST \cite{capela2025comparative}, molecular visualization software, or statistical analysis packages. This hybrid framework offers several advantages. It enhances factual reliability by grounding outputs in verifiable computational tools, mitigates hallucination during critical data retrieval processes, and facilitates incremental system updates by allowing individual components to be refined without retraining the entire model. Collectively, these advances suggest a future in which AI systems in bioinformatics evolve \x{from monolithic, encyclopedic systems into transparent, adaptive, and reliable scientific collaborators}.

In addition to the move toward modularity, there is a strong focus on improving computational efficiency and increasing accessibility \cite{liu2025phosf3c}. The high financial and environmental costs of training ever-larger foundation models highlight the need to move away from the assumption that larger models always perform better. Future efforts will focus on compute-optimal strategies that maximize performance within a set resource budget. This goal is being pursued through various methods, including creating more efficient architectures, as seen with the Nucleotide Transformer v2 \cite{dalla2025nucleotide}, performing better than its larger predecessor. The rise of PEFT techniques, for instance, LoRA, allow massive models to adapt to new tasks at a lower computational cost \cite{liu2025phosf3c}. This focus will help create a more varied ecosystem of models, including powerful and smaller open-source options \cite{mehandru2025bioagents}, ensuring that cutting-edge AI technology is accessible to more than just a few top institutions.

\begin{table*}[ht!]
\centering
\renewcommand{\arraystretch}{1.3} 
\setlength{\tabcolsep}{6pt}       
\scriptsize
\caption{Overview of GenAI methods across bioinformatics domains, including scaling and generalization, data availability and bias, integration of knowledge, interpretability, computational cost, and future strategies. The table summarizes each method’s key limitations and proposed future directions.}
\label{tab:method_limitations}
\begin{tabularx}{\textwidth}{
  @{\hspace{1pt}}
  >{\raggedright\arraybackslash}p{3.9cm} 
  @{\hspace{6pt}}
  >{\raggedright\arraybackslash}p{3cm}   
  @{\hspace{6pt}}
  c                                      
  @{\hspace{6pt}}
  >{\raggedright\arraybackslash}p{5cm}   
  @{\hspace{6pt}}
  X                                      
}
\toprule
\textbf{Type} & \textbf{Method} & \textbf{Year} & \textbf{Limitations} & \textbf{Future Directions} \\
\midrule

\multirow{4}{*}{\textbf{Scaling and Generalization}} 
& ProGen2 \cite{nijkamp2023progen2} & 2023 & Poor performance on narrow protein fitness spaces; limited transferability & Improve data quality; targeted fine-tuning for narrow biological tasks \\
& xTrimoPGLM \cite{chen2024xtrimopglm} & 2024 & Requires massive compute and specialized hardware; diminishing returns & Develop compute-efficient training; hybrid modular designs; sustainability-focused optimization \\
& ESM-2 \cite{chen2025evaluating} & 2025 & High computational cost, diminishing returns at scale & Focus on efficient transformer variants and parameter-efficient tuning \\
\midrule

\multirow{5}{*}{\textbf{Data Availability and Bias}}
& P2GPT \cite{sidorenko2024precious2gpt} & 2024 & Reliability depends on synthetic data quality & Robust validation of generated data; integration with real-world samples \\
& ESM \cite{chen2025evaluating} & 2025 & Performance drops in data-scarce domains & Augment with synthetic data; adapt transfer learning for rare proteins \\
& Nucleotide Transformer \cite{dalla2025nucleotide} & 2025 & Limited context window; sparse genomic coverage & Long-context modeling; improved genome-wide representations \\
& FuncFetch \cite{smith2025funcfetch} & 2025 & Relies on literature quality; risk of error propagation & Automated verification pipelines; cross-validation with curated datasets \\
& LORE \cite{li2025large} & 2025 & Potential inaccuracies in large-scale extraction tasks & Improve entity linking; refine extraction with expert-in-the-loop strategies \\
\midrule

\multirow{5}{*}{\textbf{Integration of Knowledge}}
& Bingo \cite{ma2024bingo} & 2024 & Dependency on contact maps as structure proxy & Integrate direct 3D experimental data; GNN with LLM hybridization \\
& ProtChatGPT \cite{wang2024protchatgpt} & 2024 & Resource intensive; multi-source embeddings not fully aligned & Modular multi-modal fusion; optimize structural \& functional reasoning \\
& GET \cite{fu2025foundation} & 2025 & Limited cell-type coverage; context-dependent biases & Scale to multi-cell datasets; integrate broader epigenomic information \\
& ESMFold \cite{chen2025evaluating} & 2025 & High reliance on scale; limited interpretability of folding predictions & Efficient folding with interpretable layers; integration with experimental structures \\
\midrule

\multirow{4}{*}{\textbf{Interpretability}}
& DNABERT \cite{ji2021dnabert} & 2021 & Black-box predictions; limited interpretability & Attention visualization and explainable embeddings \\
& TrustAffinity \cite{badkul2024trustaffinity} & 2024 & Lacks uncertainty quantification & Probabilistic calibration and confidence scoring \\
& Bingo \cite{ma2024bingo} & 2024 & GNN explanations limited in resolution & Combine structural attribution methods with interpretable embeddings \\
& PhosF3C \cite{liu2025phosf3c} & 2025 & Limited to specific phosphorylation site tasks & Generalize interpretability methods; learn cross-task XAI \\

\midrule

\multirow{5}{*}{\textbf{Computational Cost}}
& xTrimoPGLM \cite{chen2024xtrimopglm} & 2024 & Environmental impact of scaling & Green AI strategies and compute-optimal scaling laws \\
& ESM-2 \cite{chen2025evaluating} & 2025 & Extremely high training compute & LoRA and sparse transformers \\
& Nucleotide Transformer v2 \cite{dalla2025nucleotide} & 2025 & Still requires large-scale compute despite improvements & Further architecture efficiency; optimize compute–performance tradeoff \\
& PhosF3C \cite{liu2025phosf3c} & 2025 & Fine-tuning resource demand remains high & Extend LoRA-like methods; develop task-adaptive lightweight models \\
& BioAgents \cite{mehandru2025bioagents} & 2025 & Multi-agent workflows require expensive orchestration & Modular lightweight agents for distributed execution \\
\midrule

\multirow{4}{*}{\textbf{Future Strategies}}
& BRAD \cite{pickard2024language} & 2024 & Relies heavily on monolithic LLMs & Modular reasoning agents with tool augmentation \\
& GeneGPT \cite{jin2024genegpt} & 2024 & Hallucination risk when delegating tasks & Strengthen grounding in verified computational tools \\
& OLAF \cite{riffle2025olaf} & 2025 & Integration complexity across modules & Build scalable orchestration frameworks, focus on adaptability \\
& BioAgents \cite{mehandru2025bioagents} & 2025 & Coordination of agents is computationally heavy & Adaptive scheduling and decentralized execution \\
\bottomrule
\end{tabularx}
\end{table*}

Many existing models continue to represent complex biological entities as simple text strings. However, recent advanced research seeks to overcome this limitation. The successful integration of 3D structural information, functional annotations, and systems-level data, such as chromatin accessibility \cite{fu2025foundation}, demonstrates early progress in this direction. As outlined in several studies \cite{lyu2024gp, abir2025biollmnet}, the aim is to develop comprehensive \textit{in silico} models capable of jointly reasoning over sequences, structures, functions, and expression data. Such models would move beyond simple pattern recognition and enable genuine causal inference \cite{ferruz2022protgpt2}. This approach facilitates accurate prediction of \x{intervention} effects, such as gene edits or drug treatments, and supports the formulation of scientifically possible hypotheses for experimental validation. This level of integration is essential for realizing the full potential of GenAI, evolving it from a data analysis tool into an active component of scientific discovery.
These limitations and future directions are not restricted to a single challenge; however, they extend across scaling and generalization, data availability and bias, integration of knowledge, interpretability, computational cost, and future frameworks. To capture the broad spectrum, Table~\ref{tab:method_limitations} summarizes the GenAI methods, highlighting their year of publication, key limitations, and corresponding future directions within these bioinformatics domains.

\subsection{Dataset Analysis (RQ6)}
 \x{Datasets used in GenAI-based bioinformatics research can be broadly grouped into three categories: molecular-level datasets, cellular-level datasets, and textual or knowledge-based resources. These datasets serve different purposes throughout the model development process. Large-scale repositories are mainly used for pretraining, curated task-specific datasets support fine-tuning and benchmarking, and annotated knowledge bases provide grounding and retrieval information for LLM-based reasoning. Understanding the role of each dataset type is important for evaluating model generalization, identifying potential biases, and ensuring the reproducibility of GenAI applications in bioinformatics.}
 
 Comprehensive protein sequence repositories such as UniProtKB \cite{boutet2007uniprotkb}, UniRef90 \cite{suzek2007uniref}, BFD30 \cite{steinegger2018clustering}, CATH-Gene3D \cite{lewis2018gene3d}, and Pfam \cite{mistry2021pfam} provide widespread annotations that support protein language modeling and generative structure prediction. Datasets such as ProteinNet12 \cite{alquraishi2019proteinnet} and RCSB-PDB \cite{guo2023proteinchat} are helpful in training models for protein structure generation and prediction tasks, while ECOD \cite{cheng2014ecod} and TOP 1773 SUPERFAMILIES \cite{nallapareddy2023cathe} enhance model generalization across homologous protein families. {However, several issues related to data annotation and dataset representation limit the usefulness of these resources. UniProtKB~\cite{boutet2007uniprotkb} and UniRef90~\cite{suzek2007uniref}, which are widely used for pretraining protein language models such as ESM-2 and ProtT5, contain a large proportion of sequences from well-studied organisms, including Homo sapiens, E. coli, and other common model species. As a result, rare diseases, extremophiles, and many non-model organisms are less represented. This imbalance can affect the learned representations and reduce model performance when applied to less-studied biological systems. Similarly, although BFD30~\cite{steinegger2018clustering} reduces sequence redundancy through clustering, related proteins with similar structures or functions may still remain in the dataset. This can lead to overly optimistic evaluation results when training and test data originate from similar sources. In addition, Pfam~\cite{mistry2021pfam} and CATH-Gene3D~\cite{lewis2018gene3d} are widely used for protein annotation and benchmarking, yet they provide limited coverage of many uncharacterized protein families. Consequently, models evaluated only on these datasets may appear more effective than they are when applied to real-world metagenomic data or rare protein families.}

Datasets such as BindingDB \cite{gilson2016bindingdb} and ChEMBL31 \cite{mendez2019chembl} provide validated protein–ligand binding affinities through trial and error, advancing the development of generative drug design frameworks for modeling biomolecular interactions. \x{These datasets are widely used for fine-tuning and benchmarking protein-ligand interaction models. However, both datasets contain a large number of kinase inhibitors and central nervous system (CNS) drugs, while compounds such as allosteric binders, covalent inhibitors, and biologics are less represented. This imbalance may limit the ability of models to perform well across a broader range of drug classes. Similarly, SKEMPI~\cite{jankauskaite2019skempi} and ACE2~\cite{starr2020deep}, which are commonly used to evaluate mutational effects, are relatively small datasets and mainly include well-studied protein complexes. As a result, they provide only a limited assessment of how well models generalize to new and unseen biological systems.} SKEMPI \cite{jankauskaite2019skempi} and ACE2 \cite{starr2020deep} datasets focus on protein–protein interaction affinities, which enable the training of models for mutational effect prediction. Synthetic molecule repositories such as ZINC \cite{irwin2005zinc} and Mol-Instructions \cite{fang2023mol} provide vast molecular structures with textual explanations acceptable for multimodal GenAI frameworks. \x{ZINC~\cite{irwin2005zinc} is often used for pretraining generative molecular models because of its large size. However, many compounds in ZINC may be difficult or impractical to synthesize, which means that strong performance in computational evaluations does not always translate into molecules that can be readily produced in laboratory settings.} In genomics and transcriptomics, resources such as the Human Reference Genome \cite{schneider2017evaluation}, GenBank \cite{benson2012genbank}, and hg19 genome \cite{sanabria2023human} enable generative DNA modeling and variant effect prediction. Epigenomic datasets, for instance, Human DNA methylation \cite{xiong2022gmqn} and Mouse Brain scATAC-seq \cite{ma2020chromatin} contribute to generative modeling of chromatin availability and epigenetic regulation.

\x{Single-cell and multi-omics datasets are important resources for GenAI models that learn cellular diversity and gene expression patterns. CELLxGENE~\cite{czi2025cz}, Human Lung scRNA-seq~\cite{wang2020single}, the Pancreas, Immune, and Lung Atlas~\cite{luecken2022benchmarking}, and The Healthy Heart Atlas~\cite{kanemaru2023spatially} are commonly used for pretraining and benchmarking models such as scGPT~\cite{cui2024scgpt} and scBERT~\cite{yang2022scbert}. However, most samples in these datasets come from adult human tissues collected in well-funded research environments, while pediatric, elderly, and rare disease populations are less represented. Differences in tissue collection methods, sequencing platforms, and laboratory procedures can also introduce batch effects that are not always fully corrected. As a result, models may learn technical patterns in addition to true biological signals. Large cellular resources such as the Human and Mouse Cell Atlas~\cite{simon2021integration} and 10x Genomics Multiome~\cite{zheng2017massively} are widely used for multimodal pretraining and data integration, but the quality of cell-type annotations varies across studies. Rare cell populations are often grouped into broader categories or omitted because of limited sample numbers. GTEx~\cite{lonsdale2013genotype}, ARCHS4 RNA-seq~\cite{seal2023genenames}, and CCLE~\cite{barretina2012cancer} are frequently used for transcriptomic modeling, fine-tuning, and validation. However, CCLE is based on cancer cell lines and may not fully reflect the complexity of tumors in living organisms, while GTEx mainly contains healthy adult tissues and therefore provides limited coverage of disease-related gene expression patterns. CPTAC \cite{edwards2015cptac} and RPPA \cite{li2024protein} function as primary proteomic references for GenAI models focusing on protein expression synthesis and cancer proteogenomics. Integrative datasets such as OGEE \cite{gurumayum2021ogee} and RegPhos \cite{lee2011regphos} further enable predictive modeling of gene essentiality and PTMs, respectively.}

\begin{table*}[ht!]
\centering
\scriptsize
\caption{A comprehensive collection of key datasets employed in GenAI and LLM applications for bioinformatics, spanning molecular, cellular, and textual modalities.}
\begin{tabularx}{\textwidth}{p{3.6cm} p{3.7cm} p{1cm} | p{3.6cm} p{3.7cm} p{1cm}}
\toprule
\textbf{Dataset} & \textbf{Dataset Type} & \textbf{Ref.} & \textbf{Dataset} & \textbf{Dataset Type} & \textbf{Ref.} \\
\midrule
Human Reference Genome & Genomic & \cite{schneider2017evaluation} & Human DNA methylation & Epigenetic & \cite{xiong2022gmqn} \\
Mouse Brain scATAC-seq & Epigenomic & \cite{ma2020chromatin} & Mol-instructions & Biomolecule & \cite{fang2023mol} \\
CCLE & Multi-omics & \cite{barretina2012cancer} & ZINC & Synthetic small molecules & \cite{irwin2005zinc} \\
Pfam & Protein function & \cite{mistry2021pfam} & OMIM & Gene-Genetic disorder & \cite{amberger2015omim}\\
DMS & Protein function & \cite{riesselman2018deep} & UniPort & Gene-Protein pair & \cite{coudert2023annotation} \\
TOP 1773 SUPERFAMILIES & Protein sequence & \cite{nallapareddy2023cathe} & QuanTest2 & Multiple sequence alignments & \cite{sievers2020quantest2} \\
UniProtKB & Protein sequence & \cite{boutet2007uniprotkb} & CPTAC & Cancer proteomics & \cite{edwards2015cptac} \\
SecReT4 v2.0 & Protein sequence & \cite{bi2013secret4} & RPPA & Quantitative proteomics & \cite{li2024protein} \\
UniRef90 & Protein sequence &\cite{suzek2007uniref} & OGEE & Gene essentiality & \cite{gurumayum2021ogee} \\
BFD30 & Protein sequence & \cite{steinegger2018clustering} & BindingDB & Protein-Ligand interaction & \cite{gilson2016bindingdb} \\
CATH-Gene3D & Protein sequence & \cite{lewis2018gene3d} & ChEMBL31 & Protein-Ligand interaction & \cite{mendez2019chembl} \\
ProteinNet12 & Protein structure & \cite{alquraishi2019proteinnet} & RegPhos & Kinase-specific phosphorylation & \cite{lee2011regphos} \\
RCSB-PDB & Protein sequence & \cite{guo2023proteinchat} & PubMedQA & Biomedical QA & \cite{jin2019pubmedqa} \\
ECOD & Protein sequence & \cite{cheng2014ecod} & 1743 bioinformatics repositories & Bioinformatics articles & \cite{russell2018large} \\
ProtDescribe & Protein sequence & \cite{xu2023protst} & Rdatasets & General purpose archieve & \cite{arel2023rdatasets} \\
CD-HIT & Protein sequence & \cite{li2006cd} & GTDB & Taxonomic and phylogenetic & \cite{parks2018standardized} \\
UniParc & Protein sequence & \cite{leinonen2004uniprot} & GeneHop & Multi-Hop & \cite{jin2024genegpt} \\
CELLxGENE & scRNA-seq & \cite{czi2025cz} & GSC & Biomedical Text Mining & \cite{wu2021mining} \\
Human Lung scRNA-seq & scRNA-seq & \cite{wang2020single} & SKEMPI & Protein binding affinity & \cite{jankauskaite2019skempi} \\
Pancreas, human immune, and lung atlas & scRNA-seq &\cite{luecken2022benchmarking} & ACE2 & Protein binding affinity & \cite{starr2020deep} \\
The Healthy Heart & scRNA-seq & \cite{kanemaru2023spatially} & GSM8k & Math Word Problems & \cite{cobbe2021training} \\
Human and Mouse cell Atlas & scRNA-seq & \cite{simon2021integration} & GenBank & Nucleotide sequences & \cite{benson2012genbank} \\
10x Genomics Multiome & scRNA-seq & \cite{zheng2017massively} & hg19 genome & Genome assembly & \cite{sanabria2023human} \\
HomFam & Protein Sequence Alignment & \cite{sievers2011fast} & GTEx & Gene expresion & \cite{lonsdale2013genotype} \\
extHomFam & Protein Sequence Alignment & \cite{deorowicz2016famsa} & ARCHS4 RNA-seq & Gene expresion & \cite{seal2023genenames} \\
Prop3D-20sf & 3D protein structural & \cite{draizen2024deep} & PMKB-CV & Disease-gene associations & \cite{li2025large} \\
RPI1460 & RNA-Protein Interaction & \cite{huang2022lpi} & DisGenNet & Disease-gene associations & \cite{pilault2020extractive} \\
The 1000 Genomes & Human genomics & \cite{byrska2022high} & ADNI & Disease-gene associations & \cite{petersen2010alzheimer} \\
\bottomrule
\end{tabularx}
\label{tab:dataset}
\end{table*}

\x{Text-based datasets play an important role in developing LLMs for biomedical and bioinformatics applications. They are commonly used for pretraining, fine-tuning, and retrieval-based grounding. PubMedQA~\cite{jin2019pubmedqa} and GSC Biomedical Text Mining~\cite{wu2021mining} are widely used for training models in biomedical question answering and literature analysis. However, PubMedQA covers a limited range of topics, and models evaluated only on this dataset may not perform equally well on broader clinical or experimental reasoning tasks. Knowledge resources such as OMIM~\cite{amberger2015omim}, DisGeNet~\cite{pilault2020extractive}, and GeneHop~\cite{jin2024genegpt} provide structured information on gene--disease relationships and are often used to support retrieval-based and multi-step reasoning systems such as GeneGPT~\cite{jin2024genegpt} and GP-GPT~\cite{lyu2024gp}. However, these databases contain more information about common diseases and well-studied genes, while rare genetic disorders and non-coding regulatory elements are less represented. ProtDescribe~\cite{xu2023protst} and the collection of 1743 Bioinformatics Repositories~\cite{russell2018large} provide textual descriptions that support instruction-tuned and multimodal models, but the quality and completeness of these annotations differ across protein families. General-purpose datasets such as GSM8k~\cite{cobbe2021training} and Rdatasets~\cite{arel2023rdatasets} are sometimes used to improve mathematical and data reasoning abilities in biomedical-bioinformatics LLMs. However, these datasets are not specifically designed for bioinformatics, and are most suitable for tasks involving numerical analysis or structured data processing rather than biological sequence analysis.}

\x{Beyond scale and modality, dataset quality and accessibility strongly influence the reliability of GenAI models in bioinformatics. Widely used resources such as UniProtKB \cite{boutet2007uniprotkb}, ProteinNet \cite{alquraishi2019proteinnet}, CELLxGENE \cite{czi2025cz}, PubMedQA \cite{jin2019pubmedqa}, and ZINC \cite{irwin2005zinc} are openly accessible and support reproducible research, whereas datasets containing human genomic \cite{byrska2022high, sanabria2023human}, transcriptomic \cite{benson2012genbank}, or clinical data \cite{edwards2015cptac, petersen2010alzheimer} may require controlled access because of privacy and ethical considerations. Annotation quality also varies across datasets, with experimentally validated labels generally providing more reliable supervision than automatically inferred annotations. In addition, current datasets are uneven in their taxonomic and disease coverage. Resources such as UniProtKB \cite{boutet2007uniprotkb} and ProteinNet \cite{alquraishi2019proteinnet} are dominated by well-studied organisms, while many non-model species remain underrepresented. Similarly, biomedical text and omics datasets \cite{jin2024genegpt, jin2019pubmedqa, russell2018large, petersen2010alzheimer} often focus on common diseases, providing limited coverage of rare diseases and less-studied biological conditions. These limitations may affect model generalizability and lead to uneven performance across organisms and disease contexts.}

Together, these datasets lead to the creation of GenAI systems that synthesize biological knowledge on different scales, from nucleotide and protein domains, cells, and the biomedical literature. The integration of structured biological data with unstructured text archives represents a transformative shift in bioinformatics that enables foundation models to engage in generative reasoning, data augmentation, and cross-modal translation with high biological precision. \x{General-purpose datasets are sometimes incorporated into bioinformatics GenAI pipelines to improve molecular representation learning, support numerical and statistical reasoning, and evaluate a model's ability to analyze structured tabular data in biological applications.} The key datasets utilized in GenAI and LLM research in bioinformatics provide the essential foundation for corresponding models and applications discussed in this section, and a summary of these resources is presented in Table \ref{tab:dataset}.

\section{Discussion}
\label{discussion}
This review has been structured and discussed around six RQs that together present a comprehensive evaluation of GenAI in bioinformatics and highlight its rapid and transformative impact on the field. By analyzing recent advances, we outline how generative models support diverse biological tasks from protein design to \x{multi-omics} integration, along with the challenges. Research trends indicate a clear shift from classifying to generative methods, offering novel opportunities. However, ongoing issues such as data dependency, model interpretability, and the need for thorough biological validation indicate the importance of careful and systematic research as the field progresses \cite{ferruz2022protgpt2,wan2023integrating,bolton2024biomedlm,shree2023scdreamer}.

\textbf{Integration and Evolution of GenAI Applications.} In the application domain (RQ1), GenAI research demonstrates a clear shift from task-specific models toward unified and context-aware frameworks \x{across} genomics, proteomics, and single-cell biology. The introduction of foundational models such as DNABERT \cite{ji2021dnabert}, ProGen \cite{madani2020progen}, and scGPT \cite{cui2024scgpt} indicates a move toward biologically grounded architectures capable of cross-domain learning. These systems no longer rely solely on manually designed features or static datasets; \x{instead}, they learn biological representations that generalize within diverse modalities. Moreover, integration-focused tools such as sciCAN \cite{xu2022scican} and xTrimoPGLM \cite{chen2024xtrimopglm} highlight progress in multi-omics fusion and structure–function prediction, \x{extending GenAI capabilities beyond individual molecule-level models}. Interactive agents such as Top-DTI \cite{talo2025top}, ProtChatGPT \cite{wang2024protchatgpt}, and OLAF \cite{riffle2025olaf} further bridge natural language reasoning with achievable bioinformatics workflows and represent a step toward explainable and human-centered AI. Collectively, these advances show the progression of GenAI from separate predictive tools to an integrated generative framework that improves discovery and research across the bioinformatics workflow.

\textbf{Specialized Model Development and Adaptation. }Recent developments in GenAI-based bioinformatics reveal a clear transition from general-purpose language models to domain-specialized architectures (RQ2) \x{optimized} for biological sequences, structures, and graphs. Models such as DNABERT \cite{ji2021dnabert} and ESM-2 \cite{meier2021language} demonstrate the way biologically informed tokenization and domain-specific pretraining deliver substantial accuracy gains within genomics and proteomics tasks. Multimodal frameworks, for instance, TrustAffinity \cite{badkul2024trustaffinity}, Bingo \cite{ma2024bingo}, further extend this advantage by integrating structural and graph-based features that enable high precision in protein–ligand affinity prediction and cross-species gene analysis. Similarly, single-cell models such as scGPT \cite{cui2024scgpt} highlight the growing value of sequence–omics fusion for cell type annotation and response prediction under biological intervention.

\textbf{Scope of GenAI Across Bioinformatics Domains.} The review highlights that GenAI’s impact extends widely throughout core domains of bioinformatics, such as protein design, genomic, transcriptomics, and \x{multi-omics} integration (RQ3). Each domain demonstrates unique benefits enabled by domain-specific model adaptation. In protein bioinformatics, models such as ESM-2 \cite{chen2025evaluating} and ProGen2 \cite{nijkamp2023progen2} enable proper mutation effect prediction, structural modeling, and \textit{de novo} protein design. In genomics, DNABERT \cite{ji2021dnabert} and GeneGPT \cite{jin2024genegpt} interpret genomic grammar and link genetic variants to attributes, supporting breakthroughs in medical genetics and medicine. Models such as scGPT \cite{cui2024scgpt} have advanced transcriptomics by revealing cellular states, interactions, and developmental paths. In addition, \x{multi-omics} systems such as P2GPT \cite{sidorenko2024precious2gpt} and OLAF \cite{riffle2025olaf} integrate data across molecular layers and simplify complex analyses.

\textbf{From Prediction to Generation and Design.} GenAI has transformed bioinformatics from prediction to active design and interaction (RQ4). Early AI focused on classifying or predicting existing biological development. In contrast, models such as ProGen2 \cite{nijkamp2023progen2} and ProtGPT2 \cite{ferruz2022protgpt2} can generate entirely new protein sequences with predictable functions and extend to multi-omics by generating synthetic data to enrich limited datasets in bioinformatics research. AI has shifted from analyzing biological data to the active generation and design of biological systems. \x{However, a fundamental challenge is that the field mixes computational plausibility with experimental validation. Although GenAI models can generate promising proteins, molecules, synthetic omics data, and clinically relevant outputs, their practical value cannot be established through computational evaluation alone. Experimental evidence, including functional activity assays, pharmacokinetic studies, and biological validation, is essential to determine whether these generated outputs are biologically meaningful and suitable for downstream applications. Such discrepancy between simulation and practical utility in applying GenAI remains the greatest unresolved challenge in Bioinformatics.} \x{Addressing these challenges is essential for translating generative capabilities into reliable bioinformatics applications.}

\textbf{Diversity and Strategic Role of Datasets. }The dataset analysis (RQ6) reveals that GenAI in bioinformatics is supported by a highly diverse, still interdependent dataset that ranges molecular, cellular, and textual modalities. At the molecular level, resources, such as UniProtKB \cite{boutet2007uniprotkb}, ProteinNet12 \cite{alquraishi2019proteinnet}, and Pfam \cite{mistry2021pfam} indicate strong significance on protein sequence and structure learning, and also show GenAI’s advancement in protein representation and generation tasks. Cellular and multi-omics datasets, such as CELLxGENE \cite{czi2025cz}, GTEx \cite{lonsdale2013genotype}, and CCLE \cite{barretina2012cancer} highlight the growing use of integrative frameworks capable of modeling cell heterogeneity and cross-tissue interactions. Meanwhile, textual and knowledge-based datasets such as PubMedQA \cite{jin2019pubmedqa} and OMIM \cite{amberger2015omim} reveal improvement toward cross-domain learning from sequences to literature.

\textbf{Limitations and Future Research Agendas.} 
The review identifies some core limitations (RQ5) in current GenAI approaches for bioinformatics. Scaling models such as ESM-2 and xTrimoPGLM have improved biological sequence understanding; however, their training demands broad computational resources and still give decreasing performance gains across tasks \cite{chen2024xtrimopglm,chen2025evaluating}. \x{Generalization remains weak for rare proteins and underrepresented biological contexts}, as seen in ProGen2 \cite{nijkamp2023progen2}. Furthermore, LLM-augmented architecture with biomedical APIs \x{are prone to} incorrect parameter usage in scientific reasoning tasks \cite{jin2024genegpt}. Additionally, standard language models often treat DNA, RNA, and protein data as linear text sequences, overlooking the spatial and contextual complexity of biological systems \cite{pantolini2024embedding,chen2024xtrimopglm}. However, a protein’s function relies on its folded shape, and gene expression is influenced by dynamic regulatory and cellular environments \cite{ji2021dnabert,weissenow2022protein}.

The future agenda for GenAI in bioinformatics is defined by three interconnected priorities (RQ5) aimed at resolving current limitations. The initial priority is to develop a modular and extensible architecture independent of any single model, in which a generalist LLM coordinates tasks among specialized bioinformatics tools within a hybrid system to increase reliability and reduce hallucinations \cite{pickard2024language, jin2024genegpt, riffle2025olaf}. \x{The second priority is to improve computational efficiency and accessibility by moving beyond the assumption that larger models inherently perform better, instead adopting compute-optimal training strategies, parameter-efficient fine-tuning methods, and systems-level data integration approaches} \cite{dalla2025nucleotide, liu2025phosf3c, fu2025foundation, lyu2024gp}. The third and most significant future direction is the deep integration of multimodal biological knowledge to develop \textit{in silico} models that move beyond text representations and enable joint reasoning over sequences and 3D structures. Finally, frequent interpretability and uncertainty estimation issues directly affect the adoption of clinical and translational research. Future research should focus more on creating transparent GenAI models that generate interpretable latent features. The integration of these models into biologically informed deployment pipelines, in alignment with regulatory and ethical requirements, is essential to close the gap between research prototypes and trustworthy clinical \x{tools} \cite{gao2024documenting, atf2025challenge}.

\section{Conclusion}
\label{conclusion}
The involvement of GenAI in bioinformatics is redefining the approach to biological discovery, entering an era where models can design proteins, generate synthetic omics data, and collaboratively assist in advanced biological workflows. Our study addresses the gaps identified in previous literature \x{through an in-depth and integrative analysis not collectively offered by prior studies}. To achieve this, we formulated six RQs that comprehensively explore the methodological and application-oriented domains of GenAI in bioinformatics.

This work highlights GenAI-enabled de novo protein design, multi-omics integration, and biologically grounded modeling, which outperform traditional methods in accuracy and scalability. A particular strength is its comparative analysis between general-purpose and domain-specific language models, revealing how fine-tuned, biologically contextualized models outperform general-purpose LLMs. Additionally, its integrative evaluation of bioinformatics domains reveals how GenAI models collectively advance beyond the capabilities of traditional computational approaches. This work also presents a comprehensive evaluation of task-specific advancements enabled by GenAI and highlights that diverse molecular, cellular, and textual datasets collectively support generative modeling and improve biological understanding.

Our study discusses the key limitations, such as high computational cost and \x{weak} generalization, and challenges in existing GenAI applications within bioinformatics, and proposes a structured future research agenda, such as transparent and interpretable models and \x{integration} of multimodal knowledge, to guide and refine ongoing developments in the field. Finally, our review is highly beneficial as it addresses the gaps overlooked by previous studies and presents an integrative understanding of GenAI methodologies and datasets. It also supports the development of more efficient, accurate, and biologically informed computational models in bioinformatics and computational biology.

\section*{Declarations}
\noindent
\textbf{Conflict of Interests:} On behalf of all authors, the corresponding author states that there is no conflict of interest.
\textbf{Funding:} No external funding is available for this research.\\
\textbf{Data Availability Statement:} Not Applicable.\\
\textbf{Ethics Approval and Consent to Participate}. Not Applicable.
\\
\textbf{Author Contributions:}
\textit{Conceptualization and Methodology:} Wasimul Karim, Riasad Alvi, Sayeem Been Zaman, Arefin Ittesafun Abian, Mohaimenul Azam Khan Raiaan; \textit{Resources and Literature Review:} Wasimul Karim, Riasad Alvi, Sayeem Been Zaman;
\textit{Writing – Original Draft Preparation:} Wasimul Karim, Riasad Alvi, Sayeem Been Zaman, Arefin Ittesafun Abian, Mohaimenul Azam Khan Raiaan;
\textit{Validation:} Arefin Ittesafun Abian, Mohaimenul Azam Khan Raiaan, Saddam Mukta, Md Rafi Ur Rashid, Md Rafiqul Islam, Yakub Sebastian, Sami Azam;
\textit{Formal Analysis:} Arefin Ittesafun Abian, Mohaimenul Azam Khan Raiaan, Saddam Mukta, Md Rafi Ur Rashid, Md Rafiqul Islam, Yakub Sebastian, Sami Azam;
\textit{Writing – Reviewing and Finalization:} Arefin Ittesafun Abian, Mohaimenul Azam Khan Raiaan, Sami Azam;
\textit{Project Supervision:} Arefin Ittesafun Abian, Mohaimenul Azam Khan Raiaan, Sami Azam;


\end{document}